\RequirePackage{fix-cm}
\documentclass[twocolumn]{svjour3}          
\smartqed  
\usepackage{graphicx}
 \usepackage{mathptmx}      
\usepackage[table]{xcolor}
\usepackage{booktabs} 
\usepackage{amsmath}
\usepackage{amsfonts}
\usepackage{changepage,threeparttable} 
\usepackage{soul}
\usepackage{multirow}
\usepackage[utf8]{inputenc}
\usepackage{calc}
\usepackage{hyperref}

\DeclareUnicodeCharacter{2207}{$\nabla$}

\usepackage{adjustbox}
\usepackage{array}
\usepackage{pstricks} 

\usepackage[abs]{overpic}
\usepackage{pict2e}

\definecolor{brightcerulean}{rgb}{0.11, 0.67, 0.84}
\definecolor{bondiblue}{rgb}{0.0, 0.58, 0.71}
\definecolor{emerald}{rgb}{0.31, 0.78, 0.47}
\definecolor{pigment}{rgb}{0.0, 0.65, 0.31}
\newcommand{\NEW}[1]{{#1}}

\gdef\etal{\textit{et al.}}
\gdef\eg{\textit{e.g.}}
\newcommand{\x}{\mathrm{x}}
\newcommand{\n}{N}
\newcommand{\vect}[1]{\mathrm{#1}}

\newcommand{\beyLamb}[1]{\textcolor{black}{{#1}}}

\usepackage{pifont}
\newcommand{\cmark}{\ding{51}}%

\newcommand{\Aa}{$A$}
\newcommand{\Nn}{$N$}
\newcommand{\Dd}{$D$}
\newcommand{\Ss}{$S$}

\newcommand{\Ll}{$L$}

\newcommand{\shortcite}[1]{\cite{#1}}

\newcolumntype{R}[2]{%
	>{\adjustbox{angle=#1,lap=\width-(#2)}\bgroup}%
	l%
	<{\egroup}%
}
\newcommand*\rot{\multicolumn{1}{R{35}{1em}}}
\newcommand*\rotf{\multicolumn{1}{R{45}{1em}}}

\gdef\Aflat{A$_{\mbox{f}}$} 
\gdef\Asparse{A$_{\mbox{sp}}$} 
\gdef\Ssmooth{S$_{\mbox{sm}}$} 

\begin{document}
\sloppy

\title{A Survey on Intrinsic Images 
}
\subtitle{Delving Deep Into Lambert and Beyond}


\author{Elena Garces         \and
        Carlos Rodriguez-Pardo \and
        Dan Casas				\and
        Jorge Lopez-Moreno 
}


\authorrunning{E. Garces \& C. Rodriguez-Pardo \& D. Casas \& J. Lopez-Moreno} 

\institute{Elena Garces \at
           SEDDI \& Universidad Rey Juan Carlos, Madrid, Spain   \\                   
           \email{elena.garces@seddi.com}           
           \and
           Carlos Rodriguez-Pardo \at
           SEDDI \&  Universidad Carlos III, Madrid, Spain \\         
           \email{carlos.rodriguezpardo.jimenez@gmail.com}
           \and
           Dan Casas \at
		   Universidad Rey Juan Carlos, Madrid, Spain \\
		   \email{dan.casas@urjc.es}
\and
           Jorge Lopez-Moreno \at
            Universidad Rey Juan Carlos \& SEDDI, Madrid, Spain  \\
			\email{jorge@jorg3.com}             
}

\date{Received: date / Accepted: date}

\maketitle

\begin{abstract}
	Intrinsic imaging or intrinsic image decomposition has traditionally been described as the problem of decomposing an image into two layers: a reflectance, the albedo invariant color of the material; and a shading, produced by the interaction between light and geometry. Deep learning techniques have been broadly applied in recent years to increase the accuracy of those separations.
	In this survey, we overview those results in context of well-known intrinsic image data sets and relevant metrics used in the literature, discussing their suitability to predict a desirable intrinsic image decomposition. \\
	Although the Lambertian assumption is still a foundational basis for many methods, we show that there is increasing awareness on the potential of more sophisticated physically-principled components of the image formation process, that is, optically accurate material models and geometry, and more complete inverse light transport estimations. 
	We classify these methods in terms of the type of decomposition, considering the priors and models used, as well as the learning architecture and methodology driving the decomposition process. We also provide insights about future directions for research, given the recent advances in neural, inverse and differentiable rendering techniques.
\end{abstract}


\section{Introduction}

Images, as two-dimensional projections that depict the world around us, can be described as a harmonious combination of colors, shades, and shadows. Understanding how an image is generated by the complex interaction between light and matter has been a subject of research for decades: the light rays that carry all the information about a given scene are integrated into the RGB values of the camera sensor, turning the process of recovering back the original scene into an ill-posed problem.
In the computer graphics and vision literature, two main approaches target the problem of recovering the underlying properties of the scene elements (such as lights, geometry, and materials): 
inverse rendering, and intrinsic decomposition methods. 
Although they share a common root, these two problems have been traditionally tackled from two perspectives resulting in different outcomes.

Inverse rendering methods have the goal of representing the scene digitally in a way that it allows to photo-realistically \textit{re-render} novel views of it. This means estimating the parameters required by the render equation such as geometry, lights, materials, or the camera model. This is extremely challenging given a single image as input, even forward render engines sometimes merely approximate the complex light phenomena. Traditional approaches relied on manual intervention to aid modeling arbitrary geometries \cite{oh2001image}, or established priors about the shape narrowing the scope to \textit{e.g.} faces~\cite{zollhofer2018state,blanz1999morphable}, humans~\cite{kanamori2018relighting}, flat materials~\cite{dong2011appgen,dong2019deep}, or single objects~\cite{han2019image}. 
Nowadays, inverse rendering has undergone a major disruption in the way the problem is being tackled: First, deep neural networks, as powerful universal approximators, have reduced the need to define the scene elements explicitly, as shown by neural rendering methods~\cite{tewari2020starneural}. Second, differentiable renders~\cite{Li2018DMC,nimier2019mitsuba,zhao2020physics,loubet2019reparameterizing}, by allowing to compute direct derivatives of the images with respect to arbitrary scene parameters, have enabled physically-based end-to-end inverse parameter estimation.
Neural rendering techniques are reaching high degrees of realism for reproducing any kind of scene ~\cite{mildenhall2020nerf,martin2020nerf}, although this comes at the cost of limiting the manipulation of the scene parameters. Differentiable rendering, although it cannot yet cope with arbitrary scene setups, is showing promising results towards this end. 

Intrinsic decomposition, which can be seen as a simplification of inverse rendering for general scenes, aims to provide interpretable intermediate representations that prove useful for intelligent vision systems or to allow local material edits that do not require changes in lighting or viewpoint. 
The intrinsic scene model~\shortcite{barrow1978recovering} described the world as the combination of three \textit{intrinsic} components --surface reflectance, distance or surface orientation, and incident illumination. A fundamental observation for defining these layers is that the human visual system understands them independently of viewing and lighting conditions, even if it is not familiarized with the scene or the objects. In practice, most of the methods have referred to intrinsic image decomposition as the problem of separating the reflectance and shading layers, assuming a Lambertian world.
The Retinex theory~\cite{land1971lightness,horn1974determining} was fundamental for many of the algorithms developed during the last two decades~\cite{finlayson2004intrinsic,garces2012intrinsic,bi20151,bousseau2009user,bell2014intrinsic} providing some basic priors about how shading and reflectance typically behave in our retina: a change in reflectance cause sharp gradients, while a change in shading cause smooth gradient variations.
Until recently, most methods relied only on such kinds of cues (or heuristics) derived from low-level understanding of the physics of the light or empirical observations. 
With the advent of deep learning, current solutions have posed the problem as end-to-end network architectures which learn to predict the reflectance and shading layers given huge amounts of data as training. A key difference with respect to traditional solutions is that learned models also take into account the global semantics of the scene, while previous methods performed mostly at the local level (gradients and edges). 

Deep learning-based solutions for the intrinsic decomposition problem have facilitated more complex scene models beyond the Lambertian one~\cite{maxwell2008bi,tominaga1994dichromatic,SigPBR015,lafortune1994using}, and also to estimate some of the scene elements, such as illumination and geometry~\cite{janner2017self,yu2019inverserendernet,sengupta2019neural,zhou2019glosh,li2020inverse}. 
While this will ultimately enable photo-realistic arbitrary scene manipulations (goal shared with image-based inverse rendering), some critical aspects make evaluating the contributions of each new method a difficult process:
there is a variety of datasets that contain diverse types of objects, scenes, and labels; non-standardized and low-quality quantitative metrics to compare the methods with; and a lack of a unified methodology to universally evaluate the progress.

With this survey, we would like to review the current status of learning-based solutions that may serve to inspire and guide future research in the fields of computer graphics and vision. In particular: First, we review the connection of the intrinsic decomposition problem from a forward and inverse physically-based rendering perspective, hoping that this will clarify doubts and inspire more complex approaches outside the Lambertian assumption. Second, we show a taxonomy of current datasets, learning strategies, and architectures, putting them in context of traditionally non-learning-based solutions, we also discuss their main advantages and limitations. Third, we gather quantitative evaluations of these methods according to commonly used metrics and show qualitative results for difficult cases. Finally, we conclude with open research opportunities. 

\textit{In addition to the review presented in the paper, we provide a \href{http://www.elenagarces.es/projects/SurveyIntrinsicImages/}{web project} which will contain the compendium of datasets, metrics, papers and their performance, which we will keep updated with the latest research.}

\paragraph*{\textbf{Related Surveys}}~Intrinsic image decomposition problem has been reviewed before in the STAR report of Bonneel et al.~\cite{bonneel2017intrinsic}, where several --non-deep learning-based-- algorithms were evaluated in the context of image editing tasks: logo removal, shadow removal, texture replacement, and wrinkles attenuation. 
Since then, dozens of new papers have tackled the problem from a purely data-driven perspective. Our work is complementary to theirs, as we review the approaches not covered there, which propose solutions based on deep learning frameworks.
Neural rendering has been reviewed in a recent survey~\cite{tewari2020starneural}. Similarly, inverse rendering and image-based rendering has been widely studied in several surveys: for generic scenes~\cite{patow2003survey}, for particular applications like faces~\cite{zollhofer2018state} or materials~\cite{dong2019deep}, and for image-based 3D object reconstruction~\cite{han2019image}. 
Despite the vast amount of papers tackling both the problem of intrinsic decomposition and inverse rendering, the explicit connection between both fields has not been addressed before. Moreover, as we will discuss in this paper, the connection of intrinsic imaging with the actual optical properties of materials is very relevant, and thus, a survey on its representation and acquisition~\cite{Guarnera2016} could be a good reading for practitioners in the field.

\paragraph*{\textbf{Scope}}~In this survey, we cover several recent papers that propose a deep learning-based solution to estimate the intrinsic components of the scene given a single image as input. We thoroughly review those which include quantitative metrics of performance, either using the IIW dataset~\cite{bell2014intrinsic}, which contains indoor real scenes and scores given by human raters, or using the MIT Intrinsic dataset~\cite{grosse2009ground}, which contains isolated painted figurines. Although most of the papers discussed use the Lambertian material model, some of the most recent ones include more complex ones or model other scene elements such as illumination or geometry. This latter approach resembles image-based inverse rendering methods, so we further discuss the ones which explicitly model part of the scene elements (illumination, material, geometry). Among these methods, we include a brief overview of the ones which target particular domains (faces, humans, flat materials). Finally, we do not review methods that target specific applications such as relighting, colorization or texture editing, or intrinsic decomposition methods which do not provide quantitative performance comparisons, unless they prove useful to facilitate the discussion.

 \section{Theoretical Background}
\label{sec:theory}

\begin{figure}[htb]
	\includegraphics[width=\linewidth]{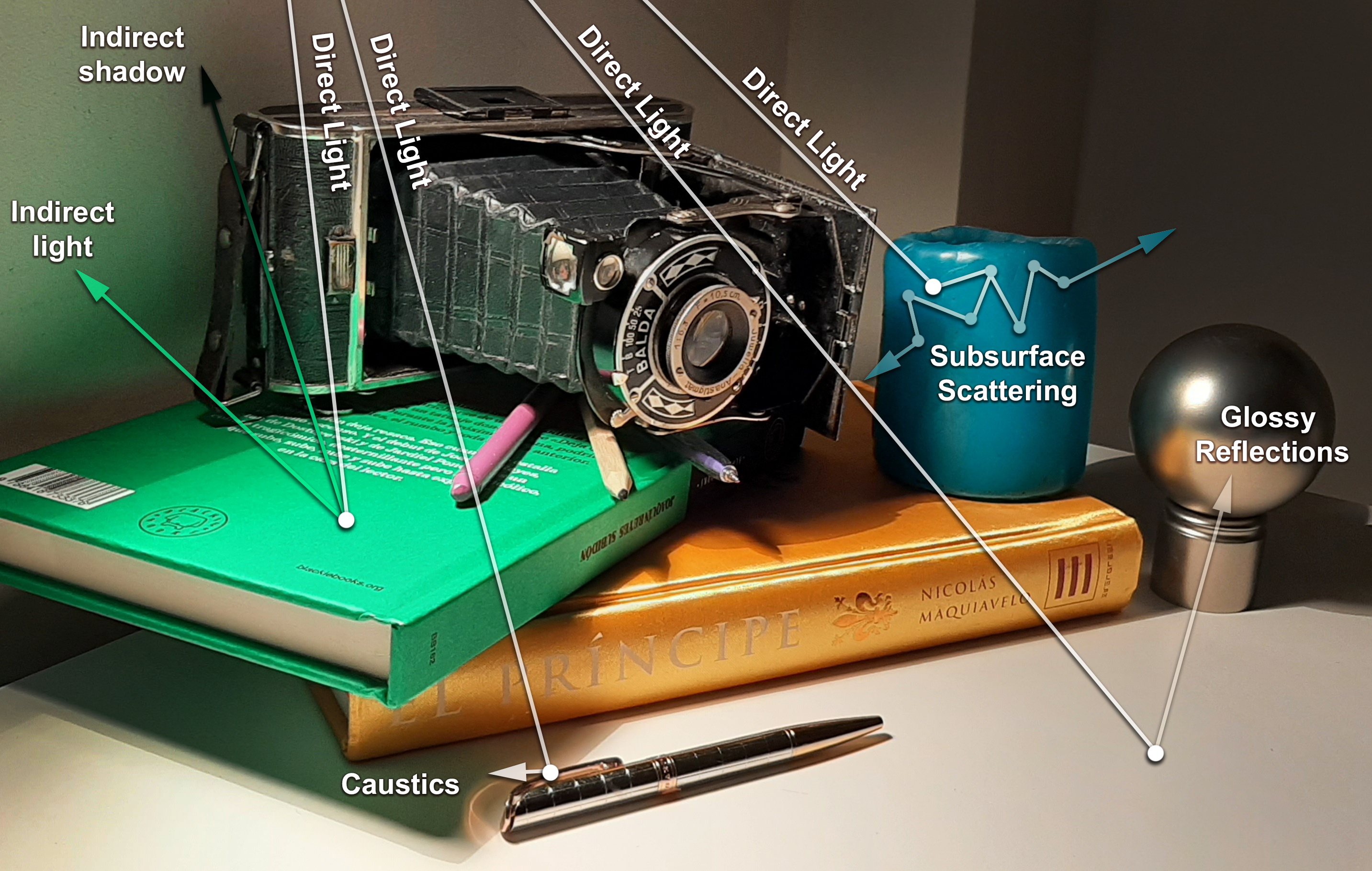}
	\caption{ Example of light transport and material interactions. Secondary bounces of light produce reflection caustics (chrome pen) and color bleeding from the green book. The wax candle exhibits multiple internal (subsurface) scattering  of photons. The yellow silk fabric of the book cover shows specular anisotropic reflections due to yarn orientation. Figure inspired by \cite{ritschel12}.}
	\label{fig:illumination_example}
\end{figure}

If we look at any simple scene surrounding us, such as the photograph in Figure \ref{fig:illumination_example}, we can find a plethora of optical interactions: indirect lighting (color bleeding), internal scattering in translucent objects, caustics, anisotropic and glossy reflections, etc. which are far from the traditional assumptions in intrinsic imaging of diffuse (lambertian) shading, direct lighting and diffuse albedo materials.

In this section we provide an overview of the theoretical background behind the image formation model, its derivation for non-diffuse materials, and the link with physically-based and inverse rendering. As we will show in following sections, only a few of these physical aspects are considered in the reviewed methods, but we think that this theory is relevant for the discussion of future lines of research.

For a deeper dive into any of the concepts quite briefly described in the following subsection, we recommend reading the book on physically based rendering by Pharr et al.\cite{Pharr16}.

\subsection{Physically-based rendering}

The color and luminosity at any point of an image, the incoming irradiance, is proportional to the sum of the outgoing radiance from all the visible points of the scene towards the camera sensor at the corresponding pixel, $I$, resulting from multiple interactions between light and matter in the scene. Naturally, this is a simplification: even if we consider the camera lenses and color filters as part of the scene, the interaction of irradiance and the sensor point affects the result, and both electronic and film cameras have specific additional image formation steps which can be simulated. In physically-based rendering, the general approach to compute this value is to use Monte Carlo estimators of the pixel and shading integral ~\cite{kajiya1986rendering}, which has the general form of:
\begin{equation}\label{eq:render_theta}
I= \int_{\chi} f(x, \Theta)\mathrm~{d}x
\end{equation}
where $f$ is a function that defines the radiance towards pixel $I$ and is defined on a domain $\chi$, generally a unit sphere, or the set of all the surfaces ($A$) in the scene, and depends on the scene parameters $\Theta$, which might include the definition of the geometry (normals, z-depth, vertices), material (albedo, BRDF), or illumination sources -- far-field environment lighting, or 3D light emitters (point, area, objects). 

The value of $I$ is defined for a given $\lambda$, which is the spectral band of the camera sensor. We could define it for as many bands as desired (including non-visible ones) but in practice, the majority of camera sensors mimics the human visual system and are commonly three-band: $\lambda \in \{R,G,B\}$. In subsequent rendering equations, we will simplify the notation by assuming a single-band, and omitting the term $\lambda$.

\begin{figure}[htb]
	\includegraphics[width=\linewidth]{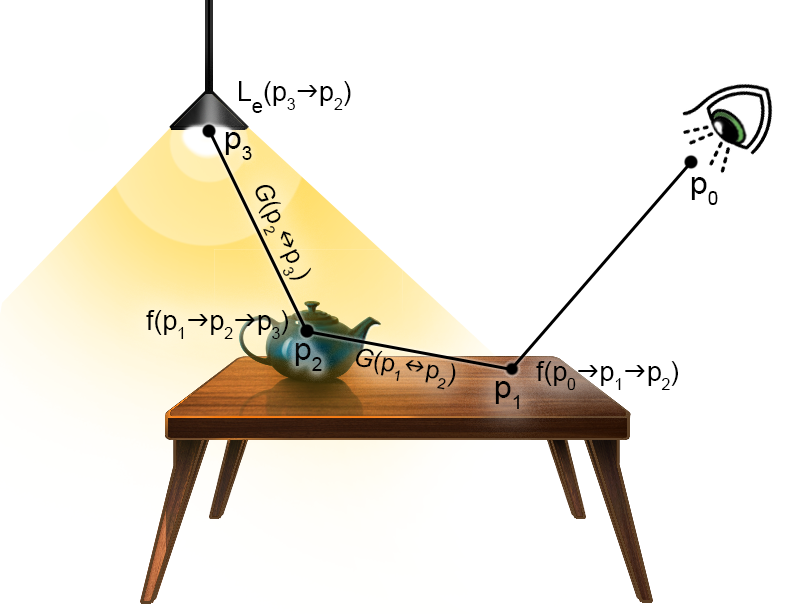}
	\caption{Example of light transport, connecting a light source $p_3$ to a pixel $I$ at $p_0$. Multiple paths like this one will need to be explored to provide a good statistical estimate of the radiance from $p_1$ to $p_0$. Figure inspired by \cite{Pharr16}.}
	\label{fig:light_transport}
\end{figure}

Equation \ref{eq:render_theta} is an integral of integrals (see Figure \ref{fig:light_transport}): to account for all the light arriving at a surface point $p_1$ to a sensor pixel at $p_0$, we have to estimate all the contributions of light from all the surfaces of the scene, recursively tracing paths bouncing in surfaces ($p_2$, $p_3$...$p_n$) until we reach the light emitted by a source $L_e(p_n \rightarrow p_{n-1})$. This is referred as the Light Transport Equation (LTE) in rendering and it is another way of seeing the equation \ref{eq:render_theta}. To compute the radiance reaching the pixel $I$, that is from $p_1$ to $p_0$ in Figure~\ref{fig:light_transport}, we would need to solve: 
\begin{equation}\label{eq:render_LTE1}
L(p_1 \rightarrow p_0) = \sum^{\infty}_{k=1} P(\bar{p}_k)
\end{equation}
with $P(\bar{p}_k)$ being the radiance scattered over a path $\bar{p}_k$ with $n+1$ vertices (\NEW{$p_0$,$p_1$},$p_2$, $p_3$...$p_n$) and computed as:
\begin{equation}\label{eq:render_LTE2}
P(\bar{p}_k)= \underbrace{\int_A\int_A ... \int_A}_{n-1} L_e(p_n \rightarrow p_{n-1}) T(\bar{p}_k)dA(p_2) ... dA(p_n)
\end{equation}
Remember that we can integrate over solid angles in the unit sphere, or surfaces (A) of the scene, being $dA(p_i)$ the differential area at point $p_i$. Note the term $T(\bar{p}_k)$, named \textit{throughput} of the path: the fraction of radiance from the light source that arrives at the camera after all of the scattering at vertices between them. The total transmitted energy will be reduced at each interaction event, as some wavelengths (colors) are absorbed or scattered away from the observer.
This is a common trade-off decision in many Monte Carlo rendering engines; longer paths per sample are costly to compute, while contributing with less and less energy with each additional vertex, but in some scenes, they might be very relevant to reduce variance (noise) and converge with less samples to an accurate image. 

Moreover, if there are interactions with non-opaque materials (human skin, cloth, marble) or participating media, such as liquids or smoke, we have to integrate volumetric scattering interactions of photons along the path between the light sources and the camera sensor pixel, requiring a more complex mathematical model such as the Radiative Transport Equation (RTE). 

\subsection{Geometry and Space} 

If we take a look at Figure~\ref{fig:light_transport}, we can observe that materials are distributed on discrete 3D objects: the table and the cup, and even the light source if we consider it as an emissive material. We can assume that one of the most important $\Theta$ parameters is geometry, usually in the form of \textbf{3D vertices and edges}, \textbf{normals} ($N$) or \textbf{depth} maps ($D$).  Please note that, in contrast to a a full 3D mesh, a single camera-view depth image (\textit{a.k.a.} Z-buffer) is an incomplete definition because the non-visible surfaces are undefined and the reflected light paths cannot be traced behind the visible objects. For example, Nimier et al.\cite{nimier2019mitsuba} require multiple views of a smoke volume in order to reconstruct its 3D density distribution by inverse rendering.

There are additional ways of defining this geometry, such as implicit surfaces, but the material distribution is more complex when there are no clear surface boundaries. That is the case of heterogeneous participating materials: human skin, airlight, mixed liquids, smoke, etc., where light transport between two points has to be in turn evaluated along the path to account for all the possible scattering effects. For instance, imagine a small cloud of vapor between points $p_1$ and $p_2$ in Figure \ref{fig:light_transport}, at each infinitesimal step along the path between those points, there will be a possibility of absorption (collision with a water particle) that will reduce the energy, but there is also a possibility of receiving incoming energy (not emitted by $p_2$), as the cloud itself is receiving direct illumination from the light source and multiple scattering events distribute the light across its volume.

In order to compute the rendering equation in volumetric media, a distribution of parameters is required at any point of the space, not only on the 3D surfaces. The usual representations include solid 3D implicit functions (for instance 3D Perlin noise, used in cloud procedural generation), meshes and distance fields, or discrete volumetric grids which store the scattering probability function at any point of the space (also known as\textit{ phase function}) by means of voxels.

\subsection{Materials}
 
Each time the light interacts with a material, there is a loss of energy and a transformation of the original wavelength reflected towards the observed direction. In rendering, the result depends on the intrinsic material response for those two angles: incident light and viewing direction (\eg: the camera, or another element of the scene). For \textit{surfaces}, this response is modeled with a Bidirectional Reflectance Distribution Function (BRDF) $f_r(\x, \omega_i,\omega_o)$ which yields at each particular 3D surface point $\x$, and for each incident direction $\omega_i$, the fraction of reflected radiance observed from a direction $\omega_o$.

The total reflected radiance $L$ at any point $\x$ can be obtained by integrating with Equation \ref{eq:render} over the positive hemisphere ${\Omega^+}$, to sample the whole incident light attenuated by the cosine term (dot product between the incident light direction $L_i$ and the normal of the surface $N$) ~\cite{kajiya1986rendering}:
\begin{equation}\label{eq:render}
L(\x, \omega_o) = \int_{\Omega^+} f_r(\x, \omega_i, \omega_o) L_i(\x, \omega_i) (\omega_i \cdot N)\mathrm~{d}\omega_i
\end{equation}

Note that by integrating the computed radiance $L(\x, \omega_o)$ of the points sampled from $p_0$ at the camera sensor ($x=p_1, \omega_o = p_1-p_0$, see Figure~\ref{fig:light_transport}), we are obtaining the irradiance at the sensor and the corresponding image pixel values ($I$ in equation \ref{eq:render_theta}). Naturally, even the pixels themselves can be sampled several times and integrated over the camera sensor with another Monte Carlo estimator to minimize aliasing effects. 

The BRDF can be extended with a Bidirectional Transmittance Distribution Function (BTDF) to conform a full Bidirectional Scattering Function (BSDF), defined in the full sphere  ${\Omega}$. The model can be further extended to account for Surface Scattering phenomena (BSSDF). 

These functions have a minimum of four dimensions (input-output pair directions in polar coordinates) and usually three RGB values as output. It is thus technically possible to choose a discrete set of  $(\omega_i, \omega_o)$ orientations an create a lookup table to interpolate the response of the material, which is captured with multiple light and camera positions (\eg with a gonioreflectometer). It is evident that the storage becomes a major drawback for tabulated data, which can only be reduced through a significant reduction of quality. Moreover, we are considering only homogeneous surface materials, which is not often the case in actual scenes (\eg, a printed paper). Those spatially-varying values (svBRDF) can be stored in a stack of textures; a Bidirectional Distribution Texture Function (BTF), increasing the dimensions and size of the table.

Beyond direct compression techniques, the most successful approach in graphics has been the use of analytic N-dimensional functions to approximate the reflectance and scattering distributions. The simplest of them, Lambertian diffuse, and Phong specular shading are also well known in the computer vision community. These functions leverage symmetry (isotropy) to model the material with a few parameters. For instance, a Lambertian material only requires to know the intrinsic albedo, a sort of base color, while the Phong model requires three additional parameters for the specular component (\eg, shininess). 
In the following subsections, we will review the most common and simple material assumptions used in recent papers, finalizing with the most sophisticated inverse material models which are starting to be studied in our field.

 \subsubsection{Lambertian Assumption}
 \label{sec:lambertian_assumption}

 The Lambertian assumption is the most common material reflectance simplification used to tackle the problem of intrinsic image decomposition.
 It consists on assuming that the BRDF of a surface is constant in all directions (diffuse) and, consequently, the observed light radiance does not depend on the viewpoint. Therefore, we can omit $\omega_o$ in the surface reflectance model $f_r$ used in Equation \ref{eq:render}.
 If the surface is diffuse, then $f_r(\omega_i) = \frac{\rho_d}{2\pi}$, with $\rho_d$ denoting the diffuse albedo: the constant ratio of incident light which is reflected in any direction, independently of the view point $\omega_o$. The image pixel value $I$ is then given by: 
 \begin{equation}\label{eq:image}
 I = \underbrace{\frac{\rho_d}{\pi}}_\text{A} \underbrace{ \int_{\Omega^+} L_i(\omega_i) (\omega_i \cdot \n)~\mathrm{d}\omega_i}_S
 \end{equation}
 The intrinsic model then can be defined as, 
 \begin{equation}
 \label{eq:intr_diff}
 I =  A\cdot S
 \end{equation}
 %
where $S$ contains all the shading variations due to the geometry of the local surface w.r.t. light direction.
In some cases, the shading image should contain contributions of all the lights in the scene ($S_1 + S_2 + ... + S_{K}$), which for discrete directional lights, can be deterministically estimated with a linear summation:
 \begin{equation}\label{eq:image_discrete}
 S = \sum_{1}^{K} L_i(\omega_i) (\omega_i \cdot \n) 
 \end{equation}
 In the case of a more realistic illumination representation, such as environment lighting, or indirect light, the shading computation requires to sample the whole hemisphere $\Omega^+$, often recursively sampling other surfaces to approximate the integral of the incoming light. The shading component $S$ within Equation \ref{eq:image} in the integral form, is difficult to compute and not so easily invertible and differentiable, so until recently, most intrinsic decomposition methods assumed the simpler formula described in Equations \ref{eq:intr_diff} and \ref{eq:image_discrete}. Note that the illumination visibility is not considered in most cases (E.g.: cast shadows).
 
 \subsubsection{Non-Lambertian Assumption}\label{sec:nonlambert}

There are two possible sources producing a Lambertian shading, either a surface which has an extremely rough micro-geometry, and thus reflects light equally in multiple random directions at any differential patch of the surface, or a very diffuse light source, coming from any direction with equal intensity (\eg, a foggy day). Both scenarios can be combined (Equation \ref{eq:render}): the shiniest object in a foggy day will look quite diffuse, while even the most diffuse materials tend to project specular reflections under focused lighting from certain view angles.
However, the majority of materials in the world are not Lambertian: even the most diffuse surface will exhibit Fresnel reflections when observed at grazing angles. Therefore, most surfaces will show the view-dependent effects that are classified as specular reflections. 
This separation between specular and Lambertian is rather pragmatic, but arbitrary, as even a simple microfacet model (shown in Figure \ref{fig:lobes}) requires multiple analytic 3D lobes to approximate the 3D reflectance response for an infinitesimal incoming light ray ($\omega_i$). The term $specular$ is usually applied to narrow lobes with high probability of scattering radiance, producing high luminance values at pixels (highlights). This family of materials are of the general form:
 \begin{align}
 L(\omega_o) = \int_{\Omega^+} \underbrace{f_r(\omega_i, \omega_o) (\omega_i \cdot \n)}_\text{$f_{NL}$} L_i(\omega_i) ~\mathrm{d}\omega_i \\
 \textbf{$f_{NL}$}(\omega_i, \omega_o, \n) = f_d (\omega_i, \n) +  f_s (\omega_i, \omega_o, \n)
 \end{align}
where $f_{NL}$ is a non-lambertian BRDF composed by two components: $f_d$, a diffuse isotropic lobe, and $f_s$, a specular lobe which depends on the camera viewpoint ($\omega_o$).

\paragraph*{Dichromatic Reflection Model. } 
 This particular Non-Lambertian model \cite{maxwell2008bi,tominaga1994dichromatic} separates the object in two reflection components ($S_d, S_s$), but considers that the specular component $S_s$ might have a color $\alpha_s$ which could differ from the color of the reflected light:
 \begin{align}
 L(\omega_o) = \alpha_d \underbrace{\int_{\Omega^+} f_d(\omega_i, \n) L_i(\omega_i) \mathrm{d}\omega_i}_\text{$S_d$} \\
 +
 \alpha_s \underbrace{\int_{\Omega^+} f_s(\omega_i, \omega_o, \n) L_i(\omega_i) \mathrm{d}\omega_i}_\text{$S_s$} 
 \end{align}
 \begin{equation}
 \label{eq:residual}
 I = \alpha_d S_d +\alpha_s S_s
 \end{equation}
 This is the case of metallic materials which, unlike dielectric ones, will show specular reflections with a change in wavelength. Some additional effects such as colored interreflections might be also captured in all layers.

\paragraph*{Phong and Blinn-Phong. } 
 The dichromatic model can be extended with one of the most adopted analytic approximations, either Phong ($f_{NL}^{P}$) or Blinn-Phong ($f_{NL}^{P}$), which could be estimated with Monte Carlo integration and arbitrary lighting, or analytically computed with directional light sources:
 \begin{align}	
 f_{NL}^{P}(\omega_i, \omega_o,  N) = \alpha_d (\omega_i \cdot \n) + \alpha_s (h \cdot \n)^k \\ 
 f_{NL}^{BP}(\omega_i, \omega_o,  N) = \alpha_s (\omega_i \cdot \n) + \alpha_s (\vect{r} \cdot \vect{v})^k   
 \end{align}
 where $\alpha_d$ and $\alpha_s$ are the colors of the diffuse and the specular reflections, the halfway vector $h = \frac{\omega_i+\omega_o}{|| \omega_i + \omega_o||}$ depends on the light direction $\omega_i$ and the view direction  $\omega_o$. The size of the specular lobe is determined by the scalar term $k \in \mathbb{R}$.

\subsubsection{Beyond Dichromatic Models: Physically-based Materials} 

Naturally, the breath of materials which can be synthesized with the previous models is very limited and not quite realistic in most cases. The advent of physically-based materials has introduced many variations \cite{SigPBR015} of the original microfacets models \cite{torrance1967}, which assume that a surface is composed of many very tiny facets that reflect light perfectly. By controlling the statistical distribution of their orientations, the roughness of the surface varies from mirror-like to almost diffuse.  Additional optical properties are introduced in these models: Fresnel view-dependent reflectivity, multiple specular lobes, metalness (conductive materials such as gold, change the color of the highlights), multiple reflection and refraction lobes, etc. 

\begin{figure}[htb]
	\includegraphics[width=\linewidth]{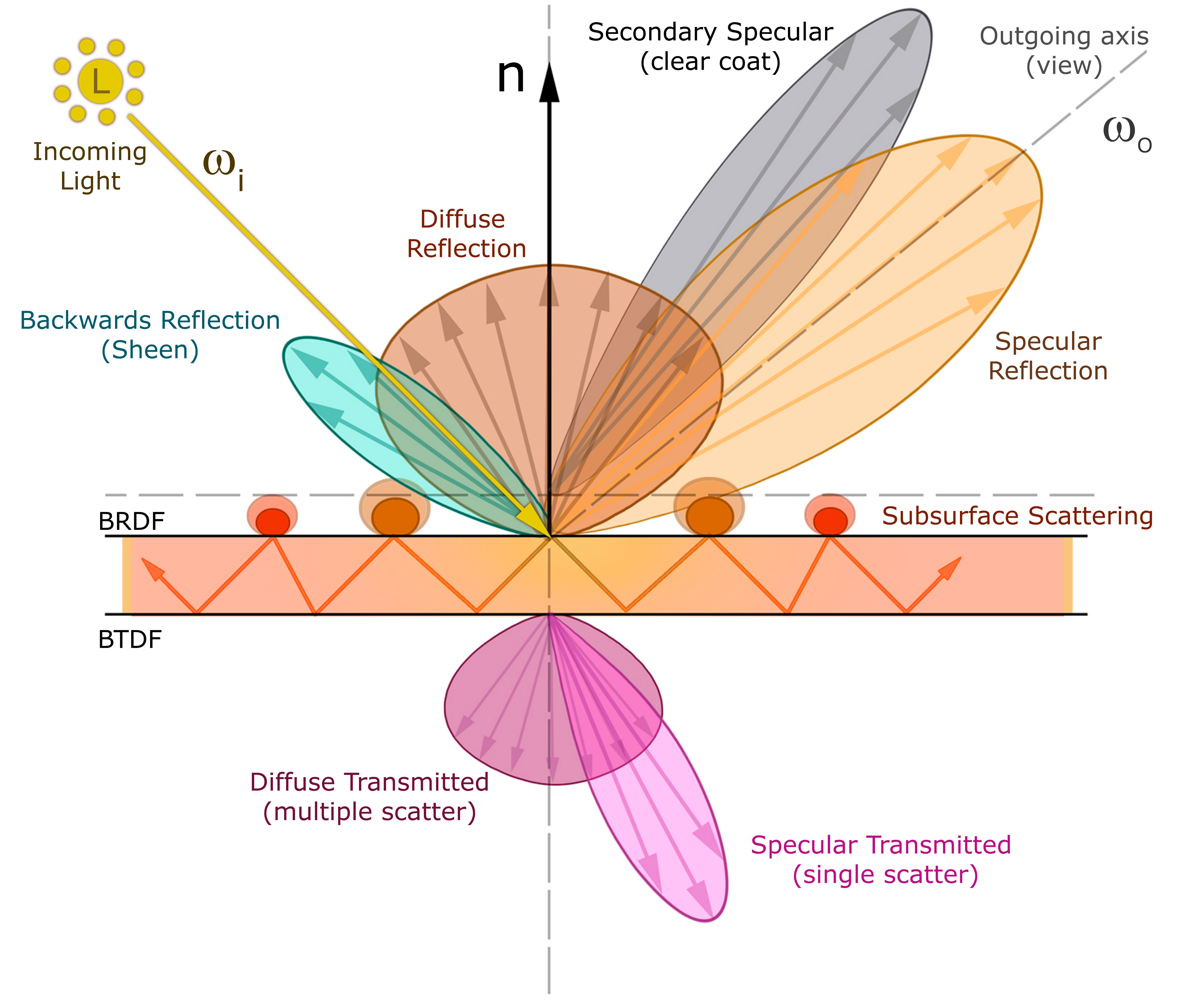}
	\caption{2D depiction of a physically-principled BSDF theorical model. In most standard representations, the continuous reflectance 4D function is discretized into a combination of analytic lobes (Cosine, GGX) which can be easily computed and used for sampling purposes. Note that subsurface scattering (photons traveling through the medium) is not composed by multiple lobes. They are depicted for descriptive purposes, but it is rather approximated by a constant value or a diffusion profile, if not explicitly computed by simulating multiple scattering events with path tracing or photon mapping.}
	\label{fig:lobes}
\end{figure}

The separation of the lobes described in the multi-lobed physically-based model shown in Figure \ref{fig:lobes} is not arbitrary, grouping reflections and refractions which share the same orientation and energy level. For instance, the main diffuse lobe is grouping multiple different orientations which are not view-dependent and share the same intensity and color. If it covers the full hemisphere with a cosine-like ratio, it is often referred as \textit{Lambertian}. Likewise, the specular transmitted lobe is grouping a view-dependent peak that the observer would only see when the translucent surface is between the light source and the camera. Even if it is often called a \textit{single scatter} lobe, likely multiple internal scattering bounces of light are also included in this group.

\subsection{Illumination}

The illumination is a significant contributor to the shading term ($S$) in most decompositions. From a rendering perspective, as shown in Equation \ref{eq:emitter}, the computation of the pixel radiance, $L$, requires to consider both the emitters $L_e$ and the irradiance: the integral of all the incoming lighting at the observed point. This incoming illumination is often neglected, considering only point light or directional analytic emitters, which simplify the shading computation by removing $L_i$ from the integral. If $f_r$ is also assumed to be Lambertian, only the form factor given by the cosine of the surface normal and the light direction remains.
\begin{equation}\label{eq:emitter}
\begin{aligned}
L(\x, \omega_o) = & L_e(\x, \omega_o)  \\
                           & + \int_{\Omega^+} f_r(\x, \omega_i, \omega_o) L_i(\x, \omega_i) \max(\omega_i \cdot \n, 0)\mathrm~{d}\omega_i\\
\end{aligned}
\end{equation}
In actual scenes, the incoming lighting is a combination of emitted or reflected illumination from distant objects (far field) and local surfaces close to the observed area (near field). The former is usually approximated in computer graphics with environment lighting, often based in High-Dynamic-Range (HDR) images mapped into an infinite sphere or cube surrounding the scene, while the latter can be derived from the far field illumination, by simulating the local secondary light bounces. If the geometry does not change, an environment map can be stored at multiple scene locations and distances, to include near field effects more accurately (Spatially Varying Environment Maps), although at a great memory cost, and only for static scenes.  

To reduce the sampling and size of environment maps and simplify the computation, Ramamoorthi and Hanrahan \cite{Ramamoorthi01} proposed their compression with Spherical Harmonics (SH), a set of orthonormal basis functions defined on the spherical domain (elevation $\theta$ and azimuth $\phi$ angles). Thus the equation \ref{eq:SH1} describes the irradiance $E$ as the sum of bases weighted by the cosine decay term $A_l$ and the illumination coefficient $L_{l,m}$. By changing the representation of the bases $\hat{Y}$ to polynomial coordinates of a unit normal $n=(x,y,z)^T$, this becomes an efficient vector dot product operation (Equation \ref{eq:SH2}). With the required modifications, this strategy is feasible with other orthogonal basis functions on the sphere.
\begin{align}\label{eq:SH1}
E(\theta, \phi) = & \sum_{l,m} \hat{A}_l L_{l,m} Y_{l,m}(\theta, \phi) \\
E= & \hat{Y}^TL \label{eq:SH2}\\
E= & T^TL  \label{eq:SH3}
\end{align}
If we want to account for near field occlusion and interreflection, it is possible to precompute those local interactions, because they depend on the object geometry and materials, and not on far field illumination. This family of techniques is known as \textit{precomputed radiance transfer} (PRT) \cite{Sloan02}: they precompute multiple events of light transport (see Figure \ref{fig:light_transport}) into the $T$ term in Equation \ref{eq:SH3} with Monte Carlo pathtracing. In this fashion, each pixel will have secondary light bounces stored in a light transport map. If only the visibility term $V(\omega_i)$ is considered for $T$, the method will be storing the ambient occlusion shadows, but not colored interreflections. 

\begin{figure}[htb]
	\includegraphics[width=\linewidth]{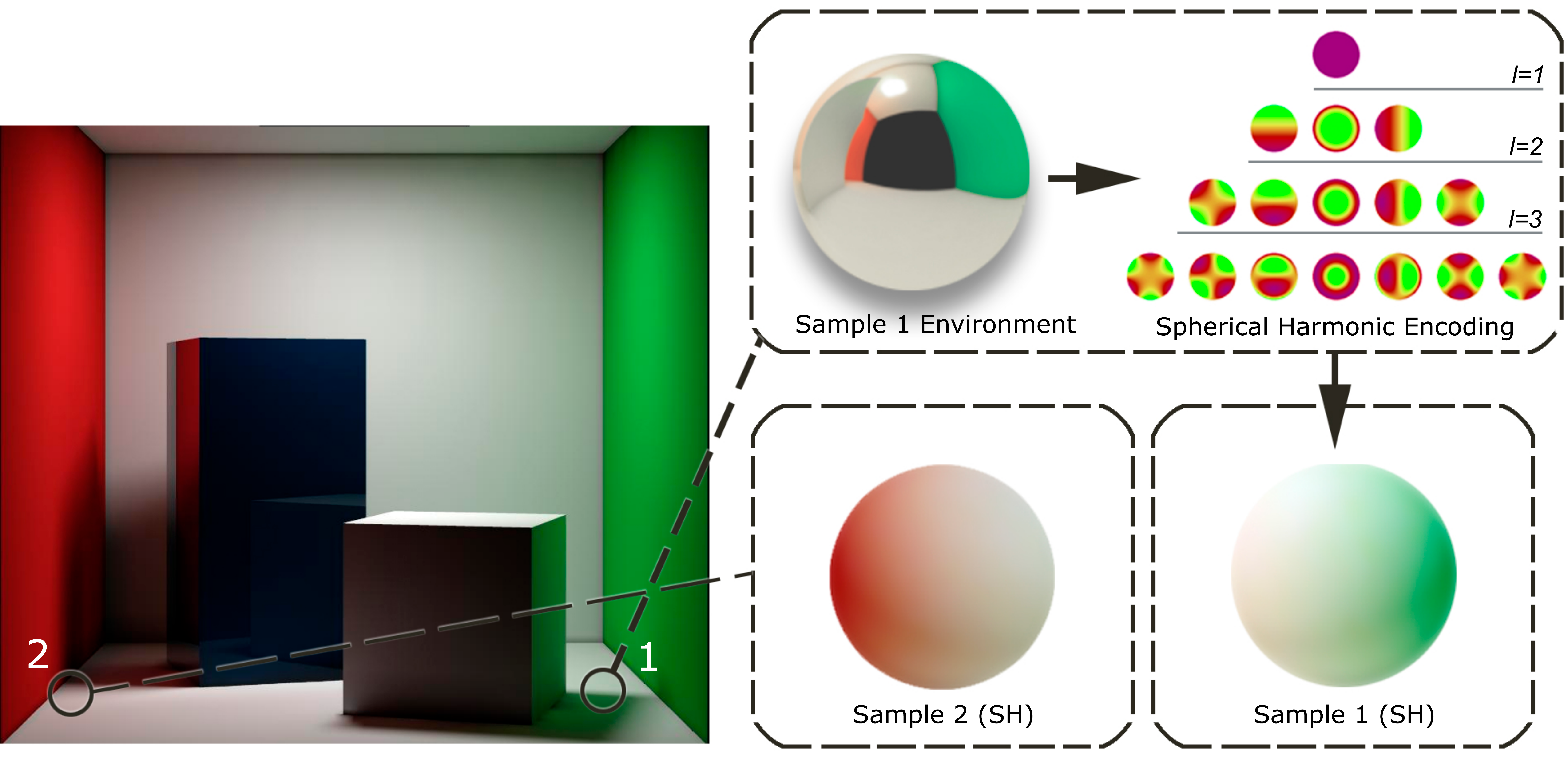}
	\caption{Example of spatially varying illumination encoding with Spherical Harmonics (SH). The incoming lighting can be computed globally for the whole scene (far field), or locally, at multiple points (near field) as we show for the two samples near each colored wall. If we project the irradiance (top row) into an SH basis we obtain a diffuse low-frequency representation (examples in bottom row).}
	\label{fig:SH}
\end{figure}

In Figure~\ref{fig:SH}, we can see a pyramid of spherical harmonics bases $Y_{l,m}$ with different coefficients. It is important to know that, although usually nine bases are considered enough to account for 99\% of the far-field irradiance at diffuse surfaces, this percentage is significantly smaller in glossy surfaces (requiring many more coefficients). Moreover, a small number of coefficients will never account for high frequency effects, such as cast shadows from high-frequency light sources (E.g: a point light representing the sun), even producing ringing artifacts if we try to increase the accuracy by adding more bases.

There are other popular basis in rendering such as Haar Wavelets or Spherical Gaussians (SG) \cite{Wang09}, which also have very interesting properties. Most of the methods analyzed in this survey use the spherical harmonics encoding (more details in Section~\ref{sec:inv}).

\section{Single Image Inverse Appearance Reconstruction}\label{sec:inv}

Given a single image, the general goal of inverse appearance reconstruction is to obtain a set of parameters, that, for a function (known or not), produce the same original image.

It can be argued that even a simple image auto-encoder performs appearance reconstruction, just by learning projections (functions) to and from deep latent space variables (parameters). However, one of the most desirable properties for those parameters in computer graphics is \textit{editability} (\eg, change the color of a wall, dim the illumination, remove a specular highlight), and such neural parameters would, in most cases, lack any meaningful semantics or intuitive control over specific components of the image formation. 
This is a well-known limitation in the field of \textit{neural rendering}~\cite{tewari2020starneural}, where the generality and editability required to create novel images with a neural network is very constrained because all the light, geometry, and material interaction events described in Section~\ref{sec:theory} are learned and embedded in an implicit unknown function. 
These models are often trained with specific parameter variations
(\eg, face relighting with environment illumination images, view synthesis with multiple lightfield views) and thus, any novel output image is limited to the parameter space sampling considered in the original training set.

The traditional approach to decompose an image into editable components is to mimic the optical process of image formation. However, as can be inferred from Equation~\ref{eq:render_LTE2}, this problem is generally ill-posed because the number of physical parameters to infer is substantially bigger than the number of known values in the system,
\NEW{resulting in several ambiguities. For example, the color of a pixel might be caused by the color of the light source or the color --reflectance-- of the material. Most of the previous work assumes gray-scale lighting, while only a minority assume colored lighting (\cite{bousseau2009user,narihira2015direct,barron2014shape}). Other ambiguities derive from the non-orthogonal parameter space, with multiple combinations yielding the same pixel value. This ambiguity, sometimes referred as the scale ambiguity, has been addressed by previous work~\cite{narihira2015direct} in the loss function, using a Scale Invariant L2 loss instead of the regular L2 (further details in Section~\ref{sec:evaluation}), or imposing priors on the albedo layer by means of bilateral filtering or L1 losses.
}

The complexity of the inverse reconstruction problem will be thus determined by the rendering function used to model the scene. As such, given a single RGB image as input, the tendency has been to simplify the rendering model, reducing it to the components with the most significant contribution to the final appearance (\eg, choosing direct over indirect illumination, diffuse Lambertian over complex BSDF models, or surface-level geometry instead of micro-geometry). 
This has ultimately lead to the \textit{intrinsic} image formation models as described in Section~\ref{ssec:img_formation_models}.
Any optical phenomena at the target image which is not reproducible by those models are thus dismissed, and either accumulated as errors in the wrong intrinsic parameter layer or stored as a residual image, trading simplicity in the parameter estimation for precision in the reconstruction. 
For example, the \textit{intrinsic diffuse model}, with albedo and shading as the only unknown parameters, is unable to capture effects like specular highlights or inter-reflections. Multiple nuanced errors are accumulated by such a simple assumption: if applied to the decomposition of a translucent object, it will generate smoother normals than the actual surface, due to the natural blurring of gradients produced by subsurface photon scattering~\cite{dong2014scattering}.
To capture those effects, along with the possibility to modify the geometry or lighting of the scene in post-processing, a complex image formulation with lights, materials, normals, and scene depth as controllable parameters is required. 

\subsection{Intrinsic Image Formation Models}
\label{ssec:img_formation_models}
\begin{figure*}[htb]
	\includegraphics[width=1\linewidth]{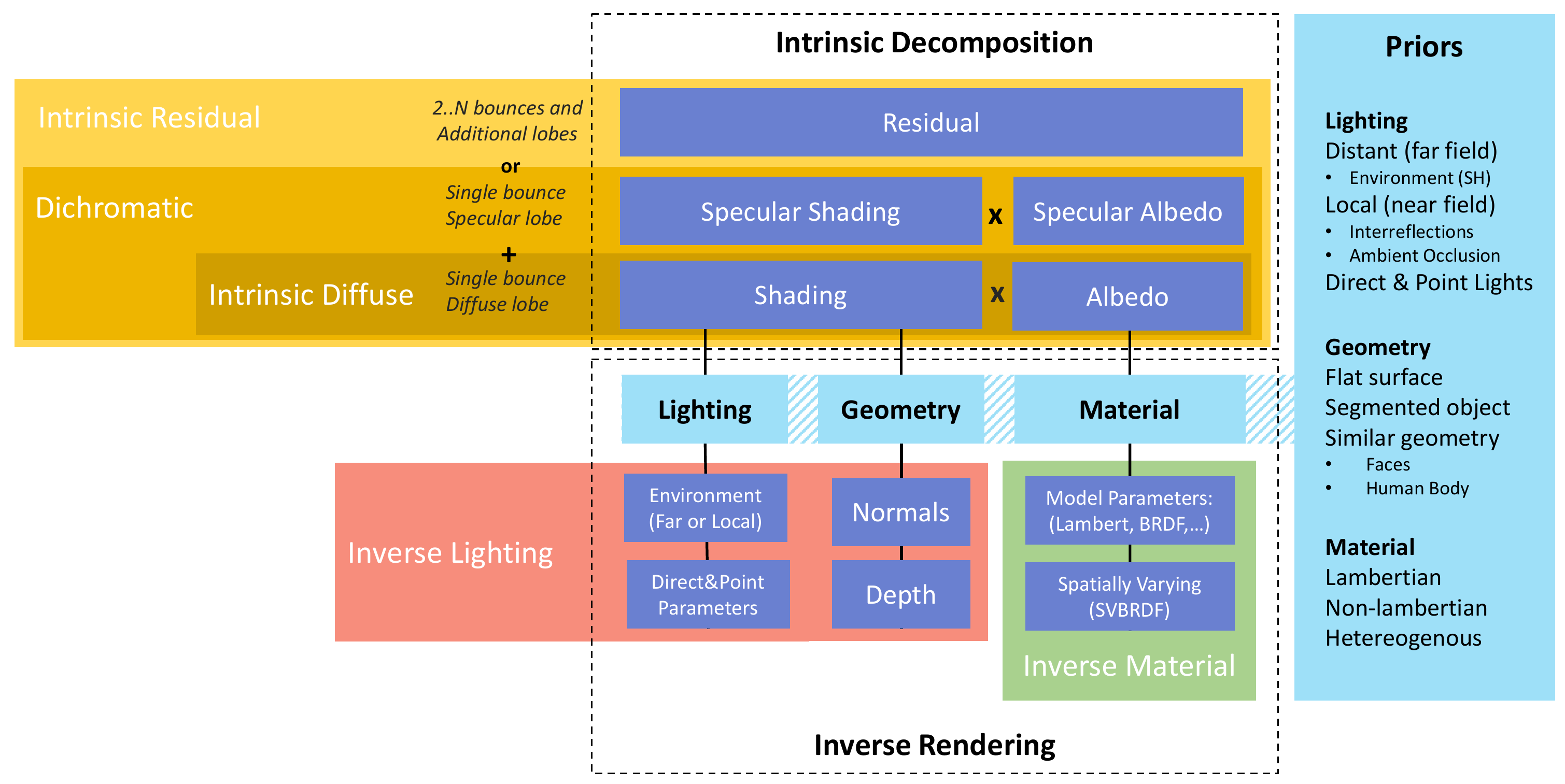}
	\caption{Taxonomy of single image inverse appearance reconstruction methods: intrinsic decomposition and inverse rendering. The intrinsic diffuse model assumes Lambertian materials and jointly couples light and geometry interaction through a shading image. The intrinsic residual captures a residual image of the sum of the other non-diffuse effects. In between, several methods in the literature have used the dichromatic reflection model, where only a colored specular reflection is taken into account. Inverse rendering methods aim at recovering the full parameterized scene (camera, lighting, geometry, and material) to synthesize novel views or to enable predictably edits.
    To simplify the problem, many methods impose priors over the scene elements, \eg: distant lighting through spherical harmonics illumination, or known proxy geometries (faces, human bodies, flat surfaces, etc.). Note that the boundary between intrinsic decomposition and inverse rendering is really fuzzy, as recent intrinsic methods are aided by implicit modeling of some scene elements (lighting through environment maps, or geometry through normals). A key difference between both approaches is that pure inverse rendering methods have the goal of modifying all the scene parameters, while intrinsic decomposition targets a more physically correct estimation of the albedo layer. 
}
	\label{fig:intrinsic_inv_render_diagram}
\end{figure*}

From a practical standpoint, most of the methods that tackle the problem of inverse appearance reconstruction for generic scenes or objects can be classified into several categories, according to the image formation model they assume (see Figure \ref{fig:intrinsic_inv_render_diagram}). All the methods reviewed in this survey are classified according to these models in Table~\ref{tab:supervision}.

\paragraph*{Intrinsic Diffuse,}~I$_d(I|A, S)$.
The input image is parameterized by the albedo ($A$) and the shading ($S$) images according to Equation \ref{eq:intr_diff}. This model assumes the materials of the objects in the scene are purely Lambertian. The majority of the methods shown in Table~\ref{tab:supervision} chose this model due to several reasons: reduced number of unknown parameters, existing priors derived from classical approaches, existing labeled datasets, and the assumption that the isotropic The Lambertian model is a good approximation to the dominant reflectance in everyday scenes 
\cite{narihira2015direct,zhou2015learning,zoran2015learning,kovacs2017shading,nestmeyer2017reflectance,janner2017self,baslamisli2018cnn,cheng2018intrinsic,fan2018revisiting,li2018cgintrinsics,li2018watching,ma2018single,bi2018deep,lettry2018darn,yu2019inverserendernet}.

\paragraph*{Intrinsic Residual,}~I$_r(I| A, S, R)$.
An image can be decomposed into an additive combination of the multiple bounces of light. The first bounce of light is directly reflected from the surface before reaching the camera sensor (direct illumination), the subsequent bounces of light hit or penetrate other surfaces a variable number of times producing global illumination effects like color bleeding, subsurface scattering, or caustics (see Figure \ref{fig:illumination_example}). 
\NEW{In the literature of intrinsic image decomposition there is a body of work focused on isolating global illumination effects, generally simplified to diffuse interreflections on lambertian surfaces. Early methods altered the illumination of the scene in order to estimate the light transport between pixels. For instance, Seitz et al.~\cite{Seitz2005} build upon shape-from-interreflection methods, computing the forward propagation of light (N bounces) and finding the cancellation operator which removes the effects of each interreflection, from a set of images in which an individual scene point is illuminated by a narrow beam of light. Similarly, Bo et al.~\cite{Bo2015} used a projector to illuminate areas of the scene and capture the multiple interactions of radiance from the target area with other surfaces, requiring a number of captures linearly dependent on the amount of recursive light bounces being estimated.}
\NEW{
The smooth nature of interreflections and user-based constraints are leveraged by Carroll et al.~\cite{carroll2011illumination} with compelling results, albeit quite sensitive to the correct placement of the strokes that guide their iterative weighted least-squares optimization process. If we consider video sequences, initial clustering can be propagated to additional frames leveraging temporal information, in such fashion, Ye et al.~\cite{Ye2014} used a Bayesian Maximum a Posteriori formulation. Similarly, Meka et al.~\cite{Meka2016} relied on iterative reweighted least squares to achieve real time decomposition. However, only direct illumination is assumed, assigning interreflections to shading variations. Finally, many methods rely on multiple views of the same scene, with different viewpoints or varying illumination conditions. In this line, Duchene et al. \cite{duchene2015} obtained separated illumination layers from outdoor photographs accounting for secondary light bounces from close surfaces, sun illumination with cast shadows, and indirect light from the sky, by propagating values in image space supported by an approximate 3D reconstruction from multiple camera views of the same scene.}
\NEW{
All these methods require information of the same scene from multiple sources, whether it is additional illumination, a novel point of view, or pixel annotation by user intervention. For instance, in the case of mirror-like reflections, the traditional approaches have estimated the shape of the reflective surface either by introducing coded lighting with projectors \cite{Balzer2011}, or capturing from multiple viewpoints, as shown by Godard et al.~\cite{Godard2015}, even accounting for self-interreflections.  It is only recently, with the development of differentiable rendering algorithms capable of physically-based inverse global illumination, such as Mitsuba \cite{nimier2019mitsuba}, that interreflections have started to become tractable in single image decomposition.}

In addition to albedo and shading parameters, the intrinsic residual model, as defined by Equation~\ref{eq:residual}, introduces an extra term ($R$) to account for all the remaining optical effects, both due to multiple interactions of light \textit{e.g.}, ambient occlusion, color bleeding from inter-reflections, caustics, and to account for additional material reflectance components \textit{e.g.}, specular reflections, translucency, scattering effects, etc.\cite{shi2017learning,meka2018lime,sengupta2019neural,zhou2019glosh,li2020inverse}.
While none of the deep learning-based methods use it, the dichromatic reflection model (Section~\ref{sec:nonlambert}) was also used to account for metallic objects and colored speculars in traditional approaches~\cite{tominaga1994dichromatic,maxwell2008bi,beigpour2011object}.

\paragraph*{Inverse Lighting,}~I$_l(S|N, L, D)$. 
A more expressive set of methods further parameterize the shading as the result of the interaction between the normal map ($N$), the illumination ($L$), or a depth map ($D$). This implies an extra level of complexity, as the amount of unknowns in the system increases. Some methods assume distant lighting (far-field) and thus use an approximate reconstruction of the environment map (E)~\cite{sengupta2019neural}, while others use spherical harmonics (SH) with a fixed number of coefficients \cite{yu2019inverserendernet} or assume directional lighting (dirL)~\cite{janner2017self} parameterized by its 3D position and intensity. Unlike other methods, which evaluate the render equation during training, Janner et al.~\cite{janner2017self} also learn the \textit{render shader} within the deep network architecture, akin to \textit{neural rendering}. Most of these methods reconstruct the surface normal $N$, as it is a required component to compute the shading. Few approaches have started to consider near-field illumination effects, such as ambient occlusion \cite{kanamori2018relighting} or local incoming illumination \cite{zhou2019glosh,li2020inverse}, as per-pixel environment maps encoded with Spherical Harmonics or Spherical Gaussians (svSH, svSG).

Most approaches combine multiple formation models, often with coupled estimation of geometry and light sources. For instance, Sengupta~\etal~\cite{sengupta2019neural} train one network to predict an intermediate representation of normals, environment maps, and albedo map, using a closed-form Lambertian shader for far-field lighting. This method also relies on the intrinsic residual model: in a second self-supervised step, they train a residual network that takes as input both the normals and environment map along with the predicted albedo, and estimates the remaining residual illumination effects which were not captured by the first pass (e.g. inter-reflections, cast shadows, near-field illumination).

\paragraph*{Inverse Material.}~
Beyond the Lambertian material model, there are a few methods which introduce more complex materials. Meka \etal~\cite{meka2018lime} use Blinn-Phong (Section~\ref{sec:nonlambert}) and regress the shininess coefficient (s-BP). Sengupta~\etal~\cite{sengupta2019neural} train the network with a dataset that contains glossy objects rendered using Phong. In this case, the use of non-Lambertian materials only serves to reinforce the estimation of the intrinsic residual term, as the network is not designed to explicitly estimate Phong material parameters. The more complete method so far is the work of Li~\etal~\cite{li2020inverse}, which assume a physically-based microfacet material model~\cite{karis2013real,SigPBR015} and predicts the albedo and roughness (r$_{SVBRDF}$) parameters, besides illumination, depth, and normals. 
Also related to this problem are the methods that estimate a coupled representation of material and illumination using reflectance maps~\cite{horn1979calculating,rematas2016deep}.

\subsection{Semantic Priors} 
\label{sec:other_domains}

Some works have focused on specific scenarios \NEW{to enforce semantic priors} and, therefore, simplify the challenging problem of inverse appearance reconstruction. 
This simplification allows to leverage known parametric models to represent surfaces, deformation, and appearances, which significantly reduce the complexity of the problem.
\NEW{Notice that such scenario-specific methods are built on top of the image formation models described in Section \ref{ssec:img_formation_models}.}
In this section we discuss several domains where learning-based methods have been proposed (\textit{e.g.}, flat materials and single objects, faces, and humans). \NEW{We explicitly link each of the methods discussed below with its corresponding underlying image formation model from Section \ref{ssec:img_formation_models}, and provide insights about the benefits of using semantic priors.}

\subsubsection{Flat Materials and Single Objects}
\label{sec:flat_and_single}

Estimating complex material parameters can be done much more easily by targeting single planar materials~\cite{dong2019deep,li2017modeling,li2018materials,vidaurre_wacv2019} or isolated objects~\cite{li2018learning}. Our survey is mainly focused on arbitrary scenes for which it is not possible to make such geometric assumptions. Nevertheless, for consistency with the literature, we overview the methods which evaluate their performance on the MIT Intrinsic dataset~\cite{grosse2009ground}, or use such dataset for training (Section~\ref{sec:objects}). There are two exceptions: the work of Meka~\etal~\cite{meka2018lime}, which we review to connect the diffuse intrinsic decomposition model with more complex svBRDF material models; and the work of Janner~\etal~\cite{janner2017self} as serves to link the problem with neural rendering methods. 

Several existing methods estimate a microfacet svBRDF model from one or several images of the material captured with a mobile device.
Deschaintre~\etal~\cite{deschaintre2018single,deschaintre2019flexible,DDB20} present a framework based on deep neural networks (UNets) trained using self-supervision and render losses. Gao~\etal~\cite{gao2019deep} uses a similar framework augmenting the training data using rendered views of the material. Recently, Guo~\etal~\cite{Guo:2020:MaterialGAN} demonstrate that GANs, StyleGAN2~\cite{karras2020analyzing}) in particular, can be powerful frameworks for estimating the reflectance properties enabling material editions using a learned latent space.

\subsubsection{Faces}
\label{sec:faces}
Many methods leverage the seminal work of Blanz and Vetter~\shortcite{blanz1999morphable} on modeling 3D faces with a low-dimensional 3D morphable model (3DMM) to incorporate geometry priors to solve the intrinsic decomposition problem.
Importantly, assuming a tight cropped image of the face, most of existing works in the area of intrinsic faces are able to train their model directly from unlabeled images in-the-wild by computing the pixel-wise difference of input and predicted image.

In the context of the different simplifications of the image formation model described at Section \ref{sec:inv}, initial methods on faces focus on the 
\textit{intrinsic diffuse} simplification (\textit{i.e.}, estimate albedo $A$ and shading $S$) \cite{shu2017neural,tewari17MoFA}, while more sophisticated approaches predict the components of the \textit{intrinsic residual} model (\textit{e.g.}, also predict specular or noise)~\cite{yamaguchi2018high}.
In order to estimate the shading layer, most methods also formulate an \textit{inverse lighting} problem and estimate normals and lighting~\cite{tewari17MoFA}.

Shu \textit{et al.}~\shortcite{shu2017neural} 
learn a subspace capable of representing face image with explicit disentanglement of normal, shading, and albedo components.
This enables
seamless edits to face images, including manipulation of expression, and adding glasses or beard.
Even if it is a self-supervised method, Shu \textit{et al.} require  intermediate constraints to prevent the network to converge to naive solutions such as shading to be constant and albedo capturing all the appearance.
To this end, they introduce a weak supervision strategy based on enforcing the estimated normals to be closed to those extracted from a 3DMM.
Tewari \etal~\cite{tewari17MoFA} also use 3DMM to reconstruct faces from monocular images by a model fitting approach. They propose a carefully designed subspace
with a latent parameters that match to a semantic encoding of facial expression, shape, illumination and albedo.
Illumination is modeled using Spherical Harmonics~\cite{muller2006spherical} with nine coefficients which tends to produce over-smooth results.

SfSNet~\cite{sengupta2018sfsnet} proposes an architecture 
to learn to separate albedo and normal layers. 
Their key observation is that in 
previous
networks \cite{shi2017learning} all high-frequency details are passed through the 
skip-connections. Therefore, the latent representation is unable to figure out whether fine details such as wrinkles or beards are due to shading or albedo. 
Consequently, they propose a new architecture that learns to separate both low and high frequency details into normal and albedo to obtain a meaningful subspace.
This is used along with the original image to predict lighting represented with spherical harmonics. This model reconstructs more detailed shape and reflectance than MoFA~\cite{tewari17MoFA} because it is not limited by the 3DMM prior.
Similarly, in their subsequent work, Tewari \textit{et al.}~\shortcite{tewari2018self} also propose a method that is not bounded by the underlying 3DMM prior. They propose and end-to-end trainable system that uses 3DMM just as a regularizer and learns corrective space for out-of-space generalization. Despite using the same illumination model as ~\cite{tewari17MoFA}, the corrective spaces enables the estimation of geometry, reflectance and lighting of higher quality, but it still assumes a Lambertian reflectance.
Follow-up research~\cite{tewari2019fml} further improve upon the use of priors and learn from scratch an appearance and geometry model to estimate the surface, albedo, and illumination of unconstrained images with unprecedented level of detail. Key to their success is a new graph-based multi-level face representation. They use both a coarse shape deformation graph and a high-resolution surface mesh, where each vertex has a color value that encodes the facial appearance.
Despite the impressive results of these works, they are all based on the assumption of a distant and smooth illumination and purely Lambertian surface properties, which prevents the modeling of any residual component (\textit{e.g.}, specular effect), which is a fundamental part intrinsic imaging.

Yamaguchi \textit{et al.}~\shortcite{yamaguchi2018high} go one step forward and focus on learning to infer \textit{high-resolution} facial reflectance, including albedo and specular layers, and fine-scale geometry from an unconstrained image.
In contrast to other approaches~\cite{shu2017neural,tewari17MoFA,sengupta2018sfsnet,tewari2018self}, their model goes beyond the Lambertian assumption, and accounts for non-trivial lighting effects such as ambient occlusion and subsurface scattering.
To this end, they use two identical 
architectures to extract specular and albedo textures, arguing that these components capture different optical features of the skin and therefore a single network easily fails in modeling conflicting features.   
Additionally, they incorporate an image completion step that generates complete texture maps.

A different approach to learn an \textit{intrinsic residual} image formation model for faces is to circumvent the use of explicit image parameters (\textit{e.g.}, $S$, $A$, or $N$) altogether, and attempt to learn a specific input-output model for a particular task. This allows the learned model to account for non-Lambertian reflectance or go beyond Spherical Harmonics (SH) illumination.
Sun \textit{et al.}~ \cite{sun2019singleimageportrait} train a
network that takes as a input a single image of a faces and a target illumination, and directly predict the relit image.
Similarly, Zhou \textit{et al.}~\cite{zhou2019deeprelight}
propose a
deep architecture to relight single images conditioned to a target lighting expressed in SH.
Meka \textit{et al.}~\cite{meka2019deepreflectance} also propose a deep learning-based approach to learn a mapping between spherical gradient images and the one-light-at-a-time (OLAT) image from a particular direction. Even if no explicit reflectance model is imposed, the residual component is captured with a task-specific perceptual loss trained to pick up specularities
and high frequency details.
Nestmeyer \textit{et al.}~\cite{nestmeyer2020relight} propose a hybrid approach where the diffuse component is represented with an explicit model, and the residual is unconstrained and modeled with a neural network. This allows for effects that are not predictable by the BRDF, such as subsurface scattering and indirect
light.

\subsubsection{Full Human Body}
\label{sec:humans}
Most of the above-discussed methods for intrinsic decomposition of faces share a common limitation: they ignore light occlusion. While this is an acceptable assumption for non-articulated and rather convex surfaces such as faces, full human bodies often present self-shadows and self-occlusions which requires more complex illumination models. This is specially evident in exceptions such as the work of Nestmeyer et al. \cite{nestmeyer2020relight} which consider a binary mask encoding the visibility from a point light source. The shadows cast by the nose or the head on the neck increase the realism of the relighting results.
Few method exist that tackle such challenge with a learning-based strategy.
Kanamori \textit{et al.}~\cite{kanamori2018relighting} propose a method that is able to learn and encode light occlusion from masked full body images. Despite being a deep-learning based approach, their loss functions explicitly minimize the intrinsic components of the images (\textit{e.g.}, albedo, light, and transport map).

\begin{figure*}[tb]
	\includegraphics[width=1\linewidth]{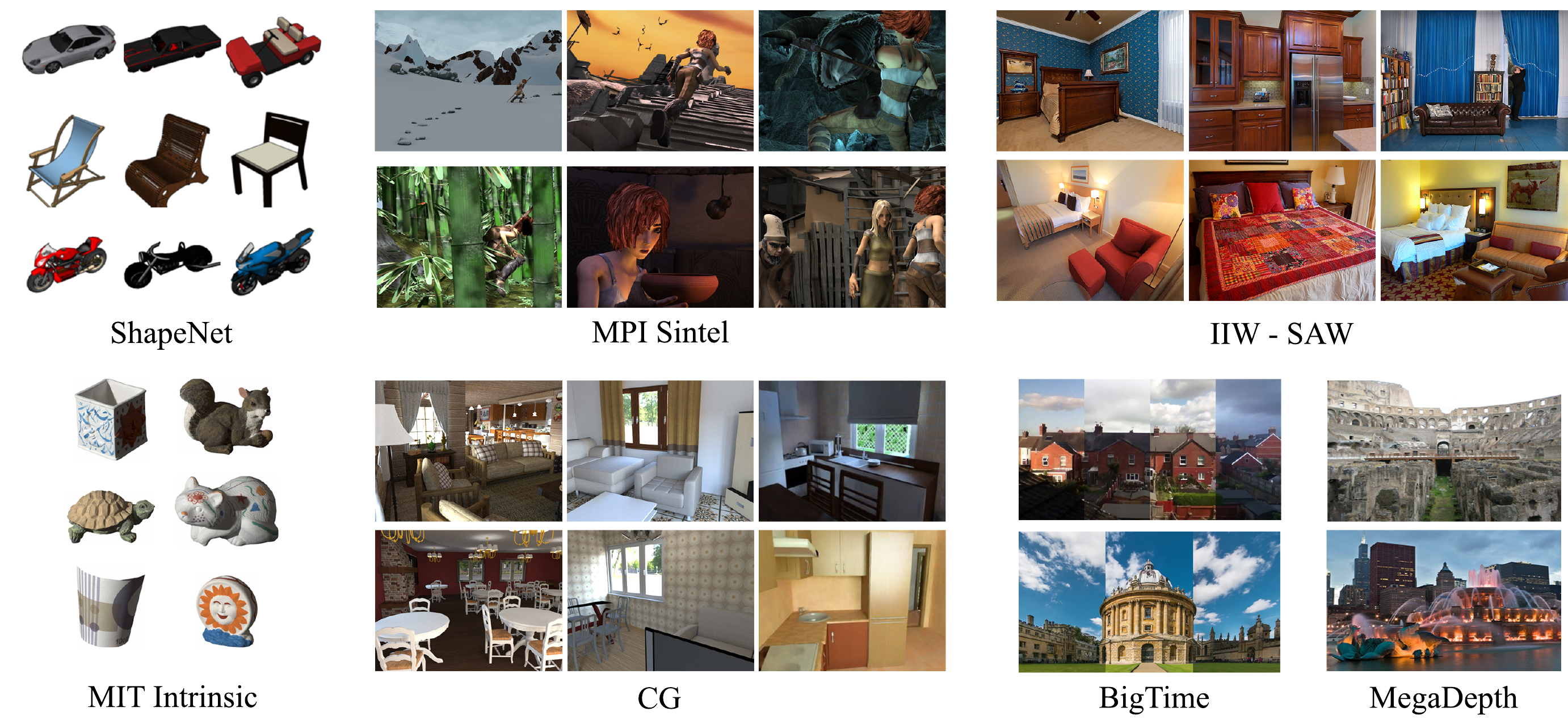}
	\caption{Representative images for the most relevant and publicly available datasets}
	\label{fig:datasets}
\end{figure*}

\begin{figure}[tb]
	\centering
	\includegraphics[width=0.9\linewidth]{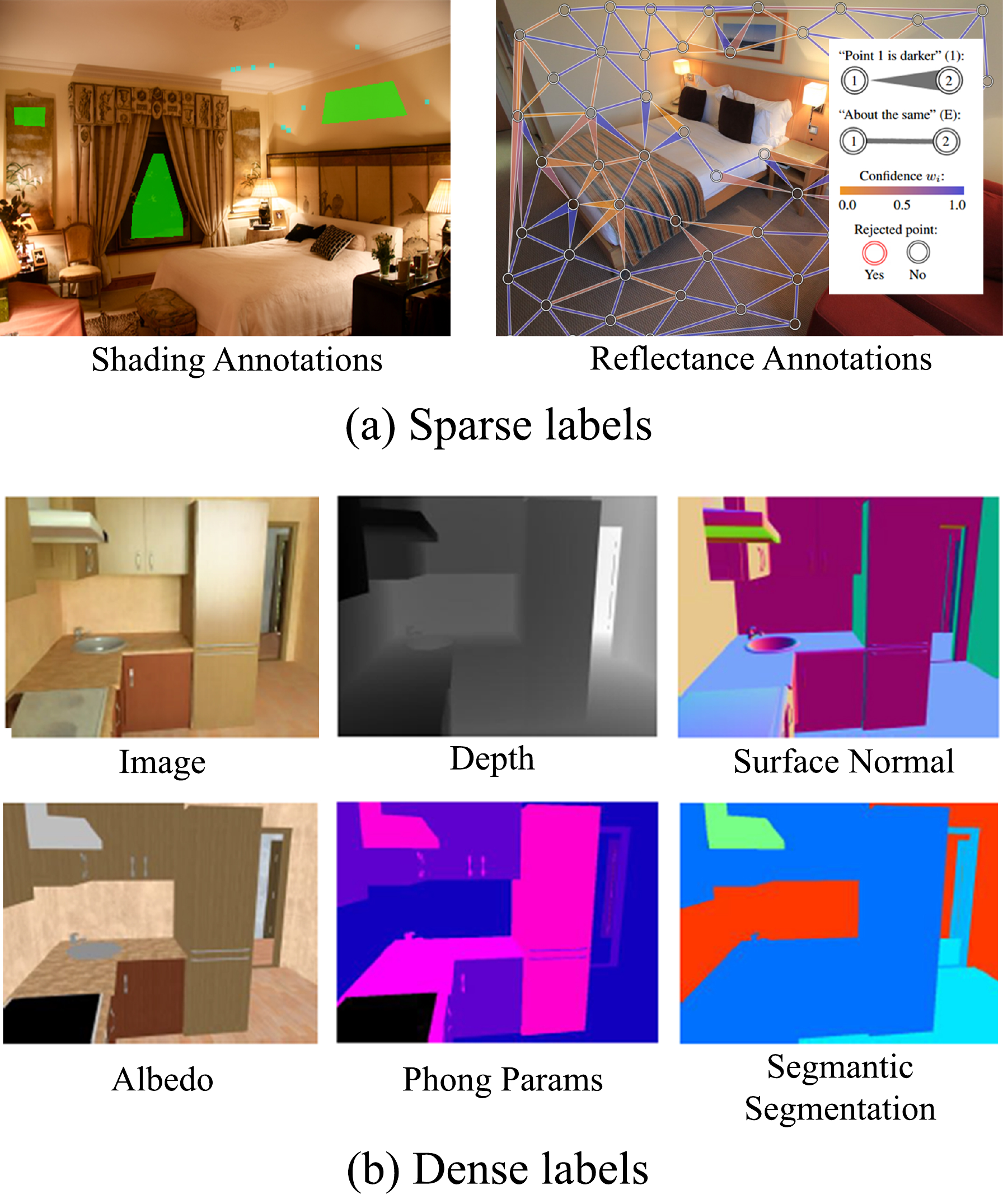}
	\caption{Difference between sparse and dense annotations. (a) Sparse labeling provided by human annotators. On the left, in Green: regions of near-constant shading but with possibly varying reflectance. In Red: edges due to discontinuities in shape (surface normal or depth). In Cyan: edges due to discontinuities in illumination (cast shadows).	Original images by \textit{herry} and \textit{uggboy} @ Flicker. 	
		(b) Dense per pixel labels provided by a synthetic dataset~\cite{sengupta2019neural}. 	
		}
\end{figure}

\section{Datasets}
\label{sec:datasets}

One of the critical pieces of any learning-based approach is the data available for training. A dataset of sufficient variety, significance, and size is required regardless of the formulation of the learning problem, e.g., supervised, semi-supervised or unsupervised.
The complexity of the intrinsic decomposition problem makes creating labeled datasets of all the individual components a highly challenging task, as this type of data cannot be freely obtained from the natural world, neither is easy to gather from human annotations. 

The very first dataset with \textit{explicit} labels for reflectance and shading was created in a laboratory setup where a few small painted figurines were coated with neutral gray to create shading images \cite{grosse2009ground}. For a long time, it was the only ground truth dataset available for quantitative evaluations, and not until recently, the boost in performance and quality of physically-based rendering engines along with the flourishing of 3D datasets has facilitated the creation of larger, more complex, and heterogeneous data for training.
The lack of labeled data motivated the use of alternative learning-based solutions. 
In this regard, the main approach has been to leverage semantic knowledge of the scene content and learn from relative measurements instead of regressing absolute radiance values. 
Relative measurements of the scene appearance can be found relatively easily, for example, humans are quite skilled at judging whether two surfaces are made of the same material despite illumination variations~\cite{bell2014intrinsic}. This property (the \textit{Albedo Invariance}), which is key to disambiguate the contribution of the intrinsic components, can also be exploited given existing and freely available datasets such as time-lapse sequences.

In this section, we review the most common datasets which are being used for training and evaluation purposes. The core datasets discussed here correspond to those explicitly published and made available to the public. Nevertheless, many methods build on these datasets to create their own without publishing it. In the following, the description of each dataset includes the methods that use them along with existing derivations. 
We organize the datasets according to the following properties:

\begin{itemize}
\item \textbf{Size (\# 3D Models, \# Sequences, \# Imgs)}: Total number of images (Imgs), sequences of the same scene with varying illumination or viewpoint (Sequences), or number of 3D renderable scenes. Note that ShapeNet~\shortcite{chang2015shapenet} and SUNCG~\cite{song2017semantic} are datasets of renderable objects/scenes so the amount of images generated to train the models depend on the particular method.
\item \textbf{Scene Content}: Images included in the dataset might be of individual objects (\textit{obj}) or complex scenes. In the latter case, some datasets might contain indoor scenes (\textit{ind}), outdoor scenes (\textit{outd}), or a combination (\textit{any}). 
\item \textbf{Scene Syn/Real}:  The image can be \textit{synthetic} or \textit{real}. In the former case, some of the datasets use physically-based rendering engines, while other have been generated with non-photorealistic ones.
   \item \textbf{Source}: The constraints used for training might be automatically generated from the data (\textit{auto}), or come from \textit{human} annotations.
	\item \textbf{Labeling}: Some datasets provide \textit{sparse} annotations of just a few pixels of the images, while others provide \textit{dense} per-pixel annotations.
	\item \textbf{Constraints}: the data can be labeled with \textit{explicit} (or \textit{absolute}) values for each of the unknown parameters, or can be used as a way to extract \textit{relative} relationships for the intrinsic components within a single image, or across several images of the same scene. If the latter, the dataset might be additionally organized in \textit{sequences}. 
\end{itemize}

The reminder of this section is organized according to the \textit{scene content}, namely objects or general scenes. The other properties will be mentioned within the description of each dataset. Please refer to Table~\ref{tab:datasets} for a comprehensive summary of datasets and their properties, and Figure~\ref{fig:datasets} for a selection of representative images.

\newcommand{\captiontabledatasets}{Summary of datasets used in the literature of deep learning-based intrinsic decompositions which are avaible to the public.  
 $^1$ The render engine used to generated the images of this dataset is a non-photorealistic one. More details about the definition of each category in the accompanying text.}

\begin{table*}[!htp]\centering
	\caption{\captiontabledatasets}\label{tab:datasets}
	\begin{tabular}{lcccccccccc}\toprule
		\textbf{Dataset} &\textbf{ \# 3D Models} &\textbf{\# Seqs} &\textbf{\# Imgs} &\textbf{\parbox{2cm}{Max Image\\ Size (approx)}} &\textbf{Scene} &\textbf{Syn/Real} &\textbf{Source} &\textbf{Labeling} &\textbf{Constraints} \\\midrule
		ShapeNet \shortcite{chang2015shapenet} &4000 &* &* &* &obj &syn &auto &dense &abs/rel \\
		MIT Intrinsics \shortcite{grosse2009ground} &- &10 &220 & 600 &obj &real &auto &dense &abs/rel \\ 
		SUNCG \shortcite{song2017semantic} &40k &* &* &* &ind &syn &auto &dense &abs \\
		MPI Sintel \shortcite{butler2012naturalistic} &- &- &890 &1024$\times$436 &any &syn$^2$ &auto &dense &abs \\
		CGIntrinsics \shortcite{li2018cgintrinsics} &- &- &20k &640$\times$480 &ind &syn &auto &dense &abs \\
		IIW \shortcite{bell2014intrinsic} &- &- &5230 &512 &ind &real &human &sparse &rel \\
		SAW \shortcite{kovacs2017shading} &- &- &6677 &512 &ind &real &human &sparse &rel \\
		BigTime \shortcite{li2018watching} &- &155 &6500 & 1080 &any &real &auto &dense &rel \\
		MegaDepth \shortcite{li2018megadepth} &* &200 &150k &1080 &outd &real &auto &dense &rel \\
		\bottomrule
	\end{tabular}
\end{table*}

\subsection{Objects}\label{sec:objects}

As a way to constrain the problem, some methods have limited the domain to isolated objects. This enables the use of additional priors about the underlying geometry and shape reducing the number of possible solutions and enabling more complex materials, and illumination models. Such is the case for methods targeted at flat surfaces, faces or humans. Although we briefly review them in Section~\ref{sec:other_domains}, in this survey we focus on the methods that deal with arbitrary object shapes or provide quantitative errors on common metrics.
Another advantage of dealing with objects instead of generic scenes is that it becomes less complex to capture/render variations (or sequences) of images of the object as seen under different perspectives or viewpoints. Such is the case for the two datasets reviewed below.

\noindent{\textbf{MIT Intrinsic~\cite{grosse2009ground}}.}  
Contains real images of 20 objects with ground truth albedo and shading captured under 11 different directional light sources, resulting in 220 images (20 sequences). Shading images were obtained by painting the object with gray spray. The objects were photographed within a controlled setup which minimized indirect illumination and allowed easy alignment between different shots.
Even though it is a small dataset, the majority of methods have used it for training or fine-tuning their models. There are two divisions of this dataset that people have consistently used to compare performance: the Barron split~\cite{barron2014shape}, which divides each sequence by image, and the Direct Intrinsics split~\cite{narihira2015direct}, which divides the sequences by objects.

\noindent{\textbf{ShapeNet ~\cite{chang2015shapenet}}.}
The ShapeNet dataset is a richly annotated dataset of 3D objects with albedo maps of over 4000 object categories. 
Different methods have rendered the available 3D models into datasets of different sizes, objects variety, and illuminations. 
They all leverage a physically-based rendering engine (e.g. Mitsuba\cite{jakob2010mitsuba} or Blender Cycles) as well as provide dense labels. Additionally, the ShapeNet 3D dataset has been used to generate sequences of images of the same object under different illumination conditions, providing relative constraints for the learning formulation. 
Meka~\etal~\shortcite{meka2018lime} render 100k images of 55 objects using Blinn-Phong materials, and 45 indoor environment maps.
Shi~\etal~\shortcite{shi2017learning} render more than 2M images of 30k objects, using Phong materials,
and 98 environment maps. They perform category-specific training using four objects (car, chair, airplane, and sofa), and evaluate cross-category generalization.
Janner~\etal~\shortcite{janner2017self} also study cross-category generalization, and use Blender Cycles to render Lambertian materials \textit{on demand} in the unsupervised setup.
Baslamisli~\etal~\shortcite{baslamisli2018cnn} render 20k images of different 3D models assigning random colors to the albedo materials to introduce more variety.
Ma~\etal~\shortcite{ma2018single} further leverage this dataset to generate multi-illuminant training sequences by randomly picking 10 different light positions per each of the 10 selected object categories (each one containing 100 objects). 
The amount of training samples obtained thanks to ShapeNet is huge, however, none of the methods published the splits so they are not available for comparing performance across different methods.

\subsection{Scenes}\label{sec:scenes}
Dealing with arbitrary types of scenes is the ultimate goal of intrinsic decomposition methods. However, the appearance of a material under different types of illumination can change dramatically (e.g. clear-sky vs cloudy day, or natural vs artificial lighting). Most of the existing datasets in this category, in particular, the ones that are synthetically generated, describe indoor scenes. 
In contrast to the ShapeNet dataset presented in the previous section, generating on-the-fly samples for arbitrary scenes is much more expensive, consequently, existing synthetic datasets in this category are static and provide absolute learning constraints (MPI Sintel, SUNCG, CGIntrinsics). 
Relative comparisons have been gathered from crowd-sourcing experiments (IIW, SAW) or from the physics of the image formation (BigTime, MegaDepth).

\noindent{\textbf{MPI Sintel}.}~This synthetic dataset~\cite{butler2012naturalistic} of animation scenes  was originally designed for optical flow evaluation, however, thanks to providing the albedo layer, it proved useful for the evaluation of intrinsic decomposition methods.
As the original renders contained complex lighting effects (specular highlights, inter-reflections, etc.), it was re-synthesized to appear purely \textit{lambertian} and coherent \cite{chen2013simple,narihira2015direct}.
It contains a total of 890 images from 18 scenes with around 50
frames each. Like MIT Intrinsic, there are two known splits: the \textit{scene split}, placing the whole scene (all frames) either completely in training or completely in testing; and the \textit{image split}, where the same scene will appear both in train and test, placing different frames in each split.
Methods have consistently used the same splits of this dataset for training and evaluating their methods.

\noindent{\textbf{SUNCG}.}~The SUNCG dataset ~\cite{song2017semantic} contains 40k manually created indoor environments with dense volumetric semantic annotations. Likewise ShapeNet, the 3D models of this dataset have served as a baseline to several methods which have rendered the scenes under different illumination conditions and materials. Zhou~\etal~\cite{zhou2019glosh} have used it to render $58949$ images of Lambertian surfaces using Mitsuba~\cite{jakob2010mitsuba} for which they also have the albedo, normal map, depth, and shading generated by setting all the materials to diffuse and reflectance to 1.
Sengupta~\etal~\shortcite{sengupta2019neural} also departs from the SUNCG dataset, enriching the existing material models using Phong~\cite{lafortune1994using}. It contains 230k images of indoor scenes physically-based rendered under multiple outdoor environment maps. It further provides the same scene rendered under both diffuse and specular settings, as well as labels about normal maps, depth, Phong model parameters, semantic and glossiness segmentation. In order to reduce ray-tracing timings, the method has used deep denoising~\cite{chaitanya2017interactive}. 

\noindent{\textbf{CGIntrinsics~\cite{li2018cgintrinsics}}.}~Taking 3D models and textures of indoor scenes from the SUNCG dataset~\cite{song2017semantic}, this dataset contains 20k images (and albedo maps) of physically-based renderings using path tracing with global illumination. The dataset also provides: the set of 50 synthetic scenes provided by Bonneel~\etal~\shortcite{bonneel2017intrinsic}, code to compute the shading image from the render and the albedo, the segmentation of each image into superpixels~\cite{achanta2012slic}, the training split, and the precomputed bilateral embedding used in the paper to guarantee shading smoothness. 

\noindent{\textbf{Intrinsic Images in the Wild (IIW)}.}~IIW dataset~\cite{bell2014intrinsic} is a sparse large-scale set of relative reflectance judgments of indoor scenes collected via crowdsourcing, containing over 900k comparisons across ~5000 photos.
Along with the dataset, the authors provide a metric for evaluating the algorithms performance, the Weighted Human Disagreement Rate (WHDR), which measures the percent of human judgments that the method predicts incorrectly, weighted by the confidence of each judgment. Although it has the limitation of the sparsity of the annotations, this dataset is being used consistently for comparing performance.

\noindent{\textbf{Shading Annotations in the Wild (SAW)}.}~Following the methodology of IIW, SAW dataset~\cite{kovacs2017shading} contains 15k sparse annotations of shading gradients over 6000 images of indoor environments: smooth shading, normal/depth discontinuity, and shadow boundary. Using Precision-Recall metrics, this dataset can also be used across methods to compare performance.

\noindent{\textbf{BigTime~\cite{li2018watching}}.}~Large dataset of real image sequences of both indoor and outdoor scenes under varying illumination. Sky, and dynamic objects such as pets, people, and cars were masked out. It contains a total of 145 sequences from indoor scenes and 50 from outdoor scenes, yielding a total of over 6500 images.

\noindent{\textbf{MegaDepth~\cite{li2018megadepth}}.}
Contains 150k images of 200 different landmarks which have been reconstructed using state-of-the-art Shape-from-Motion and Multi-View-Stereo methods. Each image is accompanied by its depth map and camera parameters, so that it is possible to reconstruct the scene from a vast range of view points.

\begin{figure*}[htb] 
	\includegraphics[width=1\linewidth]{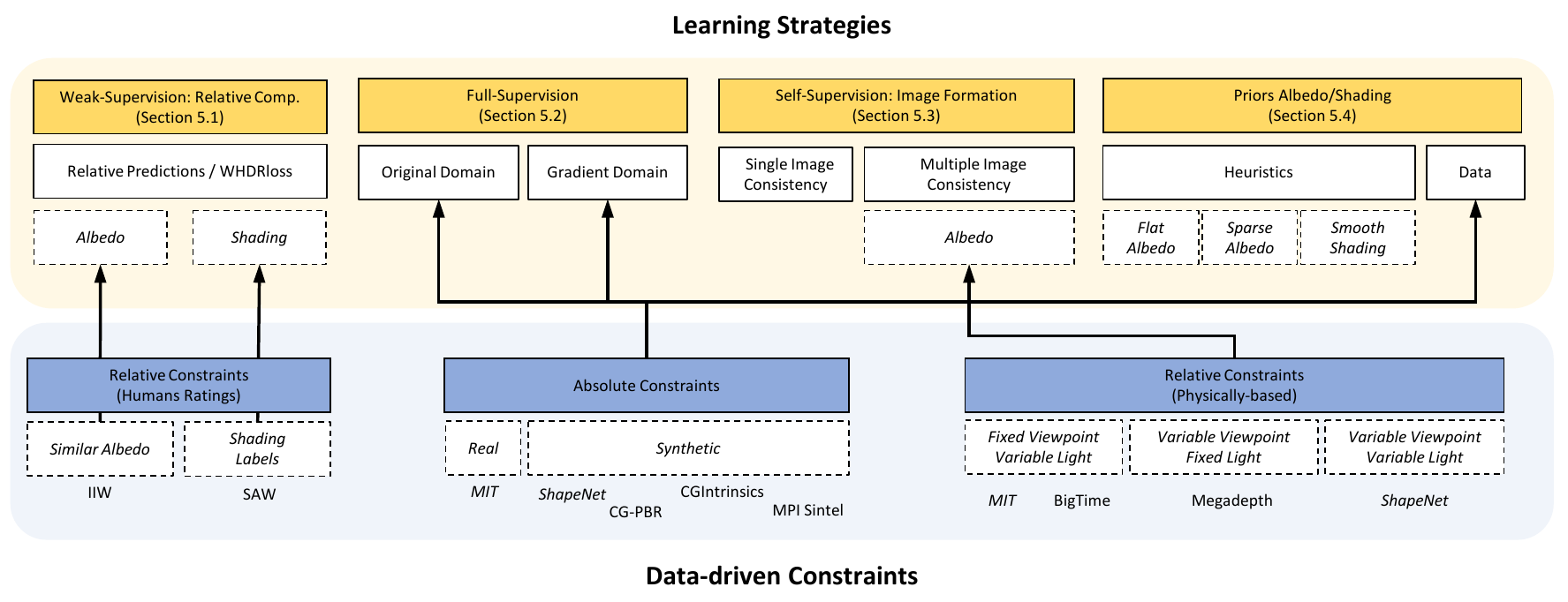}
	\caption{Learning strategies and relationship with data-driven constraints.}
	\label{fig:overview_strategies}
\end{figure*}

\begin{table*}[!htp]\centering
	\caption{Summary of methods presented in this survey. The explanation of the models is presented in Section~\ref{ssec:img_formation_models}. A full description of the learning strategy and neural architectures is presented in Sections~\ref{sec:learning_formulation} and \ref{sec:architectures}. }
	\label{tab:supervision}
	\scriptsize
\begin{tabular}{lrrrrrrrrrrrrrrrrr}\toprule
	\multirow{3}{*}{\textbf{Method}} &\multicolumn{4}{c}{\multirow{2}{*}{\textbf{Model}}} &\multicolumn{10}{c}{\textbf{Learning Strategy}} &\multicolumn{2}{c}{\multirow{2}{*}{\textbf{\parbox{1cm}{\centering Network\\Architecture}}}} \\\cmidrule{6-15}
	& & & & &\multicolumn{2}{c}{\textbf{Weak-S}} &\multicolumn{2}{c}{\textbf{Full-S}} &\multicolumn{2}{c}{\textbf{Self-S}} &\multicolumn{4}{c}{\textbf{Priors}} & & \\\cmidrule{2-17}
	&\rot{Diffuse} I$_d$ &\rot{Residual} I$_r$ &\rot{Lighting} $L$ &\rot{Material} &\rot{RP} &\rot{WHDR$_\text{loss}$} &O &$\nabla$ &SIC &MIC &\Aflat &\Asparse &\Ssmooth &Data &C &I2IT \\\midrule
	\cellcolor[HTML]{f3f3f3}\textbf{Narihira~\etal~\shortcite{narihira2015judgement}} &\cellcolor[HTML]{f3f3f3}- &\cellcolor[HTML]{f3f3f3}- &\cellcolor[HTML]{f3f3f3}- &\cellcolor[HTML]{f3f3f3}- &\cellcolor[HTML]{f3f3f3}\Ss &\cellcolor[HTML]{f3f3f3} &\cellcolor[HTML]{f3f3f3} &\cellcolor[HTML]{f3f3f3} &\cellcolor[HTML]{f3f3f3} &\cellcolor[HTML]{f3f3f3} &\cellcolor[HTML]{f3f3f3} &\cellcolor[HTML]{f3f3f3} &\cellcolor[HTML]{f3f3f3} &\cellcolor[HTML]{f3f3f3} &\cellcolor[HTML]{f3f3f3}P-W &\cellcolor[HTML]{f3f3f3} \\
	\textbf{Narihira~\etal~\shortcite{narihira2015direct}} &\cmark & & & & & &\Aa \Ss &\Aa & & & & & & & &B \\
	\cellcolor[HTML]{f3f3f3}\textbf{Zhou~\etal~\shortcite{zhou2015learning}} &\cellcolor[HTML]{f3f3f3}\cmark &\cellcolor[HTML]{f3f3f3} &\cellcolor[HTML]{f3f3f3} &\cellcolor[HTML]{f3f3f3} &\cellcolor[HTML]{f3f3f3}\Aa &\cellcolor[HTML]{f3f3f3} &\cellcolor[HTML]{f3f3f3} &\cellcolor[HTML]{f3f3f3} &\cellcolor[HTML]{f3f3f3} &\cellcolor[HTML]{f3f3f3} &\cellcolor[HTML]{f3f3f3} &\cellcolor[HTML]{f3f3f3} &\cellcolor[HTML]{f3f3f3} &\cellcolor[HTML]{f3f3f3} &\cellcolor[HTML]{f3f3f3}P-W &\cellcolor[HTML]{f3f3f3} \\
	\textbf{Zoran~\etal~\shortcite{zoran2015learning}} &\cmark & & & &\Aa & & & & & & & & & &P-W & \\
	\cellcolor[HTML]{f3f3f3}\textbf{Kovacs~\etal~\shortcite{kovacs2017shading}} &\cellcolor[HTML]{f3f3f3}\cmark &\cellcolor[HTML]{f3f3f3} &\cellcolor[HTML]{f3f3f3} &\cellcolor[HTML]{f3f3f3} &\cellcolor[HTML]{f3f3f3}\Ss &\cellcolor[HTML]{f3f3f3} &\cellcolor[HTML]{f3f3f3} &\cellcolor[HTML]{f3f3f3} &\cellcolor[HTML]{f3f3f3} &\cellcolor[HTML]{f3f3f3} &\cellcolor[HTML]{f3f3f3} &\cellcolor[HTML]{f3f3f3} &\cellcolor[HTML]{f3f3f3} &\cellcolor[HTML]{f3f3f3} &\cellcolor[HTML]{f3f3f3} &\cellcolor[HTML]{f3f3f3}B \\
	\textbf{Nestmeyer~\etal~\shortcite{nestmeyer2017reflectance}} &\cmark & & & & &\Aa & & & & &\cmark & & & & &B \\
	\cellcolor[HTML]{f3f3f3}\textbf{Shi~\etal~\shortcite{shi2017learning}} &\cellcolor[HTML]{f3f3f3} &\cellcolor[HTML]{f3f3f3}\cmark &\cellcolor[HTML]{f3f3f3} &\cellcolor[HTML]{f3f3f3} &\cellcolor[HTML]{f3f3f3} &\cellcolor[HTML]{f3f3f3} &\cellcolor[HTML]{f3f3f3}I$_r$ \Aa \Ss &\cellcolor[HTML]{f3f3f3}\Aa &\cellcolor[HTML]{f3f3f3} &\cellcolor[HTML]{f3f3f3} &\cellcolor[HTML]{f3f3f3} &\cellcolor[HTML]{f3f3f3} &\cellcolor[HTML]{f3f3f3} &\cellcolor[HTML]{f3f3f3} &\cellcolor[HTML]{f3f3f3} &\cellcolor[HTML]{f3f3f3}E-D \\
	\textbf{Janner~\etal~\shortcite{janner2017self}} &\cmark & &dirL & & & &\Aa \Nn \Ll & &\cmark & & & & & & &E-D \\
	\cellcolor[HTML]{f3f3f3}\beyLamb{\textbf{Meka~\etal~\shortcite{meka2018lime}}} &\cellcolor[HTML]{f3f3f3} &\cellcolor[HTML]{f3f3f3}\cmark &\cellcolor[HTML]{f3f3f3}E &\cellcolor[HTML]{f3f3f3}BP &\cellcolor[HTML]{f3f3f3} &\cellcolor[HTML]{f3f3f3} &\cellcolor[HTML]{f3f3f3} I$_d$ R s-BP &\cellcolor[HTML]{f3f3f3} &\cellcolor[HTML]{f3f3f3}\cmark &\cellcolor[HTML]{f3f3f3} &\cellcolor[HTML]{f3f3f3} &\cellcolor[HTML]{f3f3f3} &\cellcolor[HTML]{f3f3f3} &\cellcolor[HTML]{f3f3f3} &\cellcolor[HTML]{f3f3f3} &\cellcolor[HTML]{f3f3f3}E-D \\
	\textbf{Baslamisli~\etal~\shortcite{baslamisli2018cnn}} &\cmark & & & & & &\Aa \Ss &\Aa \Ss &\cmark & & & & & & &E-D \\
	\cellcolor[HTML]{f3f3f3}\textbf{Cheng~\etal~\shortcite{cheng2018intrinsic}} &\cellcolor[HTML]{f3f3f3}\cmark &\cellcolor[HTML]{f3f3f3} &\cellcolor[HTML]{f3f3f3} &\cellcolor[HTML]{f3f3f3} &\cellcolor[HTML]{f3f3f3} &\cellcolor[HTML]{f3f3f3} &\cellcolor[HTML]{f3f3f3}\Aa \Ss &\cellcolor[HTML]{f3f3f3} &\cellcolor[HTML]{f3f3f3}\cmark &\cellcolor[HTML]{f3f3f3} &\cellcolor[HTML]{f3f3f3}\cmark &\cellcolor[HTML]{f3f3f3}\cmark &\cellcolor[HTML]{f3f3f3}\cmark &\cellcolor[HTML]{f3f3f3} &\cellcolor[HTML]{f3f3f3} &\cellcolor[HTML]{f3f3f3}Res \\
	\textbf{Fan~\etal~\shortcite{fan2018revisiting}} &\cmark & & & & &\Aa &\Aa \Ss &\Aa \Ss $P_{\nabla A}$ & & &\cmark & & & & &Res \\
	\cellcolor[HTML]{f3f3f3}\textbf{Li~\etal~\shortcite{li2018watching}} &\cellcolor[HTML]{f3f3f3}\cmark &\cellcolor[HTML]{f3f3f3} &\cellcolor[HTML]{f3f3f3} &\cellcolor[HTML]{f3f3f3} &\cellcolor[HTML]{f3f3f3} &\cellcolor[HTML]{f3f3f3} &\cellcolor[HTML]{f3f3f3} &\cellcolor[HTML]{f3f3f3} &\cellcolor[HTML]{f3f3f3}\cmark &\cellcolor[HTML]{f3f3f3}\cmark &\cellcolor[HTML]{f3f3f3}\cmark &\cellcolor[HTML]{f3f3f3} &\cellcolor[HTML]{f3f3f3}\cmark &\cellcolor[HTML]{f3f3f3} &\cellcolor[HTML]{f3f3f3} &\cellcolor[HTML]{f3f3f3}E-D \\
	\textbf{Yu~\etal~\shortcite{yu2019inverserendernet}} &\cmark & &SH & & & &\Aa \Nn & &\cmark &\cmark & &\cmark & & & &E-D \\
	\cellcolor[HTML]{f3f3f3}\textbf{Li~\etal~\shortcite{li2018cgintrinsics}} &\cellcolor[HTML]{f3f3f3}\cmark &\cellcolor[HTML]{f3f3f3} &\cellcolor[HTML]{f3f3f3} &\cellcolor[HTML]{f3f3f3} &\cellcolor[HTML]{f3f3f3} &\cellcolor[HTML]{f3f3f3}\Aa \Ss &\cellcolor[HTML]{f3f3f3}\Aa \Ss &\cellcolor[HTML]{f3f3f3}\Aa \Ss &\cellcolor[HTML]{f3f3f3}\cmark &\cellcolor[HTML]{f3f3f3} &\cellcolor[HTML]{f3f3f3}\cmark &\cellcolor[HTML]{f3f3f3} &\cellcolor[HTML]{f3f3f3}\cmark &\cellcolor[HTML]{f3f3f3} &\cellcolor[HTML]{f3f3f3} &\cellcolor[HTML]{f3f3f3}E-D \\
	\textbf{Ma~\etal~\shortcite{ma2018single}} &\cmark & & & & & &\Aa \Ss &$P_{\nabla A}$ &\cmark &\cmark & & & & & &E-D \\
	\cellcolor[HTML]{f3f3f3}\textbf{Bi~\etal~\shortcite{bi2018deep}} &\cellcolor[HTML]{f3f3f3}\cmark &\cellcolor[HTML]{f3f3f3} &\cellcolor[HTML]{f3f3f3} &\cellcolor[HTML]{f3f3f3} &\cellcolor[HTML]{f3f3f3} &\cellcolor[HTML]{f3f3f3} &\cellcolor[HTML]{f3f3f3}\Aa \Ss &\cellcolor[HTML]{f3f3f3} &\cellcolor[HTML]{f3f3f3}\cmark &\cellcolor[HTML]{f3f3f3}\cmark &\cellcolor[HTML]{f3f3f3}\cmark &\cellcolor[HTML]{f3f3f3} &\cellcolor[HTML]{f3f3f3} &\cellcolor[HTML]{f3f3f3} &\cellcolor[HTML]{f3f3f3} &\cellcolor[HTML]{f3f3f3}E-D \\
	\textbf{Lettry~\etal~\shortcite{lettry2018darn}} &\cmark & & & & & &\Aa \Ss &\Aa \Ss &\cmark & & & & & & &Res \\
	\cellcolor[HTML]{f3f3f3}\beyLamb{\textbf{Sengupta~\etal~\shortcite{sengupta2019neural}}} &\cellcolor[HTML]{f3f3f3} &\cellcolor[HTML]{f3f3f3}\cmark &\cellcolor[HTML]{f3f3f3}E &\cellcolor[HTML]{f3f3f3}Phong &\cellcolor[HTML]{f3f3f3} &\cellcolor[HTML]{f3f3f3}\Aa &\cellcolor[HTML]{f3f3f3}\Aa \Nn \Ll &\cellcolor[HTML]{f3f3f3} &\cellcolor[HTML]{f3f3f3}\cmark &\cellcolor[HTML]{f3f3f3} &\cellcolor[HTML]{f3f3f3} &\cellcolor[HTML]{f3f3f3} &\cellcolor[HTML]{f3f3f3} &\cellcolor[HTML]{f3f3f3} &\cellcolor[HTML]{f3f3f3} &\cellcolor[HTML]{f3f3f3}E-D \\
	\textbf{Zhou~\etal~\shortcite{zhou2019glosh}} &\cmark & &svSH & & &\Aa \Ss &\Aa \Ss \Nn \Ll &\Aa \Ss \Nn &\cmark & & & & & & &E-D \\
	\cellcolor[HTML]{f3f3f3}\textbf{Liu~\etal~\shortcite{liu2020unsupervised}} &\cellcolor[HTML]{f3f3f3}\cmark &\cellcolor[HTML]{f3f3f3} &\cellcolor[HTML]{f3f3f3} &\cellcolor[HTML]{f3f3f3} &\cellcolor[HTML]{f3f3f3} &\cellcolor[HTML]{f3f3f3} &\cellcolor[HTML]{f3f3f3} &\cellcolor[HTML]{f3f3f3} &\cellcolor[HTML]{f3f3f3}\cmark &\cellcolor[HTML]{f3f3f3} &\cellcolor[HTML]{f3f3f3}\cmark &\cellcolor[HTML]{f3f3f3} &\cellcolor[HTML]{f3f3f3} &\cellcolor[HTML]{f3f3f3}\cmark &\cellcolor[HTML]{f3f3f3} &\cellcolor[HTML]{f3f3f3}Res \\
	\beyLamb{\textbf{Li~\etal~\shortcite{li2020inverse}}} &\cmark &\cmark &svSG &svBRDF & &\Aa &\parbox{1cm}{\Aa \Nn \Dd \Ll \\  r$_{svBRDF}$ } & &\cmark & &\cmark & & & & &E-D \\
	\bottomrule
\end{tabular}
\end{table*}

\section{Learning Formulation}
\label{sec:learning_formulation}
The problem of inferring the intrinsic components of a scene given a single image has been traditionally (\textit{i.e.,} before deep learning algorithms flourished) addressed with optimization-based methods that make assumptions about the contents of the scene or the physics of the imaging process. For example, some methods assume clear-sky illumination~\cite{omer2004color,finlayson2004intrinsic,barron2014shape}, monochromatic lighting~\cite{bousseau2009user,barron2014shape,chang2014bayesian}, piecewise smooth reflectance \cite{bell2014intrinsic,bi20151,rother2011recovering}, or force areas with similar texture or chromaticity to have the same reflectance~\cite{garces2012intrinsic,zhao2012closed}. 
These assumptions, formulated as statistical priors, were combined within optimization frameworks such as Conditional Random Fields (CRFs)~\cite{krahenbuhl2011efficient,bell2014intrinsic}, multi-scale gradient-based solvers~\cite{barron2014shape}, or closed-form systems of equations~\cite{zhao2012closed}. However, as in many computer vision problems, finding the optimal solution is computationally very complex, and the use of ad-hoc priors and heuristic parameters narrows the scope and the generalization capabilities of the solution. For example, the common assumption that an edge might be produced by either a change of reflectance or a change in shading~\cite{horn1974determining,tappen2003recovering,grosse2009ground}, overlooks the fact that both changes might occur simultaneously as it happens at occlusion boundaries~\cite{garces2012intrinsic}.

In recent years, Convolutional Neural Networks (CNNs) ~\cite{lecun1989backpropagation,lecun1998gradient} have become the state-of-the-art models for solving many different computer vision tasks, such as object segmentation, image classification or image-to-image translation ~\cite{zhang2020resnest,simonyan2014very,isola2017image}. A key factor to their success is the fact that they internally learn hierarchical patterns that represent image features at multiple scales, in a way that loosely mimics the behavior of the visual cortex in mammals. Additionally, the recent boost in the size and variety of datasets, as well as the expansion of available computing power, have made those models ubiquitous in computer vision problems. Consequently, in the intrinsic image decomposition literature, CNNs have gradually substituted or complemented traditional hand-crafted priors and assumptions, to features that are directly learned from data. We refer to the problem of learning an intrinsic image decomposition from images using CNNs as Deep Intrinsic Images. There are several ways in which the solution to this problem can be approximated by CNNs. 

In this section, we describe the methods according to the formulation of the learning problem, or similarly, in the way data and prior knowledge can be leveraged to train a model capable of providing a solution. Despite the variety of existing inverse image formation models described in  Section~\ref{sec:theory}, all the methods share similar learning strategies, which are frequently compatible and combined in the objective function to provide the best performance. 
We have identified four main complimentary strategies which have been used in the literature so far:
\begin{itemize}
	\item Weak-Supervision: use human judgments about the perception of materials and illumination in images (Section~\ref{sec:weak}). 
	\item Full-Supervision: leverage labeled datasets in full regression frameworks in order to learn statistical priors over the parameters domain (Section~\ref{sec:fully}).  
	\item Self-Supervision: include an image formation loss in order to guarantee that the target parameters will effectively reconstruct the original input image (Section~\ref{sec:self}).
	\item Priors: explicitly model prior knowledge about the nature of each individual intrinsic component (Section~\ref{sec:priors}).
\end{itemize}

Figure~\ref{fig:overview_strategies} shows an overview of these strategies and how they relate to the existing datasets. Although the majority of the datasets serve a single purpose, some of them can be used to feed in more than one learning strategy, particularly the \textit{ShapeNet} dataset~\cite{chang2015shapenet} as it enables the generation of samples \textit{on-the-fly} during training.
Table~\ref{tab:supervision} shows the methods covered in this survey according to the previous categorization. Note that only a few methods follow a single strategy, while the majority of them combine three or four, thus, most of them will be discussed in more than one section.

\subsection{Weak-Supervision: Learn from Relative Human Judgments} \label{sec:weak}

As opposed to computers that only understand absolute color values, humans are very skilled in judging whether two surfaces are made of the same material despite variations in illumination~\cite{land1971lightness}. 
This ability, known as color constancy, has been exploited in two recent datasets, the IIW~\cite{bell2014intrinsic} and the SAW~\cite{kovacs2017shading} datasets, that contain human judgments about regions of images sharing \textit{similar albedo},  or with assigned \textit{shading labels} (smooth shading, normal/depth discontinuity, and shadow boundary), respectively.
These judgments are provided as sparse sets of pairwise comparisons and accompanied by a confidence score, the \textit{Weighted Human Disagreement Rate} ($WHDR$)  ~\cite{narihira2015judgement}, that takes into account disagreements between the raters.
Several methods have used these relative constraints to train, fine-tune or evaluate their models. 
Below we detail the two main strategies to leverage such weak supervision to train deep networks.

\subsubsection{Relative Predictions [RP]} 
Early approaches use this kind of data to train models that predict relative scores between two image regions. Then, they combine the output of these sparse predictions within existing inference frameworks to provide smooth estimations.
Narihira \etal~\shortcite{narihira2015judgement} (trained on IIW) predict local lightness relationships by means of a CNN used as feature descriptor combined with ridge ranking regression. They provide relative predictions, but unlike other methods, do not attempt to reconstruct the intrinsic components.
Using the SAW dataset, Kovacs \etal~\shortcite{kovacs2017shading} predict the source of a shading gradient --smooth shading, normal/depth discontinuity, and shadow boundary-- using a CNN along with a linear classifier. Then, they use the local prediction within a classical Retinex formulation~\cite{zhao2012closed} to estimate the intrinsic components.
Zhou \etal~\shortcite{zhou2015learning} builds on a dense CRF framework~\cite{krahenbuhl2011efficient,bell2014intrinsic} initialized by the output of a siamese network used as binary classifier which behaves as a prior for reflectance.
Concurrently, Zoran \etal~\shortcite{zoran2015learning} propose a framework to reason about ordinal relationships between local patches of the images; besides depth estimation, they prove to be successful on the IIW dataset for intrinsic decomposition. In this case, they use quadratic programming to propagate and smooth the estimations using superpixels~\cite{achanta2012slic}.

\subsubsection{Relative Comparisons as Weak Supervision [WHDRloss]} 
The second group of methods that use data from relative human judgments leverages such weak annotations for an extra supervision in the learning loss to constrain the reflectance, the shading, or both, depending on the dataset used.
Instead of combining the prediction with external frameworks, these methods estimate the intrinsic components using only the predictions of their neural networks. 
To this end, they use the  ($WHDR$) which, as mentioned, is a distance metric developed to evaluate the quality of the automatic estimations with respect to the human ratings.
Nestmeyer \etal~\shortcite{nestmeyer2017reflectance} proposed the only method to use this loss without further supervision to provide dense estimations within an end-to-end deep framework. They were followed by other methods \cite{fan2018revisiting,sengupta2019neural,li2018cgintrinsics,zhou2019glosh}, which included this strategy as a form of weak-supervision to fine-tune the models trained with other sources of data.

\subsection{Full-Supervision: Learn from Labeled Data} \label{sec:fully}

The use of labeled datasets for training machine learning models is a common training strategy, often necessary for obtaining successful models.
The majority of methods mentioned in this section use absolute label values in regression losses. That is, for each labeled image, the error function is penalized if the estimated intrinsic components do not conform with the ground truth labeled ones. Only using this kind of supervision~\cite{narihira2015direct,shi2017learning}, however, is not a guarantee that the estimated components will faithfully reconstruct the input image. Consequently, the  majority of methods combine full regression with other forms of supervision, as shown in Table~\ref{tab:supervision}.
An alternative way of using labeled data without explicitly connecting the labels in the loss function by means of regression is proposed by Liu~\etal~\cite{liu2020unsupervised}. In such solution, large sets of images of each unknown intrinsic component were used to train individual latent spaces representations of each domain. We discuss such case in Section~\ref{sec:priors}.

\NEW{Jointly training with real and synthetically generated data is becoming a popular and successful trend (see Table~\ref{tab:method_dataset}). Synthetic data may be useful to teach the model the global shapes and common geometries, while real data is necessary to fill the domain gap necessary for the methods to work well with images taken under real-world illuminations and materials. 
}

As common with any image processing algorithms, the regression of the intrinsic components can be done in the original domain of the image pixels or in the gradient domain. 
This may help the neural networks learn a better representation of the problem, but also hinder their learning capabilities by restricting the space of solutions they can find.
We have classified the methods in this section according to whether the regression problem is formulated in the original domain of the intrinsic parameters, or whether they are transformed to the gradient domain before applying $\ell^1$, $\ell^2$, or perceptual regression losses. Note that other transformations applied to the intrinsic components before performing regression are also possible, for example, applying a bilateral filter to the reflectance layer. We discuss such cases in Section~\ref{sec:priors}.

\subsubsection{Original Domain [\textbf{$O$}]} Most of the methods regress the intrinsic components using either $\ell^1$ or $\ell^2$ losses. This strategy is sometimes referred as Direct Intrinsic \cite{narihira2015direct,shi2017learning,baslamisli2018cnn,fan2018revisiting,li2018watching,ma2018single,bi2018deep,lettry2018darn,zhou2019glosh}, and can be used to initialize just part of the network ~\cite{janner2017self,sengupta2019neural,meka2018lime}.

\subsubsection{Gradient Domain [\textbf{$\nabla$}]} Inspired by classical Retinex approaches \cite{horn1974determining,grosse2009ground}, which obtain the shading image by first predicting the gradients and then integrating them with Poisson-like reconstruction, a set of methods explicitly regress the gradients of the components. This loss has been applied only to the albedo~\cite{narihira2015direct,shi2017learning}, or both the albedo and shading layers~\cite{baslamisli2018cnn,fan2018revisiting,li2018cgintrinsics,lettry2018darn,zhou2019glosh}.
The probabilistic version of working in the gradient domain is presented by the methods that predict the probability of an albedo gradient ($P_{\nabla A}$). 
Fan \etal~\shortcite{fan2018revisiting} complement a Direct Intrinsic network with a Guidance Network that is trained to predict binary albedo edges emulating the response of applying a L1 flattening~\cite{bi20151}, method used as ground truth. Similarly, Ma \etal~\shortcite{ma2018single} explicitly predict a soft assignment mask with the probability of an albedo edge which, unlike before~\cite{fan2018revisiting}, is trained using the ground truth albedo and not a flattened version of it.

\subsection{Self-Supervision: Learn from the Image Formation Model} \label{sec:self}

There are several problems that make a fully supervised approach with explicit labels insufficient: First, it is not determined how perceptual differences will be distributed between the different intrinsic components. Second, there is no guarantee that the reconstructed image from the inferred components will exactly match the input. Finally, full regression methods require a huge amount of labeled data in order to generalize reasonably. 
In this context, self-supervised strategies formulated as the \textit{Render Loss} or the \textit{Image Formation Loss} have evolved to address these limitations.
The key of these strategies is to introduce, at training time, per-pixel reconstruction of the input image as an additional signal to guide the learning process.
\NEW{This operation can be done for each single image of the training dataset (\textit{Single-Image Consistency}), or across images for multi-image datasets with relative constraints (\textit{Multi-Image Consistency}). }
Thus, it is critical to perform this reconstruction efficiently, as it has to be performed thousands of times during training.
We discuss the trade-offs of choosing an image formation model in Section~\ref{sec:inv} and recent trends that allow more expressive models in Section~\ref{sec:future}. Here we assume that reconstructing the image from a set of intrinsic parameters can be done in a negligible amount of time.

\subsubsection{Single-Image Consistency [\textbf{$SVC$}]}

One of the most critical decisions in the problem of single image inverse reconstruction is to choose the model that should reconstruct the scene. As discussed in Section~\ref{sec:inv}, there is a trade-off between the desired complexity of the target scene --geometry, material, illumination--, and the model complexity.
For this reason, most of the methods dealing with arbitrary scenes choose simple formulations, either the intrinsic diffuse or intrinsic residual, with limited number of parameters that can be evaluated in real time. 
The \textit{intrinsic diffuse} model receives as input the albedo and shading images, which, when multiplied together, reconstruct the input image \cite{cheng2018intrinsic,li2018cgintrinsics,lettry2018darn,li2018watching,yu2019inverserendernet,ma2018single,bi2018deep,lettry2018darn,baslamisli2018cnn}. The \textit{intrinsic residual} has an additional parameter to capture the specular reflections and other light effects that do not belong to the diffuse behavior of light \cite{meka2018lime,sengupta2019neural}.
Regardless of the inverse model and the internal architecture used to capture the intermediate steps of the physics of the image formation process (more details in Section~\ref{sec:inv} and Section~\ref{sec:architectures}), reconstructions losses guarantee that the estimated components will be able to reconstruct the input image.

\subsubsection{Multi-Image Consistency [$MIC$]}

One important characteristic of the diffuse albedo of the materials is that its value remains constant despite variations of other scene properties such as the illumination or the view angle. This property has been exploited as an extra form of supervision by leveraging existing datasets which might not be specifically collected for the purpose of intrinsic decomposition and lack of explicit labels.

\textit{Time-lapse sequences}, \textit{i.e.} sequences of images of a static scene with varying illumination, have been widely used for years as input to estimate the intrinsic components. 
Weiss \etal~\shortcite{weiss2001deriving} learn from the data that shading images of outdoor scenes convolved with a derivate filter are 
sparse, and apply it as a prior to estimate the intrinsic components.
Sunkavalli \etal~\shortcite{sunkavalli2007factored} additionally decomposed the scene into shadows, shading and reflectance under the assumption of clear-sky illumination. Later on, Laffont \etal~\shortcite{laffont2015intrinsic} used the albedo invariance to constrain the decomposition within a classical optimization framework. 
\textit{Multi-view sequences} contain the same scene under different perspectives or viewpoints. This kind of datasets have been mostly gathered with the purpose of 3D reconstruction, although occasionally used within the context of intrinsic decomposition. The method of Laffont \etal~\shortcite{laffont2012coherent} used online photo collections as input to guide the decomposition, leveraging cues from partial 3D reconstruction. Duchene \etal~\shortcite{duchene2015} further provide a full 3D model of the scene enabling relighting applications of outdoor scenes.  
As opposed to the methods reviewed in this survey, which leverage these datasets for training only, these approaches require as input the whole sequence of images to decompose a single view of the scene.

The key idea of Multi-Image Consistency is to combine different estimations for reflectance and shading images taken from the same \textit{sequence} but from different \textit{images} of the sequence.
The most frequent approach is to combine these cues with fully supervised training \cite{bi2018deep,ma2018single}.
Li \etal~\shortcite{li2018watching} propose the only method that not requiring any explicitly labeled data for training but heavily relying on heuristic priors (as described in the next section).
Finally, Yu \etal~\shortcite{yu2019inverserendernet} are the first to use a multi-view stereo dataset which contains rich variations in illumination to train a single image inverse method, capable of recovering albedo, normals and two spherical harmonic lighting coefficients. The network is mainly trained in a self-supervised manner by cross-projecting the views using depth maps and camera projection matrices, imposing coherency in the reconstruction and in the inferred albedo, as previous methods do ~\cite{li2018watching,ma2018single,bi2018deep}.

\subsection{Priors} \label{sec:priors}

Priors are the existing beliefs about a problem.
In classical non-learning based approaches, the priors about each intrinsic component were, first, observed from the data and then modeled as hand-crafted image heuristics, taking into account each phenomena independently. 
For example, it was observed that under natural daylight and for narrow-band camera sensors, pixels with the same reflectance form a single line in logRGB space. Such observation was used, for example, to identify shadows boundaries~\cite{finlayson2004intrinsic}.
An extensive overview of these priors and assumptions from a classical perspective is presented in Bonneel \etal ~\shortcite{bonneel2017intrinsic}. 
Using deep learning architectures has deemed the use of such priors unnecessary in the majority of situations, as they are now implicitly learned by the deep model during training. However, due the complexity of the inverse reconstruction problem, some of these priors have proven to be still useful nowadays. In the following we focus on these heuristic priors, as well as present an different approach to learn them from the data.

\subsubsection{Albedo is Piece-wise Flat [\Aflat]} 
Observing that the human visual system perceives colors locally and constantly with independence of the illumination conditions, the Retinex theory~\cite{land1971lightness} was fundamental for the development of the most popular computational prior on the albedo, which assumes that it is
piece-wise flat, of high frequency, and sparse~\cite{bousseau2009user,bi20151}.
In a deep learning formulation, this prior can be applied in two ways: First, as an additional loss term that applies to the albedo only, being the L1 loss the most natural way to impose such constrain \cite{li2018watching,li2018cgintrinsics,ma2018single,cheng2018intrinsic}. Second, as an explicit filtering operation applied over such component to guide the learning process in a more aggressive way.
In the latter case, the l1 flattening algorithm~\cite{bi20151} or different variations of the bilateral filter~\cite{gastal2012adaptive,poole2016fast,tomasi1998bilateral} have been the most popular and successful strategies ~\cite{bi2018deep,cheng2018intrinsic,fan2018revisiting,nestmeyer2017reflectance}.

\subsubsection{Albedo is Sparse  [\Asparse]} In order to reduce the complexity of the problem, previous work relied on two strategies also related to the appearance of the albedo component. First, assuming that similar chromaticities values of the input image are likely to have the same albedo values, and second, that the amount of different albedos within a natural image is sparse~\cite{garces2012intrinsic,bell2014intrinsic,shen2011intrinsic}.
Yu \etal~\shortcite{yu2019inverserendernet} explicity take into account the former by apply a pixel-wise weighted penalty according to chromaticity values of the input image.
The latter was implicitly considered by Cheng~\etal~\shortcite{cheng2018intrinsic}, who use a deep perceptual loss in order to preserve textured details~\cite{johnson2016perceptual}.

\subsubsection{Shading is Smooth [\Ssmooth]} Also derived from Retinex, and from the assumption of convex and smooth 3D geometries, this prior assumes that smooth image variation are mostly due to the interaction of light with a continuous smooth surface~\cite{horn1974determining}. Existing work has modeled it with the $\ell^1$ or $\ell^2$ norms~\cite{cheng2018intrinsic,li2018cgintrinsics}, or minimizing second-order shading gradients~\cite{ma2018single}.

\subsubsection{Data-driven Priors [Data]} 

In a learning-based approach, the priors are directly learned from the available data as probability distributions.
In a fully-supervised setup, the most common way to leverage labeled data is by feeding the network with one image along with its corresponding intrinsic components. 
After enough data and iterations of the training process, the deep neural network will have learned an internal and \textit{coupled} implicit representation of each of the intrinsic components.
An alternative way to use labeled data for training was proposed by Liu~\etal~\cite{liu2020unsupervised}. They present a deep network architecture that do no require aligned relationships between the intrinsic components and the input image in order to learn the prior distribution of each of them. Instead, it feeds in the network with independent sets of images of each intrinsic component. In this regard, the prior information is given by the actual data, and the work of understanding the unique features of each component is a learning task.

\section{Deep Neural Network Architectures} \label{sec:architectures}
In the previous section, we discussed the problem of deep intrinsic decomposition from a learning formulation perspective. The design of the deep neural network architecture that is used to learn to decompose shading and reflectance is also important, as different network architectures are biased towards finding solutions of different characteristics. The neural network architecture categorization proposed herein is summarized in Table~\ref{tab:supervision}. In this section, we will describe the methods according to their network architecture design choices. We have identified two main groups that have been used in the literature so far: 

\begin{itemize}
    \item[-] Networks designed to learn \textbf{pairwise comparisons} between patches of an image (Section~\ref{sec:AR_PWC}).
    \item[-] Networks designed to perform an \textbf{image-to-image translation}, from the input image to either an intermediate representation of the decomposition, or to the full image decomposition (Section~\ref{sec:AR_I2IT}). 
\end{itemize}

\subsection{Pairwise Comparison}  \textbf{(P-W C)}\label{sec:AR_PWC}
As discussed in Section~\ref{sec:weak}, early deep intrinsic decomposition methods relied on sparse local judgments of either reflectance or shading. Some methods proposed deep architectures that learned to solve that exact problem: given two patches of an image, find their relative magnitude of lightness. Narihira \etal~\shortcite{narihira2015judgement} fine-tune an Alexnet network \cite{krizhevsky2012imagenet} pre-trained on Imagenet ~\cite{deng2009imagenet}, and use its last fully-connected layer as a feature descriptor vector of the input image. To compare the relative lightness of two patches of images, they perform ridge-ranking regression using their two feature vectors as input. Other methods argue that using only local patches does not provide the model with enough information, as the context of the whole image is lost. Zhou \etal~\shortcite{zhou2015learning} propose a three-stream deep convolutional architecture that performs the relative lightness prediction by combining --inside a shared vector-- the features extracted of three images: the two patches and the global input image. This vector is enriched with the spatial coordinates of the two patches. A set of fully connected layers uses this shared vector to predict the relative lightness. Similarly, Zoran \etal~\shortcite{zoran2015learning} propose a multi-stream convolutional architecture that receives both patches, the global image, as well as masks for both patches, a bounding box and a \textit{region of interest} image. The predictions of the convolutional networks within the architecture are then aggregated in a feature vector, used by a block of fully-connected layers to perform the relative lightness prediction.

\subsection{Image-to-Image Translation} \label{sec:AR_I2IT}
Instead of relying on pairwise comparisons between patches of images, many methods perform a direct prediction of either the shading, the reflectance, or both maps, in an end-to-end fashion. This type of framework reduces the amount of post-processing needed to complete the intrinsic decomposition, and are more easily integrated with other differentiable modules (e.g. a differentiable render, differentiable filtering layers, etc.), which may help increase their learning capabilities. We have divided those image-to-image translation networks into three groups, depending on how they learn the mapping between input image and intrinsic decomposition, from a network architecture perspective. As we will see in Section~\ref{sec:AR_IT2IT_BM}, earlier methods rely on a simple set of connected convolutional layers to solve this problem, without the use of any skip or mirror connections. Previous work~\cite{isola2017image,ronneberger2015u,newell2016stacked} show that the use of skip connections between layers that represent features at different scales is helpful for preserving rich details on the output maps. Those skip connections can be included in encoder-decoder architectures (Section~\ref{sec:AR_IT2IT_ED}) or using residual connections (Section~\ref{sec:AR_IT2IT_RN}). 

\subsubsection{Baseline Methods} (\textbf{B I2IT}) \label{sec:AR_IT2IT_BM}
Some methods propose a deep architecture similar to AlexNet ~\cite{krizhevsky2012imagenet} or VGG-16 ~\cite{simonyan2014very}, which are comprised of a set of convolutional layers, followed by a block of fully-connected layers. Such is the case of the architecture in Kovacs \etal~\shortcite{kovacs2017shading}, where they use a VGG network (trained on Image-Net) and add to it 3 fully-connected layers, which help complete their shading prediction. The use of fully-connected layers is limiting, as it forces the input image to have some specific dimensions. Consequently, multiple methods use only convolutional layers for their predictions. Nestmeyer \etal~\shortcite{nestmeyer2017reflectance} propose a fully-convolutional architecture, which outputs a reflectance intensity prediction, then transformed to reflectance and shading maps using differentiable operations. The use of a fully-convolutional architecture is also proposed in Narihira \etal~\shortcite{narihira2015direct}. Their method performs directly the decomposition estimation by combining two networks that process the input image at different scales.

\subsubsection{Encoder-Decoder Networks}  (\textbf{E-D I2IT}) \label{sec:AR_IT2IT_ED}
A common approach for deep intrinsic image decomposition is to use an encoder-decoder architecture~\cite{ronneberger2015u}. Those architectures are composed of an \textit{encoder}, which translates the input image to a dense, information-rich, latent space, which is then transformed into another image by a \textit{decoder}. In the deep intrinsic images literature, all the methods that use encoder-decoder architectures transform the input image using a shared encoder, followed by different decoders for every map they want to predict. To preserve rich spatial details, they enhance their models with skip connections between mirrored layers in the encoder and decoder networks. Learning different decoders for the shading and reflectance layers is a common approach \cite{li2018cgintrinsics,baslamisli2018cnn,ma2018single}. Different enhancements can be performed to this architecture, such as predicting a residual layer~\cite{shi2017learning}, an illumination color~\cite{li2018watching}, an environment map~\cite{sengupta2019neural}, or sharing activation values between decoders~\cite{bi2018deep}. This type of architecture can also be used to predict normal~\cite{janner2017self,yu2019inverserendernet,li2020inverse,zhou2019glosh} or specular maps~\cite{meka2018lime,li2020inverse}, which can then be processed to estimate shading. 

\subsubsection{Residual Networks} (\textbf{Res I2IT})  \label{sec:AR_IT2IT_RN} 
A different way of preserving rich multi-scale spatial information is by using residual skip connections ~\cite{he2016deep}. In a residual block within a convolutional neural network, the input is processed by a set of convolutional and non-linear activation layers, the result of which is added to the original input. This helps the flow of gradients during back-propagation, and adds spatial details to image-to-image translation networks. Instead of only using these skip connections between mirrored layers, as in the methods in Section~\ref{sec:AR_IT2IT_ED}, those skip connections are used in every layer of the network. Residual networks for intrinsic image decomposition has been proposed in different fashions. The model of Fan \etal~\shortcite{fan2018revisiting} contains two residual networks: one which predicts albedo intensities, and other that estimates the probability of each pixel corresponding to an edge on the albedo. A fully-convolutional residual network is proposed in Lettry \etal~\shortcite{lettry2018darn}, which predicts the shading of the input image. The residual network proposed in Cheng \etal~\shortcite{cheng2018intrinsic} predicts a laplacian pyramid, which is then collapsed into the decomposition prediction. The architecture proposed in \cite{liu2020unsupervised} performs the intrinsic decomposition by leveraging residual deep latent spaces for unsupervised image domain translations. It is worth mentioning that the Intrinsic Residual Network (IRN) proposed in Sengupta \etal~\shortcite{sengupta2019neural} uses residual layers as part of their encoder-decoder architecture.

\begin{table}[!htp]\centering	
	\caption{Methods presented in the paper and datasets used for Training (T) and Evaluation (E) purposes. *These datasets contain only isolated objects. $^1$ Uses BigTime~\shortcite{li2018watching}; $^2$ Uses MegaDepth~\shortcite{li2018megadepth}; $^3$ Uses a custom multi-illumination dataset.}\label{tab:method_dataset}
	\small
	\begin{tabular}{lcccccc}
		\toprule
		\multirow{2}{*}{\textbf{Method}} 
		&\multicolumn{3}{c}{\textbf{Synthetic }} 
		&\multicolumn{3}{c}{\textbf{Real}} \\
		\cmidrule(r){2-4} \cmidrule(l){5-7}
		&\rotf{*ShapeNet-D} 
		&\rotf{MPI Sintel} 
		&\rotf{CG} 
		&\rotf{*MIT Intrinsic} 
		&\rotf{IIW/SAW} 
		&\rotf{Other} \\
		\midrule
		Narihira~\etal~\shortcite{narihira2015judgement} & & & & &TE & \\
		Narihira~\etal~\shortcite{narihira2015direct} & &TE & &TE & & \\
		Zhou~\etal~\shortcite{zhou2015learning} & & & & &TE & \\
		Zoran~\etal~\shortcite{zoran2015learning} & & & & &TE & \\
		Kovacs~\etal~\shortcite{kovacs2017shading} & & & & &TE & \\		
		Nestmeyer~\etal~\shortcite{nestmeyer2017reflectance} & & & & &T & \\
		Shi~\etal~\shortcite{shi2017learning} &T & & &TE & & \\
		Janner~\etal~\shortcite{janner2017self} &T & & & & & \\
		Meka~\etal~\shortcite{meka2018lime} &T & & & & & \\
		Baslamisli~\etal~\shortcite{baslamisli2018cnn} &T & & &E & & \\
		Cheng~\etal~\shortcite{cheng2018intrinsic} & &TE & &TE & & \\
		Fan~\etal~\shortcite{fan2018revisiting} & &TE & &TE &TE & \\
		Li~\etal~\shortcite{li2018watching} & & & &TE &E/E &T$^1$ \\
		Yu~\etal~\shortcite{yu2019inverserendernet} & & & & &E &T$^2$ \\
		Li~\etal~\shortcite{li2018cgintrinsics} & & &TE & &TE/TE & \\
		Ma~\etal~\shortcite{ma2018single} &T & & &TE & & \\
		Bi~\etal~\shortcite{bi2018deep} & &TE & & &E &T$^3$ \\
		Lettry~\etal~\shortcite{lettry2018darn} & &TE & &TE & & \\
		Sengupta~\etal~\shortcite{sengupta2019neural} & & &TE & &TE & \\
		Zhou~\etal~\shortcite{zhou2019glosh} & & &TE & &TE/TE & \\
		Liu~\etal~\shortcite{liu2020unsupervised} & &TE &TE &TE &E & \\
		Li~\etal~\shortcite{li2020inverse} & & &TE & &TE & \\
		\bottomrule
	\end{tabular}
	
\end{table}

\section{Evaluation}\label{sec:evaluation}

The quality of an intrinsic decomposition method is challenging to evaluate.
As we argue in Sections \ref{sec:theory} and \ref{sec:inv}, one of the reasons is that the common assumption about Lambertian materials does not hold in real life. Consequently, the methods wrongly assign the additional light effects either to the shading or to the albedo layers. 
The existence of explicitly labeled datasets has facilitated the task, which was done for a long time by means of visual side-by-side comparisons, and most of the methods nowadays quantify similarity to the ground truth using pixel-wise errors metrics. 
As opposed to using explicitly labeled datasets, several methods have evaluated the algorithmic performance by comparing it with humans doing the same task. These forms of evaluation have some trade-offs that we discuss in Section~\ref{sec:discussion}. In the following, we describe the error metrics along with the reported performance of each of the reviewed works. We will report the error metrics for the two most commonly used datasets to compare different approaches for intrinsic decomposition.

\subsection{Pixel-wise Error Metrics}
A common way of evaluating image regression models is through pixel-wise error metrics. In them, the regressed output and the target image are compared pixel-by-pixel, using a given distance metric, such as the $\ell^1$ or $\ell^2$ norms. Pixel-wise error metrics provide an estimation of the quality of the output of the model that fails to take into account spatial information, as individual pixels are considered to be independent from their neighboring pixels. Consequently, such error metrics are not able to provide an estimation of the perceptual quality of the decomposition, in ways that deep perceptual-aware metrics are capable of (see ~\cite{gatys2017controlling,huang2017arbitrary,zhang2018unreasonable}). 

Nevertheless, pixel-wise distances have been widely used in the intrinsic decomposition literature, particularly for evaluating the quality of the results obtained on the MIT dataset~\cite{grosse2009ground}. Pixel-wise approaches can be used for evaluating the quality of intrinsic decomposition methods for this dataset, because it contains densely-annotated data, as discussed in Section~\ref{sec:datasets}. In particular, the most common way of reporting errors for this dataset has been through the Scale Invariant L2 Loss, which is a modified version of the $\ell^2$ norm that ignores the scale of the (log) shading and albedo maps. When averaging this loss over overlapping windows, the Local Mean Squared Error (LSME) error can be computed. A more detailed description of these error metrics is found in ~\cite{narihira2015direct}. 

In Table~\ref{tab:MIT_error}, a comprehensive comparison of LMSE metrics on the MIT dataset can be found. In particular, we detail the self-reported average LMSE (averaged between shading and albedo LSME), for multiple train and test splits. Many of the methods do not provide information about the test dataset they are evaluating their method on, so we simply include the self-reported evaluation metrics, as well as their own comparisons with other methods in the intrinsic decomposition literature. As previously discussed, the MIT dataset does not contain sufficient data for a deep learning model to learn from. Consequently, a common approach is to train the deep learning model using other datasets, and evaluate on a test split of the MIT dataset. Nevertheless, many methods further fine-tune their model using a portion of the MIT dataset, so as to improve the results on its test split. We include both the baseline and the fine-tuned models' results for the methods that perform this fine-tuning on the MIT dataset. As it can be seen, despite the fact that comparing methods on this dataset is difficult (because in many cases the train/test split is unreported), there is a trend towards smaller LMSE errors on this dataset. This may indicate that there has been progress on the quality of the deep learning models and learning frameworks used to train them.

\begin{table*}[!htp]\centering
	\caption{Average LMSE (between shading and reflectance maps) reported for the MIT dataset~\cite{grosse2009ground}, for multiple train/test splits. We only show the results for each paper's best-performing method. Many authors provide the results of their deep learning model with and without fine-tuning their weights on a training set of the MIT dataset. We include both results both results for fairness when comparing methods. $*$~\cite{narihira2015direct} also splits the test set by objects, instead of by images, as done in \cite{barron2014shape}. $\dagger$ These methods are all evaluated using the same train/test split, which is unknown.}\label{tab:MIT_error}
	\begin{tabular}{lrrrrrrrrr}\toprule
		\multirow{2}{*}{\textbf{Method}} &\multicolumn{2}{c}{\textbf{\shortcite{barron2014shape} split}} &\multicolumn{2}{c}{\textbf{Other / unknown split}} &\multicolumn{4}{c}{\textbf{Error reported for other methods}} \\\cmidrule{2-9}
		&\textbf{Baseline} &\textbf{Finetune on \shortcite{grosse2009ground}} &\textbf{Baseline} &\textbf{Finetune on \shortcite{grosse2009ground}} &\textbf{\shortcite{barron2014shape}} &\textbf{\shortcite{narihira2015direct}} &\textbf{\shortcite{shi2017learning}} &\textbf{\shortcite{fan2018revisiting}} \\\cmidrule{1-9}
		Narihira~\etal~\shortcite{narihira2015judgement} & &0.0218 &$0.0234^*$ &$0.0224^*$ &0.0125 &0.0218* & & \\
		Shi~\etal~\shortcite{shi2017learning} & & &$0.0535^\dagger$ &$0.0375^\dagger$ &0.0292 &0.044 &$0.0375$ & \\
		Baslamisli~\etal~\shortcite{baslamisli2018cnn} \\ (IntrinsicNet) & & &$0.0226^\dagger$ & & &0.044 &0.0535 & \\
		Cheng~\etal~\shortcite{cheng2018intrinsic} &0.0133 &0.0121 & & &0.0125 &0.0239 &0.0271 &0.02 \\
		Fan~\etal~\shortcite{fan2018revisiting} &0.0203 & & & &0.0125 & &0.0271&0.0203 \\
		Li~\etal~\shortcite{li2018watching} &0.0297 & & & &0.0292&0.044 &0.0372 & \\
		Ma~\etal~\shortcite{ma2018single} & & &$0.0105^\dagger$ &$0.0063^\dagger$ &0.005 & &0.0098 & \\
		Lettry~\etal~\shortcite{lettry2018darn} & & & &0.00055 & &0.00091 & & \\
		\bottomrule
	\end{tabular}
	
\end{table*}

\subsection{Human Disagreement Metrics}
A different way of evaluating the quality of the intrinsic decomposition performed by machine learning models is by comparing the results of their outputs to judgments of humans that were asked to assess the relative lightness of two patches of an image. Such approach is helpful for datasets in which no ground-truth values are available, but for which sparse humans annotations can be found.

A dataset with these characteristics is the Intrinsic Images in the Wild (IIW) dataset ~\cite{bell2014intrinsic}, which is annotated by humans using relative reflectance values, as described in Section \ref{sec:datasets}. In the images in this dataset, there are sparse human annotations of the relative lightness (lighter, darker, same lightness) of pairs of patches of the image. This dataset has been widely used on the intrinsic decomposition literature, and it provides a specific train/test split, which helps make comparisons fair. 

To evaluate the quality of the results on the images in this dataset, most methods have used the \textit{Weighted Human Disagreement Rate} (WHDR), which compares the proportion of human judgments that the models disagrees with, weighted by the confidence of each pairwise judgments. This error metric may be more perceptually accurate than using simple pixel-wise metrics, but it fails to account for the intensity of the relative lightness values (e.g. how much lighter one patch of the image is compared to another), and it only measures the quality of the predictions on a subset of the images, as annotations are sparse. 

A summary of the self-reported WHDR values for the methods that use the IIW dataset for evaluation can be found on Table~\ref{tab:IIW_Error}. For each method, we provide their self-reported WHDR value and the values they report for other methods, as the latter do not necessarily agree with their own self-reported values. As it can be seen, there is a downward trend of the error metrics, which indicates that the most recent approaches, that include self-supervision, better priors, or more sophisticated neural architectures, may be more suitable for performing the intrinsic decomposition task. However, smaller values on this error metric does not necessarily mean that the decomposition performed by the model is physically accurate, as we will discuss in Section \ref{sec:discussion}. 

\begin{table*}[!htp]\centering
	\caption{Reported WHDR error for the IIW dataset~\cite{bell2014intrinsic}. $\dagger$These methods were not trained on IIW human ratings. $*$ indicate the values obtained using the test set proposed in~\cite{zoran2015learning}.}\label{tab:IIW_Error}
	\begin{tabular}{lrrrrrrrrrrr}\toprule
		\multirow{2}{*}{\textbf{Method}} &\multirow{2}{*}{\textbf{WHDR}} &\multicolumn{9}{c}{\textbf{Error reported for other methods}} \\\cmidrule{3-11}
		& &\textbf{\cite{narihira2015direct}} &\textbf{\cite{narihira2015judgement}} &\textbf{\cite{zhou2015learning}} &\textbf{\cite{zoran2015learning}} &\textbf{\cite{nestmeyer2017reflectance}} &\textbf{\cite{shi2017learning}} &\textbf{\cite{fan2018revisiting}} &\textbf{\cite{li2018watching}} &\textbf{\cite{bi2018deep}} \\\midrule
		Narihira~\etal~\cite{narihira2015judgement} &18.1 & &18.1 & & & & & & & \\
		Zhou~\etal~\cite{zhou2015learning} &15.7 & &18.1 &15.7 & & & & & & \\
		Zoran~\etal~\cite{zoran2015learning}* &17.86 & & & &17.86 & & & & & \\
		Nestmeyer~\etal~\cite{nestmeyer2017reflectance} &17.69 & & &19.95 &17.85 &17.69 & & & & \\
		Fan~\etal~\cite{fan2018revisiting} &14.45 & & &19.95 &17.85 &17.69 & &14.45 & & \\
		Li~\etal~\cite{li2018cgintrinsics} &14.8 &37.3 & &19.9 & &17.7*/19.5 &59.4 & & &17.7 \\
		\beyLamb{Sengupta~\etal~\cite{sengupta2019neural}} &16.7 & & &19.9 & &19.5 & & & & \\
		\beyLamb{Li~\etal~\cite{li2020inverse}} &15.93/$\dagger$21.99 & & & & & & & & & \\
		Li~\etal~\cite{li2018watching} &$\dagger$20.3 &37.3 &18.1 &15.7*/19.9 & & &59.4 & &20.3 & \\
		Bi~\etal~\cite{bi2018deep} &$\dagger$17.18 &40.9 & &19.9 &17.85 &17.69 &54.44 & & &17.18 \\
		Yu~\etal~\cite{yu2019inverserendernet} &$\dagger$21.4 &37.3 & &19.9 & &19.5 &59.4 &14.5 & & \\
		Zhou~\etal~\cite{zhou2019glosh} &15.2 & & &19.9 & &19.5 & &15.4 & & \\
		Liu~\etal~\cite{liu2020unsupervised} &$\dagger$18.69 & &18.1 &15.7 & & & &14.45 &20.3 &20.94 \\
		\bottomrule
	\end{tabular}
\end{table*}

\begin{figure*}[htb]
\begin{overpic}[width=\linewidth,scale=1.0,unit=1mm]{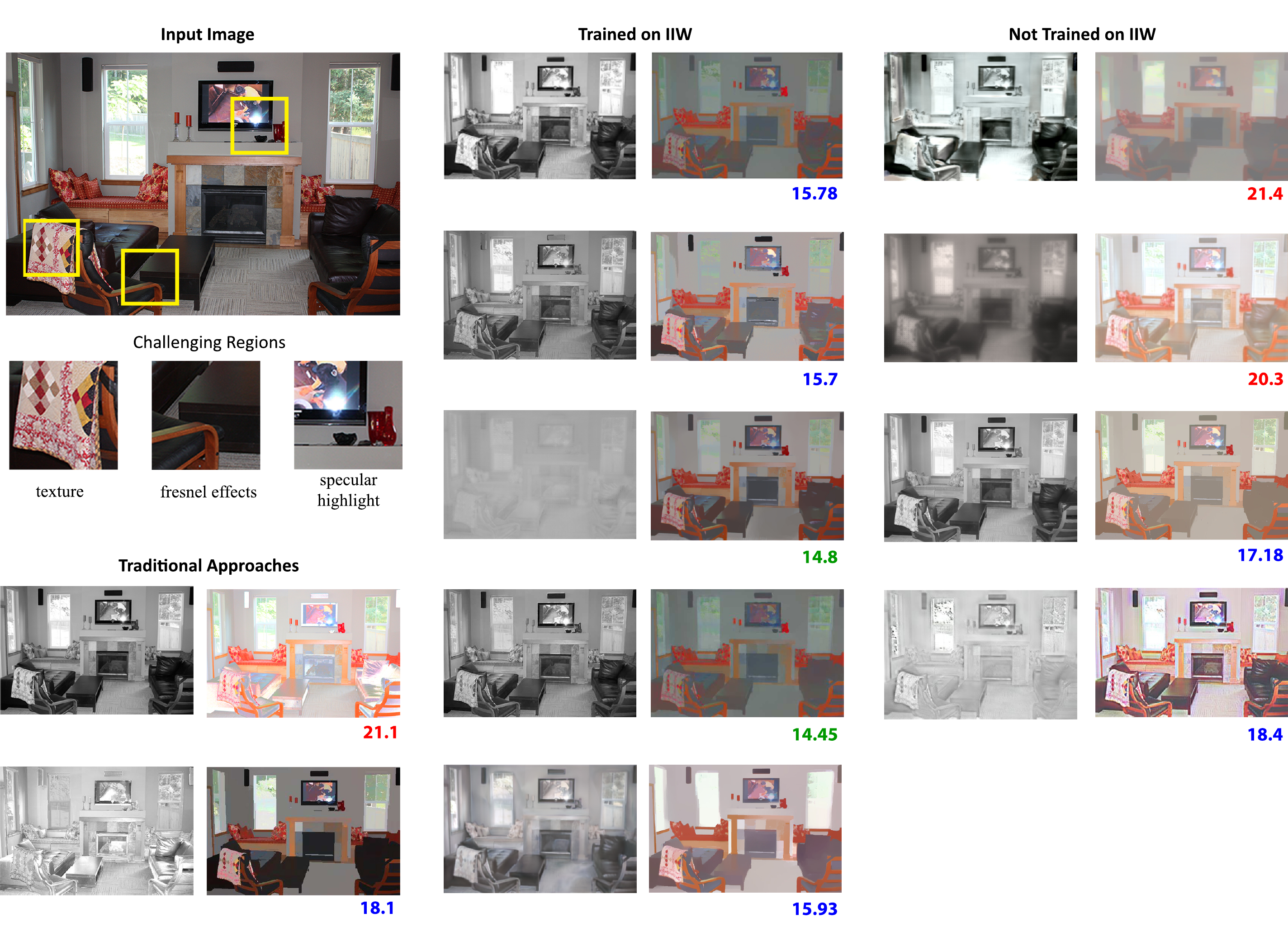}
	\put(28,26){\makebox(0,0){\small{Baseline: Luminance - Chromaticity}}}
	\put(28,1.5){\makebox(0,0){\small{Bi \etal~\cite{bi20151}}}}
	\put(88,98){\makebox(0,0){\small{Nestmeyer \etal~\cite{nestmeyer2017reflectance}}}}
	\put(88,74){\makebox(0,0){\small{Zhou \etal~\cite{zhou2015learning}}}}
	\put(88,50){\makebox(0,0){\small{\textit{CGIntrinsics} Li \etal~\cite{li2018cgintrinsics}}}}
	\put(88,26){\makebox(0,0){\small{Fan \etal~\cite{fan2018revisiting}}}}
	\put(88,1.5){\makebox(0,0){\small{Li \etal~\cite{li2020inverse}}}}	
	\put(147,98){\makebox(0,0){\small{Yu \etal~\cite{yu2019inverserendernet}}}}
	\put(147,74){\makebox(0,0){\small{Li \etal~\cite{li2018watching}}}}
	\put(147,50){\makebox(0,0){\small{Bi \etal~\cite{bi2018deep}}}}
	\put(147,26){\makebox(0,0){\small{Liu \etal~\cite{liu2020unsupervised}}}}

\end{overpic}
\caption{Qualitative comparison with an input image from the IIW dataset. Here we compare results between traditional non-learning based solutions, methods trained or fine-tuned on the IIW dataset, and methods not trained on such dataset. The quantitative error is shown below (errors below 15 on green, above 20 on red). Note that a similar score does not necessarily mean the same qualitative result, as shown by the two methods with green score. Three interesting effects are highlighted in the input image. We recommend the reader to zoom-in and analyze the results of the method at these specific areas. Image ID: \textit{quiltsalad\_3711222369} 
}
\label{fig:iiw}
\end{figure*}

\section{Discussion}\label{sec:discussion}

As shown in Sections~\ref{sec:learning_formulation} and~\ref{sec:evaluation}, recent methods are showing considerable advances in their ability to learn with non explicitly labeled datasets, purely computer graphics datasets, estimate extra scene elements (such as normals or depth), and take into account more complex light-surface interactions beyond Lambertian material models. Nevertheless, this progress is not clearly obvious by looking at quantitative metrics or qualitative results, often showing very different outcomes for similar error values. In this section, our aim is to discuss this problem as well as other factors which are making it challenging to objectively track advances in this field.

\subsection{Outperforming Learning Strategies}
\NEW{
As discussed in detail in Section~\ref{sec:learning_formulation}, there are several ways to use datasets and priors to train a neural network to address the intrinsic decomposition problem. In the following, we will discuss how the different strategies have been combined by the most successful methods according to Section~\ref{sec:evaluation}. In particular, to show generalization capabilities of the different training strategies we will center the discussion around the methods that show compelling results on the IIW dataset without using such data for training (see Table~\ref{tab:IIW_Error} $\dagger$). 
}

\NEW{
According to the WHDR error reported in Table~\ref{tab:IIW_Error}, the method of Bi~\etal~\cite{bi2018deep} is the most compelling one.
It combines a twofold strategy using full and self-supervision (single-image and multi-image), as well as synthetic data from the MPI Sintel and real data from a custom built multi-illuminant dataset. It further includes prior information for the albedo layer by means of a learned bilateral filter~\cite{barron2016fast}.
The method of Liu~\etal~\cite{liu2020unsupervised} is the second best with a score under twenty. Their approach is fully unsupervised and their key idea is to train latent feature spaces for a domain-invariant image-to-image translation architecture. Their models are trained using mostly synthetic images of ShapeNet, MPI Sintel, and CG dataset, and real from MIT Intrinsics.
}

\NEW{
The methods of Yu~\etal~\cite{yu2019inverserendernet} and Li~\etal~\cite{li2020inverse} which perform similarly also follow  a similar approach to encode and infer explicit illumination using spherical harmonics. The former by means of a single environment map, while the latter encodes it in a spatially-varying basis. The idea of encoding a spatially-varying material map is further followed by Li's which leverages a large synthetic dataset for training.  Yu's method is however trained to obtain a pure diffuse model using real scenes from MegaDepth dataset for which normals and albedo maps have been estimated using multi-view stereo. 
}
\NEW{
Finally, the method of Li~\etal~\cite{li2018watching}, while performing similarly to the above one, is the most different in terms of learning strategy. It relies on a multi-image consistency loss trained using timelapse sequences and smoothness priors for the shading and reflectance layers.
}

\NEW{
In light of these results, it is worth noting that the most successful strategy seems to be the combination of self-supervised learning along with fine-tuning on a specific dataset or domain. Using priors on the albedo layer by means of some form of bilateral filter appears to be also an interesting approach to reduce the amount of training data required. The main disadvantage of self-supervision for this problem is that the reconstruction loss needs to be computed efficiently during the progress of training, difficulting the use of global-illumination based reconstruction losses. Taking into account global-illumination effects requires the computation of multiple light bounces that is prohibitive in this context, which is the reason why the majority of the methods rely on direct illumination and lambertian materials models. As discussed in Section 9.3, a promising research direction in this area involves the use of differentiable rendering techniques. 
}

\subsection{Training and Testing Data}

It is well known that one of the reasons for the superior performance of CNNs is their capability of capturing local and global --semantic, object level-- features. CNNs are empowered by two main factors: translation invariance and hierarchical definition of image-level patterns. Although important for several computer vision applications, such as image classification or semantic segmentation, those properties might be problematic for the intrinsic decomposition problem. 

Semantic dependence during training impacts the generalization of the method to arbitrary scenes. 
First, the diversity of objects and materials of the scene is given by its semantics, so a network trained with a dataset only containing indoor scenes will not likely generalize to outdoor scenes. In other words, the prior distribution of scene components learned by the network will not cover the diversity of scenes that are present in the real world. 
A similar problem is faced with illumination, as the same object and material might have a vastly different appearance under different lighting conditions, so if the training data is not targeted at disentangling this relationship, the network will perform inadequately.
Such a limitation is common in most deep intrinsic decomposition models. Currently, many of the existing datasets, either synthetic (SUNCG, CGIntrinsics) or real (IIW, SAW), used for training and testing contain a majority of indoor scenes. 
The methods of Sengupta~\etal~\shortcite{sengupta2019neural} and Li~\etal~\shortcite{li2020inverse} account for this bias and limit the scope of the contributions to indoor scenes, as well as capturing an inverse model of indoor lighting through environment maps.

Testing on data outside the training set (without fine-tuning) is done by only a few methods: 
on the MIT dataset~\cite{baslamisli2018cnn}, and IIW dataset~\cite{bi2018deep,li2018watching,yu2019inverserendernet,liu2020unsupervised}.
As shown in Table~\ref{tab:method_dataset}, the majority of methods are both trained and tested using the same dataset. 
In Figure~\ref{fig:iiw}, we can appreciate better quantitative performance for methods both trained and tested on IIW. Nevertheless, the qualitative improvement is less clearly observable, suggesting that the quantitative evaluation done to those algorithms may not account for perceptual factors. We discuss this issue next.

\subsection{Quantitative Errors as Indicator of Performance}

Quantitative error metrics are the common way of measuring the performance of machine learning algorithms. 
\NEW{However, as explained in Section~\ref{sec:evaluation}, the intrinsic decomposition problem lacks properly established benchmarks that facilitate these comparisons. This is evidenced in Tables~\ref{tab:MIT_error} and \ref{tab:IIW_Error} where we can observe a diverse range of reported errors for the same method. There are two key factors that might be the cause of this discrepancy: first, choosing different splits for the train/test sets, and second, the selection of the reported error metric when the methods report several errors for different training conditions (e.g. with and without fine-tuning with a specific dataset).}

In the following discussion, we will focus on the IIW dataset and reported WHDR errors, shown in Table~\ref{tab:IIW_Error}, \NEW{as it contains the most consistent train/test splits and is the most common.} As it can be seen, there is a consistent trend towards small errors in most recent approaches that include complex inverse models or network architectures. However, a smaller error value does not necessarily correlate with a more physically accurate decomposition. Furthermore, similar error values do not necessarily imply similar decompositions, as we discuss next.%

In Figure~\ref{fig:iiw}, we show the intrinsic components of one of the scenes of the IIW dataset. We compare the performance among learning-based methods: 1) trained or fine-tuned with IIW human annotations, 2) methods only tested on that dataset, and 3) two baseline decompositions that do not require training data, first, using the luminance channel as shading component and, second, the decomposition provided by Bi \etal \shortcite{bi20151} which relied on a clustering-based strategy. 
In terms of quantitative metrics, it is shown that methods trained on such dataset perform slightly better. However, the differences between methods with similar error values are noticeable, for example, for methods trained on IIW, between Li's \cite{li2018cgintrinsics} and Fan's\cite{fan2018revisiting}, and between Yu's \cite{yu2019inverserendernet} and Li's\cite{li2018watching}. 
At the same time, it seems quite difficult to qualitatively judge which decomposition is more accurate. It could be argued that Li's\cite{li2018cgintrinsics}, Bi's \cite{bi20151} and Liu's \cite{liu2020unsupervised} produce the best results as the shading images have less albedo remnants. 
Bonneel \etal~\cite{bonneel2017intrinsic} already exposed this problem and presented a thorough evaluation of intrinsic decomposition methods in the context of image editing tasks. In their study, none of the existing methods was robust enough to be used for such kind of operations.

\subsection{\NEW{The Influence of the Material Model}}

For a long time, intrinsic decomposition methods assumed ~Lambertian material models, a design decision which impacted also the variability of the datasets used to train and evaluate the methods. As we have shown in Section~\ref{sec:theory}, this significantly simplifies the image formation model, making the inverse problem tractable from a practical standpoint.
However, the use of these models and datasets have had two main problems: First, as \textit{the model is not rich enough}, it will not be able to inversely reproduce the scene correctly. 
For example, looking at the world around us, it is easy to discover a majority of non-Lambertian surfaces: glasses, plastics, or cloth, among others. 
Second, if \textit{the data used for training does not contain enough variability and realism}, it will be hard to predict how the methods will behave on real scenes containing a broad range of material types. 

\NEW{Only a few of the latest methods on intrinsic decomposition have incorporated more complex material models. Meka~\etal~\cite{meka2018lime} estimate the parameters of a Blinn-Phong material model, however, as it just deals with single isolated objects, a fair comparison is unfeasible.
The two methods that deal with general scenes and complex materials are Sengupta~\etal~\cite{sengupta2019neural} that incorporate the specular lobe of Phong shader, and the method of Li~\etal~\cite{li2020inverse} that goes a step further and estimates the parameters of a spatially-varying microfacet BRDF.
It is worth mentioning that the latter performs reasonably well on the IIW dataset after fine-tuning, and even provides the error estimation for a model which wasn’t fine-tuned on such a dataset, suggesting a positive trend in attempting to generalize to new scenes. 
In Figure 9, we observe that while the reflectance layer looks coherent, the shading layer tends to have artifacts due to the increased complexity of having to estimate an underlying geometry of the scene in the form of normals and lighting. These extra layers (normals and lighting) are nevertheless beneficial to improve the editability of the scene, for example, to change illumination or material properties.}

When a specular image is tested with a method that assumes a Lambertian model, the
specular highlight will be wrongly placed either completely or partially in either component, because the intrinsic diffuse model does not consider such a property of the materials. We can see an example of such effect in the television highlight of Figure~\ref{fig:iiw}. Note that all the methods shown in that figure, except Li~\etal~\cite{li2020inverse}, follow the intrinsic diffuse model.
At the same time, the Lambertian assumption and the \textit{classical} intrinsic decomposition framework has yielded a set of explicitly labeled datasets which describe a limited set of Lambertian materials~\cite{grosse2009ground,li2018cgintrinsics}.

\section{Research Opportunities and Future Directions}\label{sec:future}

Deep learning has changed the way the intrinsic decomposition problem is addressed, transitioning from sometimes \textit{manual} or ad-hoc heuristics to implicit rules defined by the data. Among the data-driven methods reviewed, we have also observed a trend, from simple regression-based approaches to recent methods that take into account the physics of light, for example, modeling complex scene elements: geometry through normals, lighting using spherical harmonics, or non-lambertian material models using microfacets. Nevertheless, the world is highly complex and the problem of fully reconstructing arbitrary scenes from single images is still in an open research area. In the following, we discuss, among other topics, how the field can benefit from novel machine learning techniques learn more efficiently from data, initiatives that promote replicability and facilitate comparative evaluation frameworks, as well as the potential of differentiable and neural rendering.

\subsection{Enhancing Generalization}

Addressing the problem of intrinsic image decomposition with a purely data-driven approach is particularly risky: the content of the dataset guides the variability and amount of object-material and light-material phenomena that a model learns to disambiguate. 
Some approaches have focused on indoor scenes \cite{bell2014intrinsic,li2018cgintrinsics,sengupta2019neural} dominated by furniture, decorative objects, and certain types of lighting; and others have limited the scope to a subset of objects within the ShapeNet dataset~\cite{chang2015shapenet}. 
Recently, domain specific datasets have been created for outdoor road scenes~\cite{krahenbuhl2018free}, or planar materials \cite{dong2019deep}. In contrast to the brute force approach of training a massive network able to generalize blindly for each domain (akin to language models~\cite{brown2020language}), a more environment-friendly and scalable approach would include combining the variety of domain-specific network in a single framework without needing to re-train the whole system, \eg, pre-classifying the scenes to decide the best networks from an ensemble, defining accessible APIs, or using networks designed for cross-domain generalization, as proposed by Rebuffi~\etal~\cite{rebuffi2017learning}. 

Most intrinsic decomposition datasets are smaller than many popular datasets used in the computer vision literature, such as ImageNet~\cite{deng2009imagenet} or COCO~\cite{lin2014microsoft}. The patterns found by early layers in CNNs trained on ImageNet have been found to generalize to multiple computer vision problems, which could benefit deep learning models applied to the intrinsic decomposition problems. Pre-training a CNN on ImageNet using recent unsupervised~\cite{gidaris2018unsupervised} or self-supervised learning~\cite{chen2020simple} approaches, then fine-tune such network for the intrinsic decomposition problem, could provide a solution for the lack of sufficient training data in the datasets we discussed.

Besides, generative adversarial networks have proven successful at generating photo-realistic images in many domains \cite{karras2020analyzing}. Recent work on neural rendering and inverse graphics \cite{zhang2020image} suggests that those generative models learn representations in which geometry, light and texture can be disentangled. Such approaches could be used to generate new synthetic data samples that allow for larger intrinsic decomposition datasets. Furthermore, some data points are more informative than others. Active learning methods \cite{brust2018active,wang2016cost} could help efficiently generate new (either synthetic or real) samples that help intrinsic decomposition methods improve their generalization capabilities. 

Additionally, most neural network architectures used for intrinsic decomposition are CNNs. Recent work on attention mechanisms indicate that traditional convolutional layers, while powerful, may be limiting the potential of deep learning for computer vision problems~\cite{dosovitskiy2020image,wang2018non,jetley2018learn}. Moving beyond traditional deep CNN architectures may provide intrisic decomposition algorithms with more sophisticated inductive biases.

\subsection{Evaluation Frameworks} 

As we have seen in the previous section, existing evaluation metrics~\cite{bell2014intrinsic,grosse2009ground} are not necessarily representative of the complexity of the real world, neither capture consistently the actual accuracy and consistency of the decomposition results (see Figure~\ref{fig:iiw}). 
A particularly interesting opportunity in this field would include creating a common evaluation framework (a benchmark, or a challenge), following existing initiatives in the computer vision and graphics communities~\cite{Erofeev2015,rhemann2009perceptually,robustvision2020,merzbach2020bonn}. 

Creating an evaluation dataset and metrics for the intrinsic decomposition problem is challenging, as the intrinsic components do not freely exist in the nature. We propose several ideas to facilitate this process. 
First, leverage hyper-realistic computer generated scenes, like existing works~\cite{li2018cgintrinsics,bonneel2017intrinsic,sengupta2019neural} but shaping the data in the form of a benchmark so that it is easily accessible, and comparable. The main limitations in this case are the render engine not being able to reproduce physically complex light phenomena, the cost of manually creating a variety of scenes of several semantics, as well the required render time. 
Second, exploit scene characteristics such as the albedo invariance to illumination changes. This can be done by leveraging time-lapse or multi-view scenes (~\cite{li2018megadepth,li2018watching}) which are more easily accessible, so that the final evaluation should take into account the quality of the reconstruction as well as the invariance of the albedo layer (mimicking the multi-view learning strategy presented in Section~\ref{sec:self}).
Finally, an ideal dataset for a benchmark would also include: a variety of scene semantics (indoor, outdoors, single objects, humans, etc.), the same scene with multiple illuminations, and complex light phenomena useful to understand the path of light. Those characteristics might also serve as intermediate low-level step towards a higher-level task, \textit{e.g.}, specular layers, scattering effects, a layer for caustics, or a layer for shadows or occlusion boundaries. Having these layers would be useful for many other tasks, such as material editing, object segmentation, or scene compositing.

\subsection{Beyond Lambert} 
\label{subsection:complex_material-lighting}

We have observed in recent works an increasing use of complex illumination models \cite{yu2019inverserendernet,sengupta2019neural,zhou2019glosh,li2020inverse}, learning from synthetic datasets that contain both global illumination and inter-reflections, and some relying on the \textit{intrinsic residual} model to enable spatially varying light effects. 
In these methods, light transport simulation has been restricted to real time techniques like directional lights and spherical harmonics environments, with limitations in realism at high frequency shadows, occlusions and inter-reflections.

The advent of differentiable ray tracing methods \cite{Li2018DMC,Azinovic2019,nimier2019mitsuba}, have opened a new opportunity to include accurate light transport path simulations, not only as a costly part of the error function in the learning process, but also as a neural network component. Also, the success of the neural networks to learn parts of the render equation~\cite{janner2017self,sengupta2019neural,li2020inverse,zhou2019glosh} have a great potential to encapsulate complex lighting interactions, as shown by neural rendering techniques~\cite{tewari2020starneural}.

The definition of reflectance has also reached increasing levels of sophistication, the material being no longer just a base albedo color, but also including specular or Phong coefficients in more recent methods\cite{meka2018lime,li2020inverse}.
However, there is great potential for improvement. The success in estimating physically-based material models for flat surfaces~\cite{deschaintre2018single} or objects~\cite{li2018learning} suggest an interesting avenue for future work in such direction in order to generalize such findings to arbitrary scenes.
Without a full BSDF model~\cite{vidaurre_wacv2019} and its multiple parameters linked to the reflectance layer, some intrinsic decomposition applications such as relighting will never produce realistic results. For instance, the human relighting technique by Kanamori et al. \cite{kanamori2018relighting} relies on advanced illumination (spherical harmonics and local occlussion) but fails to produce realistic relighting because their material model lacks transmittance and subsurface parameters which are paramount for the final appearance of human skin and cloth.

We have barely mentioned several optical phenomena in this survey: participating media, caustics, subsurface scattering (highly relevant in many materials such as skin, cloth or liquids) \cite{jensen2001practical}, energy transfer between wavelengths (re-radiance, fluorescence), polarization, interference (iridescence), etc. Although we have seen impressive results in the recent years for problems such as inferring the scattering parameters an density of volumetric media \cite{nimier2019mitsuba}, many of these effects will require further advances of the inverse rendering field to be applied to intrinsic imaging.

The challenges ahead include finding an adequate balance between inverse and neural rendering techniques, and the right distribution of parameters into multiple intrinsic layers for each desired application. For instance, editing applications will required more physically-principled decompositions through inverse rendering, so more parameters are modifiable. On the other side, constrained parts of the problem, like restricted geometries (faces, bodies), partial light transport paths, or material shaders, will surely benefit from neural rendering strategies.

\section{Conclusions}

Deep learning has changed the way the intrinsic decomposition problem is addressed, transitioning from sometimes \textit{manual} or ad-hoc heuristics to implicit rules defined by the data. In this survey, we have reviewed this transition discussing learning frameworks, architectures, and datasets used, putting them in the context of traditional non-learning-based solutions. Through this revision, we have identified several problems that might prevent the field to develop further: the semantic dependence on the training data, the uncorrelation between current evaluation metrics and qualitative results, and the limitations of the widely used Lambertian material model to capture complex materials and light phenomena.

In light of the recent advances in neural rendering and differentiable rendering, we also believe that the intrinsic decomposition would greatly benefit from a physically-based --inverse rendering-- perspective. For this reason, in this survey, we also provide a thorough explanation of the physics of light from a rendering and inverse rendering perspective, as well as make explicit connections with other inverse methods that deal with specific domains such as faces, humans, flat materials, or objects.

\paragraph{\textbf{Acknowledgments}} Elena Garces was partially supported by a Torres Quevedo Fellowship (PTQ2018-009868). The work was also funded in part by the Spanish
Ministry of Science (RTI2018-098694-B-I00 VizLearning).

\bibliographystyle{spmpsci}      
\bibliography{intrinsic}   

\begin{thebibliography}{100}
\providecommand{\url}[1]{{#1}}
\providecommand{\urlprefix}{URL }
\expandafter\ifx\csname urlstyle\endcsname\relax
  \providecommand{\doi}[1]{DOI~\discretionary{}{}{}#1}\else
  \providecommand{\doi}{DOI~\discretionary{}{}{}\begingroup
  \urlstyle{rm}\Url}\fi

\bibitem{robustvision2020}
{Robust Vision Challenge 2020}.
\newblock Accessed: 2020-10-31

\bibitem{achanta2012slic}
Achanta, R., Shaji, A., Smith, K., Lucchi, A., Fua, P., S{\"u}sstrunk, S.: Slic
  superpixels compared to state-of-the-art superpixel methods.
\newblock IEEE Transactions on Pattern Analysis and Machine Intelligence
  \textbf{34}(11), 2274--2282 (2012)

\bibitem{Azinovic2019}
Azinovic, D., Li, T.M., Kaplanyan, A., Niessner, M.: Inverse path tracing for
  joint material and lighting estimation.
\newblock In: Proceedings of the IEEE Conference on Computer Vision and Pattern
  Recognition (2019)

\bibitem{Balzer2011}
Balzer, J., Höfer, S., Beyerer, J.: Multiview specular stereo reconstruction
  of large mirror surfaces.
\newblock In: CVPR 2011, pp. 2537--2544 (2011).
\newblock \doi{10.1109/CVPR.2011.5995346}

\bibitem{barron2014shape}
Barron, J.T., Malik, J.: Shape, illumination, and reflectance from shading.
\newblock IEEE Transactions on Pattern Analysis and Machine Intelligence
  \textbf{37}(8), 1670--1687 (2014)

\bibitem{barron2016fast}
Barron, J.T., Poole, B.: The fast bilateral solver.
\newblock In: European Conference on Computer Vision, pp. 617--632. Springer
  (2016)

\bibitem{barrow1978recovering}
Barrow, H., Tenenbaum, J., Hanson, A., Riseman, E.: Recovering intrinsic scene
  characteristics.
\newblock Computer Vision Systems \textbf{2}(3-26), 2 (1978)

\bibitem{baslamisli2018cnn}
Baslamisli, A.S., Le, H.A., Gevers, T.: Cnn based learning using reflection and
  retinex models for intrinsic image decomposition.
\newblock In: Proceedings of the IEEE Conference on Computer Vision and Pattern
  Recognition, pp. 6674--6683 (2018)

\bibitem{beigpour2011object}
Beigpour, S., Van De~Weijer, J.: Object recoloring based on intrinsic image
  estimation.
\newblock In: Proceedings of the IEEE International Conference on Computer
  Vision, pp. 327--334. IEEE (2011)

\bibitem{bell2014intrinsic}
Bell, S., Bala, K., Snavely, N.: Intrinsic images in the wild.
\newblock ACM Transactions on Graphics (TOG) \textbf{33}(4), 1--12 (2014)

\bibitem{bi20151}
Bi, S., Han, X., Yu, Y.: An l 1 image transform for edge-preserving smoothing
  and scene-level intrinsic decomposition.
\newblock ACM Transactions on Graphics (TOG) \textbf{34}(4), 1--12 (2015)

\bibitem{bi2018deep}
Bi, S., Kalantari, N.K., Ramamoorthi, R.: Deep hybrid real and synthetic
  training for intrinsic decomposition.
\newblock Computer Graphics Forum (Proc.~Eurographics Symposium on Rendering)
  (2018)

\bibitem{blanz1999morphable}
Blanz, V., Vetter, T.: A morphable model for the synthesis of 3d faces.
\newblock In: Proceedings of the 26th Annual Conference on Computer Graphics
  and Interactive Techniques, pp. 187--194 (1999)

\bibitem{bonneel2017intrinsic}
Bonneel, N., Kovacs, B., Paris, S., Bala, K.: Intrinsic decompositions for
  image editing.
\newblock In: Computer Graphics Forum (Proc.~Eurographics STAR), vol.~36, pp.
  593--609 (2017)

\bibitem{bousseau2009user}
Bousseau, A., Paris, S., Durand, F.: User-assisted intrinsic images.
\newblock In: Proceedings of the 2009 SIGGRAPH Asia Conference, pp. 1--10
  (2009)

\bibitem{brown2020language}
Brown, T.B., Mann, B., Ryder, N., Subbiah, M., Kaplan, J., Dhariwal, P.,
  Neelakantan, A., Shyam, P., Sastry, G., Askell, A., Agarwal, S.,
  Herbert-Voss, A., Krueger, G., Henighan, T., Child, R., Ramesh, A., Ziegler,
  D.M., Wu, J., Winter, C., Hesse, C., Chen, M., Sigler, E., Litwin, M., Gray,
  S., Chess, B., Clark, J., Berner, C., McCandlish, S., Radford, A., Sutskever,
  I., Amodei, D.: Language models are few-shot learners.
\newblock arXiv preprint arXiv:2005.14165  (2020)

\bibitem{brust2018active}
Brust, C.A., K{\"a}ding, C., Denzler, J.: Active learning for deep object
  detection.
\newblock arXiv preprint arXiv:1809.09875  (2018)

\bibitem{butler2012naturalistic}
Butler, D.J., Wulff, J., Stanley, G.B., Black, M.J.: A naturalistic open source
  movie for optical flow evaluation.
\newblock In: Proceedings of the European Conference on Computer Vision (ECCV),
  pp. 611--625. Springer (2012)

\bibitem{carroll2011illumination}
Carroll, R., Ramamoorthi, R., Agrawala, M.: Illumination decomposition for
  material recoloring with consistent interreflections.
\newblock In: ACM SIGGRAPH 2011 papers, pp. 1--10 (2011)

\bibitem{chaitanya2017interactive}
Chaitanya, C.R.A., Kaplanyan, A.S., Schied, C., Salvi, M., Lefohn, A.,
  Nowrouzezahrai, D., Aila, T.: Interactive reconstruction of monte carlo image
  sequences using a recurrent denoising autoencoder.
\newblock ACM Transactions on Graphics (TOG) \textbf{36}(4), 1--12 (2017)

\bibitem{chang2015shapenet}
Chang, A.X., Funkhouser, T., Guibas, L., Hanrahan, P., Huang, Q., Li, Z.,
  Savarese, S., Savva, M., Song, S., Su, H., et~al.: Shapenet: An
  information-rich 3d model repository.
\newblock arXiv preprint arXiv:1512.03012  (2015)

\bibitem{chang2014bayesian}
Chang, J., Cabezas, R., Fisher, J.W.: Bayesian nonparametric intrinsic image
  decomposition.
\newblock In: Proceedings of the European Conference on Computer Vision (ECCV),
  pp. 704--719. Springer (2014)

\bibitem{chen2013simple}
Chen, Q., Koltun, V.: A simple model for intrinsic image decomposition with
  depth cues.
\newblock In: Proceedings of the IEEE International Conference on Computer
  Vision, pp. 241--248 (2013)

\bibitem{chen2020simple}
Chen, T., Kornblith, S., Norouzi, M., Hinton, G.: A simple framework for
  contrastive learning of visual representations.
\newblock arXiv preprint arXiv:2002.05709  (2020)

\bibitem{cheng2018intrinsic}
Cheng, L., Zhang, C., Liao, Z.: Intrinsic image transformation via scale space
  decomposition.
\newblock In: Proceedings of the IEEE Conference on Computer Vision and Pattern
  Recognition, pp. 656--665 (2018)

\bibitem{deng2009imagenet}
Deng, J., Dong, W., Socher, R., Li, L.J., Li, K., Fei-Fei, L.: Imagenet: A
  large-scale hierarchical image database.
\newblock In: Proceedings of the IEEE Conference on Computer Vision and Pattern
  Recognition, pp. 248--255. Ieee (2009)

\bibitem{deschaintre2018single}
Deschaintre, V., Aittala, M., Durand, F., Drettakis, G., Bousseau, A.:
  Single-image svbrdf capture with a rendering-aware deep network.
\newblock ACM Transactions on Graphics (ToG) \textbf{37}(4), 1--15 (2018)

\bibitem{deschaintre2019flexible}
Deschaintre, V., Aittala, M., Durand, F., Drettakis, G., Bousseau, A.: Flexible
  svbrdf capture with a multi-image deep network.
\newblock In: Computer Graphics Forum, vol.~38, pp. 1--13. Wiley Online Library
  (2019)

\bibitem{DDB20}
Deschaintre, V., Drettakis, G., Bousseau, A.: Guided fine-tuning for
  large-scale material transfer.
\newblock Computer Graphics Forum (Proceedings of the Eurographics Symposium on
  Rendering) \textbf{39}(4) (2020).
\newblock \urlprefix\url{http://www-sop.inria.fr/reves/Basilic/2020/DDB20}

\bibitem{Bo2015}
Dong, B., Dong, Y., Tong, X., Peers, P.: Measurement-based editing of diffuse
  albedo with consistent interreflections.
\newblock ACM Trans. Graph. \textbf{34}(4) (2015)

\bibitem{dong2014scattering}
Dong, B., Moore, K.D., Zhang, W., Peers, P.: Scattering parameters and surface
  normals from homogeneous translucent materials using photometric stereo.
\newblock In: Proceedings of the IEEE Conference on Computer Vision and Pattern
  Recognition, pp. 2291--2298 (2014)

\bibitem{dong2019deep}
Dong, Y.: Deep appearance modeling: A survey.
\newblock Visual Informatics \textbf{3}(2), 59--68 (2019)

\bibitem{dong2011appgen}
Dong, Y., Tong, X., Pellacini, F., Guo, B.: Appgen: interactive material
  modeling from a single image.
\newblock In: Proceedings of the 2011 SIGGRAPH Asia Conference, pp. 1--10
  (2011)

\bibitem{dosovitskiy2020image}
Dosovitskiy, A., Beyer, L., Kolesnikov, A., Weissenborn, D., Zhai, X.,
  Unterthiner, T., Dehghani, M., Minderer, M., Heigold, G., Gelly, S., et~al.:
  An image is worth 16x16 words: Transformers for image recognition at scale.
\newblock arXiv preprint arXiv:2010.11929  (2020)

\bibitem{duchene2015}
Duch\^{e}ne, S., Riant, C., Chaurasia, G., Moreno, J.L., Laffont, P.Y., Popov,
  S., Bousseau, A., Drettakis, G.: Multiview intrinsic images of outdoors
  scenes with an application to relighting.
\newblock ACM Transactions on Graphics (TOG) \textbf{34}(5) (2015)

\bibitem{Erofeev2015}
Erofeev, M., Gitman, Y., Vatolin, D., Fedorov, A., Wang, J.: Perceptually
  motivated benchmark for video matting.
\newblock In: Proceedings of the British Machine Vision Conference (BMVC), pp.
  99.1--99.12. BMVA Press (2015)

\bibitem{fan2018revisiting}
Fan, Q., Yang, J., Hua, G., Chen, B., Wipf, D.: Revisiting deep intrinsic image
  decompositions.
\newblock In: Proceedings of the IEEE conference on computer vision and pattern
  recognition, pp. 8944--8952 (2018)

\bibitem{finlayson2004intrinsic}
Finlayson, G.D., Drew, M.S., Lu, C.: Intrinsic images by entropy minimization.
\newblock In: Proceedings of the European Conference on Computer Vision (ECCV),
  pp. 582--595. Springer (2004)

\bibitem{gao2019deep}
Gao, D., Li, X., Dong, Y., Peers, P., Xu, K., Tong, X.: Deep inverse rendering
  for high-resolution svbrdf estimation from an arbitrary number of images.
\newblock ACM Transactions on Graphics (TOG) \textbf{38}(4), 1--15 (2019)

\bibitem{garces2012intrinsic}
Garces, E., Munoz, A., Lopez-Moreno, J., Gutierrez, D.: Intrinsic images by
  clustering.
\newblock In: Computer Graphics Forum, vol.~31, pp. 1415--1424 (2012)

\bibitem{gastal2012adaptive}
Gastal, E.S., Oliveira, M.M.: Adaptive manifolds for real-time high-dimensional
  filtering.
\newblock ACM Transactions on Graphics (TOG) \textbf{31}(4), 1--13 (2012)

\bibitem{gatys2017controlling}
Gatys, L.A., Ecker, A.S., Bethge, M., Hertzmann, A., Shechtman, E.: Controlling
  perceptual factors in neural style transfer.
\newblock In: Proceedings of the IEEE Conference on Computer Vision and Pattern
  Recognition, pp. 3985--3993 (2017)

\bibitem{gidaris2018unsupervised}
Gidaris, S., Singh, P., Komodakis, N.: Unsupervised representation learning by
  predicting image rotations.
\newblock arXiv preprint arXiv:1803.07728  (2018)

\bibitem{Godard2015}
Godard, C., Hedman, P., Li, W., Gabriel, J.: Multi-view reconstruction of
  highly specular surfaces in uncontrolled environments.
\newblock In: 3DV (2015).
\newblock \doi{10.1109/3DV.2015.10}

\bibitem{grosse2009ground}
Grosse, R., Johnson, M.K., Adelson, E.H., Freeman, W.T.: Ground truth dataset
  and baseline evaluations for intrinsic image algorithms.
\newblock In: Proceedings of the IEEE International Conference on Computer
  Vision, pp. 2335--2342. IEEE (2009)

\bibitem{Guarnera2016}
Guarnera, D., Guarnera, G., Ghosh, A., Denk, C., Glencross, M.: Brdf
  representation and acquisition.
\newblock Computer Graphics Forum \textbf{35}(2), 625--650 (2016)

\bibitem{Guo:2020:MaterialGAN}
Guo, Y., Smith, C., Ha\v{s}an, M., Sunkavalli, K., Zhao, S.: Materialgan:
  Reflectance capture using a generative svbrdf model.
\newblock ACM Trans. Graph. \textbf{39}(6), 254:1--254:13 (2020)

\bibitem{han2019image}
Han, X., Laga, H., Bennamoun, M.: Image-based 3d object reconstruction:
  State-of-the-art and trends in the deep learning era.
\newblock IEEE Transactions on Pattern Analysis and Machine Intelligence
  (2019)

\bibitem{he2016deep}
He, K., Zhang, X., Ren, S., Sun, J.: Deep residual learning for image
  recognition.
\newblock In: Proceedings of the IEEE Conference on Computer Vision and Pattern
  Recognition, pp. 770--778 (2016)

\bibitem{SigPBR015}
Hill, S., McAuley, S., Burley, B., Chan, D., Fascione, L., Iwanicki, M.,
  Hoffman, N., Jakob, W., Neubelt, D., Pesce, A., Pettineo, M.: Physically
  based shading in theory and practice.
\newblock In: ACM SIGGRAPH 2015 Courses, SIGGRAPH '15. Association for
  Computing Machinery, New York, NY, USA (2015)

\bibitem{horn1974determining}
Horn, B.K.: Determining lightness from an image.
\newblock Computer graphics and image processing \textbf{3}(4), 277--299 (1974)

\bibitem{horn1979calculating}
Horn, B.K., Sjoberg, R.W.: Calculating the reflectance map.
\newblock Applied optics \textbf{18}(11), 1770--1779 (1979)

\bibitem{huang2017arbitrary}
Huang, X., Belongie, S.: Arbitrary style transfer in real-time with adaptive
  instance normalization.
\newblock In: Proceedings of the IEEE International Conference on Computer
  Vision, pp. 1501--1510 (2017)

\bibitem{isola2017image}
Isola, P., Zhu, J.Y., Zhou, T., Efros, A.A.: Image-to-image translation with
  conditional adversarial networks.
\newblock In: Proceedings of the IEEE Conference on Computer Vision and Pattern
  Recognition, pp. 1125--1134 (2017)

\bibitem{jakob2010mitsuba}
Jakob, W.: Mitsuba renderer (2010)

\bibitem{janner2017self}
Janner, M., Wu, J., Kulkarni, T.D., Yildirim, I., Tenenbaum, J.:
  Self-supervised intrinsic image decomposition.
\newblock In: Advances in Neural Information Processing Systems, pp. 5936--5946
  (2017)

\bibitem{jensen2001practical}
Jensen, H.W., Marschner, S.R., Levoy, M., Hanrahan, P.: A practical model for
  subsurface light transport.
\newblock In: Proceedings of the 28th annual conference on Computer graphics
  and interactive techniques, pp. 511--518 (2001)

\bibitem{jetley2018learn}
Jetley, S., Lord, N.A., Lee, N., Torr, P.H.: Learn to pay attention.
\newblock arXiv preprint arXiv:1804.02391  (2018)

\bibitem{johnson2016perceptual}
Johnson, J., Alahi, A., Fei-Fei, L.: Perceptual losses for real-time style
  transfer and super-resolution.
\newblock In: Proceedings of the European Conference on Computer Vision (ECCV),
  pp. 694--711. Springer (2016)

\bibitem{kajiya1986rendering}
Kajiya, J.T.: The rendering equation.
\newblock In: Proceedings of the 13th annual conference on Computer graphics
  and interactive techniques, pp. 143--150 (1986)

\bibitem{kanamori2018relighting}
Kanamori, Y., Endo, Y.: Relighting humans: occlusion-aware inverse rendering
  for full-body human images.
\newblock ACM Transactions on Graphics (TOG) \textbf{37}(6), 1--11 (2018)

\bibitem{karis2013real}
Karis, B., Games, E.: Real shading in unreal engine 4.
\newblock Proc. Physically Based Shading Theory Practice \textbf{4}, 3 (2013)

\bibitem{karras2020analyzing}
Karras, T., Laine, S., Aittala, M., Hellsten, J., Lehtinen, J., Aila, T.:
  Analyzing and improving the image quality of stylegan.
\newblock In: Proceedings of the IEEE Conference on Computer Vision and Pattern
  Recognition, pp. 8110--8119 (2020)

\bibitem{kovacs2017shading}
Kovacs, B., Bell, S., Snavely, N., Bala, K.: Shading annotations in the wild.
\newblock In: Proceedings of the IEEE Conference on Computer Vision and Pattern
  Recognition, pp. 6998--7007 (2017)

\bibitem{krahenbuhl2018free}
Kr{\"a}henb{\"u}hl, P.: Free supervision from video games.
\newblock In: Proceedings of the IEEE Conference on Computer Vision and Pattern
  Recognition, pp. 2955--2964 (2018)

\bibitem{krahenbuhl2011efficient}
Kr{\"a}henb{\"u}hl, P., Koltun, V.: Efficient inference in fully connected crfs
  with gaussian edge potentials.
\newblock In: Advances in Neural Information Processing Systems, pp. 109--117
  (2011)

\bibitem{krizhevsky2012imagenet}
Krizhevsky, A., Sutskever, I., Hinton, G.E.: Imagenet classification with deep
  convolutional neural networks.
\newblock In: Advances in Neural Information Processing Systems, pp. 1097--1105
  (2012)

\bibitem{laffont2015intrinsic}
Laffont, P.Y., Bazin, J.C.: Intrinsic decomposition of image sequences from
  local temporal variations.
\newblock In: Proceedings of the IEEE International Conference on Computer
  Vision, pp. 433--441 (2015)

\bibitem{laffont2012coherent}
Laffont, P.Y., Bousseau, A., Paris, S., Durand, F., Drettakis, G.: Coherent
  intrinsic images from photo collections.
\newblock ACM Transactions on Graphics (TOG) \textbf{31}(6), 1--11 (2012)

\bibitem{lafortune1994using}
Lafortune, E.P., Willems, Y.D.: Using the modified phong reflectance model for
  physically based rendering  (1994)

\bibitem{land1971lightness}
Land, E.H., McCann, J.J.: Lightness and retinex theory.
\newblock Josa \textbf{61}(1), 1--11 (1971)

\bibitem{lecun1989backpropagation}
LeCun, Y., Boser, B., Denker, J.S., Henderson, D., Howard, R.E., Hubbard, W.,
  Jackel, L.D.: Backpropagation applied to handwritten zip code recognition.
\newblock Neural computation \textbf{1}(4), 541--551 (1989)

\bibitem{lecun1998gradient}
LeCun, Y., Bottou, L., Bengio, Y., Haffner, P.: Gradient-based learning applied
  to document recognition.
\newblock Proceedings of the IEEE \textbf{86}(11), 2278--2324 (1998)

\bibitem{lettry2018darn}
Lettry, L., Vanhoey, K., Van~Gool, L.: Darn: a deep adversarial residual
  network for intrinsic image decomposition.
\newblock In: 2018 IEEE Winter Conference on Applications of Computer Vision
  (WACV), pp. 1359--1367. IEEE (2018)

\bibitem{Li2018DMC}
Li, T.M., Aittala, M., Durand, F., Lehtinen, J.: Differentiable monte carlo ray
  tracing through edge sampling.
\newblock ACM Trans. Graph. (Proc. SIGGRAPH Asia) \textbf{37}(6), 222:1--222:11
  (2018)

\bibitem{li2017modeling}
Li, X., Dong, Y., Peers, P., Tong, X.: Modeling surface appearance from a
  single photograph using self-augmented convolutional neural networks.
\newblock ACM Transactions on Graphics (ToG) \textbf{36}(4), 1--11 (2017)

\bibitem{li2020inverse}
Li, Z., Shafiei, M., Ramamoorthi, R., Sunkavalli, K., Chandraker, M.: Inverse
  rendering for complex indoor scenes: Shape, spatially-varying lighting and
  svbrdf from a single image.
\newblock In: Proceedings of the IEEE Conference on Computer Vision and Pattern
  Recognition, pp. 2475--2484 (2020)

\bibitem{li2018cgintrinsics}
Li, Z., Snavely, N.: Cgintrinsics: Better intrinsic image decomposition through
  physically-based rendering.
\newblock In: Proceedings of the European Conference on Computer Vision (ECCV),
  pp. 371--387 (2018)

\bibitem{li2018watching}
Li, Z., Snavely, N.: Learning intrinsic image decomposition from watching the
  world.
\newblock In: Proceedings of the IEEE Conference on Computer Vision and Pattern
  Recognition, pp. 9039--9048 (2018)

\bibitem{li2018megadepth}
Li, Z., Snavely, N.: Megadepth: Learning single-view depth prediction from
  internet photos.
\newblock In: Proceedings of the IEEE Conference on Computer Vision and Pattern
  Recognition, pp. 2041--2050 (2018)

\bibitem{li2018materials}
Li, Z., Sunkavalli, K., Chandraker, M.: Materials for masses: Svbrdf
  acquisition with a single mobile phone image.
\newblock In: Proceedings of the European Conference on Computer Vision (ECCV),
  pp. 72--87 (2018)

\bibitem{li2018learning}
Li, Z., Xu, Z., Ramamoorthi, R., Sunkavalli, K., Chandraker, M.: Learning to
  reconstruct shape and spatially-varying reflectance from a single image.
\newblock ACM Transactions on Graphics (TOG) \textbf{37}(6), 1--11 (2018)

\bibitem{lin2014microsoft}
Lin, T.Y., Maire, M., Belongie, S., Hays, J., Perona, P., Ramanan, D.,
  Doll{\'a}r, P., Zitnick, C.L.: Microsoft coco: Common objects in context.
\newblock In: Proceedings of the European Conference on Computer Vision (ECCV),
  pp. 740--755. Springer (2014)

\bibitem{liu2020unsupervised}
Liu, Y., Li, Y., You, S., Lu, F.: Unsupervised learning for intrinsic image
  decomposition from a single image.
\newblock In: Proceedings of the IEEE Conference on Computer Vision and Pattern
  Recognition (2020)

\bibitem{loubet2019reparameterizing}
Loubet, G., Holzschuch, N., Jakob, W.: Reparameterizing discontinuous
  integrands for differentiable rendering.
\newblock ACM Transactions on Graphics (TOG) \textbf{38}(6), 1--14 (2019)

\bibitem{ma2018single}
Ma, W.C., Chu, H., Zhou, B., Urtasun, R., Torralba, A.: Single image intrinsic
  decomposition without a single intrinsic image.
\newblock In: Proceedings of the European Conference on Computer Vision (ECCV),
  pp. 201--217 (2018)

\bibitem{martin2020nerf}
Martin-Brualla, R., Radwan, N., Sajjadi, M.S., Barron, J.T., Dosovitskiy, A.,
  Duckworth, D.: Nerf in the wild: Neural radiance fields for unconstrained
  photo collections.
\newblock arXiv preprint arXiv:2008.02268  (2020)

\bibitem{maxwell2008bi}
Maxwell, B.A., Friedhoff, R.M., Smith, C.A.: A bi-illuminant dichromatic
  reflection model for understanding images.
\newblock In: 2008 IEEE Conference on Computer Vision and Pattern Recognition,
  pp. 1--8. IEEE (2008)

\bibitem{meka2019deepreflectance}
Meka, A., H\"{a}ne, C., Pandey, R., Zollh\"{o}fer, M., Fanello, S., Fyffe, G.,
  Kowdle, A., Yu, X., Busch, J., Dourgarian, J., Denny, P., Bouaziz, S.,
  Lincoln, P., Whalen, M., Harvey, G., Taylor, J., Izadi, S., Tagliasacchi, A.,
  Debevec, P., Theobalt, C., Valentin, J., Rhemann, C.: Deep reflectance
  fields: High-quality facial reflectance field inference from color gradient
  illumination.
\newblock ACM Transactions on Graphics (TOG) \textbf{38}(4) (2019)

\bibitem{meka2018lime}
Meka, A., Maximov, M., Zollhoefer, M., Chatterjee, A., Seidel, H.P., Richardt,
  C., Theobalt, C.: Lime: Live intrinsic material estimation.
\newblock In: Proceedings of the IEEE Conference on Computer Vision and Pattern
  Recognition, pp. 6315--6324 (2018)

\bibitem{Meka2016}
Meka, A., Zollh\"{o}fer, M., Richardt, C., Theobalt, C.: Live intrinsic video.
\newblock ACM Trans. Graph. \textbf{35}(4) (2016)

\bibitem{merzbach2020bonn}
Merzbach, S., Klein, R.: Bonn appearance benchmark.
\newblock The Eurographics Association (2020)

\bibitem{mildenhall2020nerf}
Mildenhall, B., Srinivasan, P.P., Tancik, M., Barron, J.T., Ramamoorthi, R.,
  Ng, R.: Nerf: Representing scenes as neural radiance fields for view
  synthesis.
\newblock In: Proceedings of the IEEE Conference on Computer Vision and Pattern
  Recognition (2020)

\bibitem{muller2006spherical}
M{\"u}ller, C.: Spherical harmonics, vol.~17.
\newblock Springer (2006)

\bibitem{narihira2015direct}
Narihira, T., Maire, M., Yu, S.X.: Direct intrinsics: Learning albedo-shading
  decomposition by convolutional regression.
\newblock In: Proceedings of the IEEE international conference on computer
  vision, pp. 2992--2992 (2015)

\bibitem{narihira2015judgement}
Narihira, T., Maire, M., Yu, S.X.: Learning lightness from human judgement on
  relative reflectance.
\newblock In: Proceedings of the IEEE Conference on Computer Vision and Pattern
  Recognition, pp. 2965--2973 (2015)

\bibitem{nestmeyer2017reflectance}
Nestmeyer, T., Gehler, P.V.: Reflectance adaptive filtering improves intrinsic
  image estimation.
\newblock In: Proceedings of the IEEE Conference on Computer Vision and Pattern
  Recognition, pp. 6789--6798 (2017)

\bibitem{nestmeyer2020relight}
Nestmeyer, T., Lalonde, J.F., Matthews, I., Lehrmann, A.: Learning
  physics-guided face relighting under directional light.
\newblock In: Proceedings of the IEEE Conference on Computer Vision and Pattern
  Recognition (2020)

\bibitem{newell2016stacked}
Newell, A., Yang, K., Deng, J.: Stacked hourglass networks for human pose
  estimation.
\newblock In: Proceedings of the European Conference on Computer Vision (ECCV),
  pp. 483--499. Springer (2016)

\bibitem{nimier2019mitsuba}
Nimier-David, M., Vicini, D., Zeltner, T., Jakob, W.: Mitsuba 2: A retargetable
  forward and inverse renderer.
\newblock ACM Transactions on Graphics (TOG) \textbf{38}(6), 1--17 (2019)

\bibitem{oh2001image}
Oh, B.M., Chen, M., Dorsey, J., Durand, F.: Image-based modeling and photo
  editing.
\newblock In: Proceedings of the 28th annual conference on Computer graphics
  and interactive techniques, pp. 433--442 (2001)

\bibitem{omer2004color}
Omer, I., Werman, M.: Color lines: Image specific color representation.
\newblock In: Proceedings of the IEEE Conference on Computer Vision and Pattern
  Recognition, vol.~2, pp. II--II. IEEE (2004)

\bibitem{patow2003survey}
Patow, G., Pueyo, X.: A survey of inverse rendering problems.
\newblock In: Computer Graphics Forum, vol.~22, pp. 663--687 (2003)

\bibitem{Pharr16}
Pharr, M., Jakob, W., Humphreys, G.: Physically Based Rendering: From Theory to
  Implementation, 3rd edn.
\newblock Morgan Kaufmann Publishers Inc., San Francisco, CA, USA (2016)

\bibitem{poole2016fast}
Poole, B., Barron, J.T.: The fast bilateral solver.
\newblock In: Proceedings of the European Conference on Computer Vision (ECCV),
  pp. 617--632 (2016)

\bibitem{Ramamoorthi01}
Ramamoorthi, R., Hanrahan, P.: An efficient representation for irradiance
  environment maps.
\newblock In: Proceedings of the 28th Annual Conference on Computer Graphics
  and Interactive Techniques, SIGGRAPH '01, p. 497–500. Association for
  Computing Machinery, New York, NY, USA (2001)

\bibitem{rebuffi2017learning}
Rebuffi, S.A., Bilen, H., Vedaldi, A.: Learning multiple visual domains with
  residual adapters.
\newblock In: Advances in Neural Information Processing Systems, pp. 506--516
  (2017)

\bibitem{rematas2016deep}
Rematas, K., Ritschel, T., Fritz, M., Gavves, E., Tuytelaars, T.: Deep
  reflectance maps.
\newblock In: Proceedings of the IEEE Conference on Computer Vision and Pattern
  Recognition, pp. 4508--4516 (2016)

\bibitem{rhemann2009perceptually}
Rhemann, C., Rother, C., Wang, J., Gelautz, M., Kohli, P., Rott, P.: A
  perceptually motivated online benchmark for image matting.
\newblock In: 2009 IEEE Conference on Computer Vision and Pattern Recognition,
  pp. 1826--1833. IEEE (2009)

\bibitem{ritschel12}
Ritschel~T Dachsbacher~C, G.T.K.J.: The state of the art in interactive global
  illumination.
\newblock Computer Graphics Forum \textbf{31}(1) (2012)

\bibitem{ronneberger2015u}
Ronneberger, O., Fischer, P., Brox, T.: U-net: Convolutional networks for
  biomedical image segmentation.
\newblock In: International Conference on Medical image computing and
  computer-assisted intervention, pp. 234--241. Springer (2015)

\bibitem{rother2011recovering}
Rother, C., Kiefel, M., Zhang, L., Sch{\"o}lkopf, B., Gehler, P.V.: Recovering
  intrinsic images with a global sparsity prior on reflectance.
\newblock In: Advances in Neural Information Processing Systems, pp. 765--773
  (2011)

\bibitem{Seitz2005}
Seitz, S., Matsushita, Y., Kutulakos, K.: A theory of inverse light transport.
\newblock In: Tenth IEEE International Conference on Computer Vision (ICCV'05)
  Volume 1, vol.~2, pp. 1440--1447 Vol. 2 (2005)

\bibitem{sengupta2019neural}
Sengupta, S., Gu, J., Kim, K., Liu, G., Jacobs, D.W., Kautz, J.: Neural inverse
  rendering of an indoor scene from a single image.
\newblock In: Proceedings of the IEEE International Conference on Computer
  Vision, pp. 8598--8607 (2019)

\bibitem{sengupta2018sfsnet}
Sengupta, S., Kanazawa, A., Castillo, C.D., Jacobs, D.W.: Sfsnet: Learning
  shape, reflectance and illuminance of facesin the wild'.
\newblock In: Proceedings of the IEEE Conference on Computer Vision and Pattern
  Recognition, pp. 6296--6305 (2018)

\bibitem{shen2011intrinsic}
Shen, L., Yeo, C.: Intrinsic images decomposition using a local and global
  sparse representation of reflectance.
\newblock In: CVPR 2011, pp. 697--704. IEEE (2011)

\bibitem{shi2017learning}
Shi, J., Dong, Y., Su, H., Yu, S.X.: Learning non-lambertian object intrinsics
  across shapenet categories.
\newblock In: Proceedings of the IEEE Conference on Computer Vision and Pattern
  Recognition, pp. 1685--1694 (2017)

\bibitem{shu2017neural}
Shu, Z., Yumer, E., Hadap, S., Sunkavalli, K., Shechtman, E., Samaras, D.:
  Neural face editing with intrinsic image disentangling.
\newblock In: Proceedings of the IEEE Conference on Computer Vision and Pattern
  Recognition, pp. 5541--5550 (2017)

\bibitem{simonyan2014very}
Simonyan, K., Zisserman, A.: Very deep convolutional networks for large-scale
  image recognition.
\newblock arXiv preprint arXiv:1409.1556  (2014)

\bibitem{Sloan02}
Sloan, P.P., Kautz, J., Snyder, J.: Precomputed radiance transfer for real-time
  rendering in dynamic, low-frequency lighting environments.
\newblock ACM Transactions on Graphics (TOG) \textbf{21}(3), 527–536 (2002)

\bibitem{song2017semantic}
Song, S., Yu, F., Zeng, A., Chang, A.X., Savva, M., Funkhouser, T.: Semantic
  scene completion from a single depth image.
\newblock In: Proceedings of the IEEE Conference on Computer Vision and Pattern
  Recognition, pp. 1746--1754 (2017)

\bibitem{sun2019singleimageportrait}
Sun, T., Barron, J.T., Tsai, Y.T., Xu, Z., Yu, X., Fyffe, G., Rhemann, C.,
  Busch, J., Debevec, P., Ramamoorthi, R.: Single image portrait relighting.
\newblock ACM Transactions on Graphics (TOG) \textbf{38}(4) (2019)

\bibitem{sunkavalli2007factored}
Sunkavalli, K., Matusik, W., Pfister, H., Rusinkiewicz, S.: Factored time-lapse
  video.
\newblock In: ACM SIGGRAPH 2007 papers, pp. 101--es (2007)

\bibitem{tappen2003recovering}
Tappen, M.F., Freeman, W.T., Adelson, E.H.: Recovering intrinsic images from a
  single image.
\newblock In: Advances in Neural Information Processing Systems, pp. 1367--1374
  (2003)

\bibitem{tewari2019fml}
Tewari, A., Bernard, F., Garrido, P., Bharaj, G., Elgharib, M., Seidel, H.P.,
  P{\'e}rez, P., Z{\"o}llhofer, M., Theobalt, C.: {FML: Face Model Learning
  from Videos}.
\newblock In: Proceedings of the IEEE Conference on Computer Vision and Pattern
  Recognition, pp. 10812--10822 (2019)

\bibitem{tewari2020starneural}
Tewari, A., Fried, O., Thies, J., Sitzmann, V., Lombardi, S., Sunkavalli, K.,
  Martin-Brualla, R., Simon, T., Saragih, J., Nießner, M., Pandey, R.,
  Fanello, S., Wetzstein, G., Zhu, J.Y., Theobalt, C., Agrawala, M., Shechtman,
  E., Goldman, D.B., Zollhöfer, M.: {State of the Art on Neural Rendering}.
\newblock Computer Graphics Forum \textbf{39}(2), 701--727 (2020)

\bibitem{tewari2018self}
Tewari, A., Zollh{\"o}fer, M., Garrido, P., Bernard, F., Kim, H., P{\'e}rez,
  P., Theobalt, C.: Self-supervised multi-level face model learning for
  monocular reconstruction at over 250 hz.
\newblock In: Proceedings of the IEEE Conference on Computer Vision and Pattern
  Recognition, pp. 2549--2559 (2018)

\bibitem{tewari17MoFA}
Tewari, A., Zoll{\"o}fer, M., Kim, H., Garrido, P., Bernard, F., Perez, P.,
  Christian, T.: {MoFA: Model-based Deep Convolutional Face Autoencoder for
  Unsupervised Monocular Reconstruction}.
\newblock In: Proceedings of the IEEE International Conference on Computer
  Vision (2017)

\bibitem{tomasi1998bilateral}
Tomasi, C., Manduchi, R.: Bilateral filtering for gray and color images.
\newblock In: Proceedings of the IEEE International Conference on Computer
  Vision, pp. 839--846. IEEE (1998)

\bibitem{tominaga1994dichromatic}
Tominaga, S.: Dichromatic reflection models for a variety of materials.
\newblock Color Research \& Application \textbf{19}(4), 277--285 (1994)

\bibitem{torrance1967}
Torrance, K.E., Sparrow, E.M.: Theory for off-specular reflection from
  roughened surfaces.
\newblock Josa \textbf{57}(9), 1105--1114 (1967)

\bibitem{vidaurre_wacv2019}
Vidaurre, R., Casas, D., Garces, E., Lopez-Moreno, J.: Brdf estimation of
  complex materials with nested learning.
\newblock In: IEEE Winter Conference on Applications of Computer Vision (WACV)
  (2019)

\bibitem{Wang09}
Wang, J., Ren, P., Gong, M., Snyder, J., Guo, B.: All-frequency rendering of
  dynamic, spatially-varying reflectance \textbf{28}(5), 1–10 (2009)

\bibitem{wang2016cost}
Wang, K., Zhang, D., Li, Y., Zhang, R., Lin, L.: Cost-effective active learning
  for deep image classification.
\newblock IEEE Transactions on Circuits and Systems for Video Technology
  \textbf{27}(12), 2591--2600 (2016)

\bibitem{wang2018non}
Wang, X., Girshick, R., Gupta, A., He, K.: Non-local neural networks.
\newblock In: Proceedings of the IEEE Conference on Computer Vision and Pattern
  Recognition, pp. 7794--7803 (2018)

\bibitem{weiss2001deriving}
Weiss, Y.: Deriving intrinsic images from image sequences.
\newblock In: Proceedings of the IEEE International Conference on Computer
  Vision, vol.~2, pp. 68--75. IEEE (2001)

\bibitem{yamaguchi2018high}
Yamaguchi, S., Saito, S., Nagano, K., Zhao, Y., Chen, W., Olszewski, K.,
  Morishima, S., Li, H.: High-fidelity facial reflectance and geometry
  inference from an unconstrained image.
\newblock ACM Transactions on Graphics (TOG) \textbf{37}(4), 1--14 (2018)

\bibitem{Ye2014}
Ye, G., Garces, E., Liu, Y., Dai, Q., Gutierrez, D.: Intrinsic video and
  applications.
\newblock ACM Trans. Graph. \textbf{33}(4) (2014)

\bibitem{yu2019inverserendernet}
Yu, Y., Smith, W.A.: Inverserendernet: Learning single image inverse rendering.
\newblock In: Proceedings of the IEEE Conference on Computer Vision and Pattern
  Recognition, pp. 3155--3164 (2019)

\bibitem{zhang2020resnest}
Zhang, H., Wu, C., Zhang, Z., Zhu, Y., Zhang, Z., Lin, H., Sun, Y., He, T.,
  Mueller, J., Manmatha, R., et~al.: Resnest: Split-attention networks.
\newblock arXiv preprint arXiv:2004.08955  (2020)

\bibitem{zhang2018unreasonable}
Zhang, R., Isola, P., Efros, A.A., Shechtman, E., Wang, O.: The unreasonable
  effectiveness of deep features as a perceptual metric.
\newblock In: Proceedings of the IEEE Conference on Computer Vision and Pattern
  Recognition, pp. 586--595 (2018)

\bibitem{zhang2020image}
Zhang, Y., Chen, W., Ling, H., Gao, J., Zhang, Y., Torralba, A., Fidler, S.:
  Image gans meet differentiable rendering for inverse graphics and
  interpretable 3d neural rendering.
\newblock arXiv preprint arXiv:2010.09125  (2020)

\bibitem{zhao2012closed}
Zhao, Q., Tan, P., Dai, Q., Shen, L., Wu, E., Lin, S.: A closed-form solution
  to retinex with nonlocal texture constraints.
\newblock IEEE Transactions on Pattern Analysis and Machine Intelligence
  \textbf{34}(7), 1437--1444 (2012)

\bibitem{zhao2020physics}
Zhao, S., Jakob, W., Li, T.M.: Physics-based differentiable rendering: from
  theory to implementation.
\newblock In: ACM SIGGRAPH 2020 Courses, pp. 1--30 (2020)

\bibitem{zhou2019deeprelight}
Zhou, H., Hadap, S., Sunkavalli, K., Jacobs, D.W.: {Deep Single-Image Portrait
  Relighting}.
\newblock In: Proceedings of the IEEE International Conference on Computer
  Vision, pp. 7194--7202 (2019)

\bibitem{zhou2019glosh}
Zhou, H., Yu, X., Jacobs, D.W.: Glosh: Global-local spherical harmonics for
  intrinsic image decomposition.
\newblock In: Proceedings of the IEEE International Conference on Computer
  Vision, pp. 7820--7829 (2019)

\bibitem{zhou2015learning}
Zhou, T., Krahenbuhl, P., Efros, A.A.: Learning data-driven reflectance priors
  for intrinsic image decomposition.
\newblock In: Proceedings of the IEEE International Conference on Computer
  Vision, pp. 3469--3477 (2015)

\bibitem{zollhofer2018state}
Zollh{\"o}fer, M., Thies, J., Garrido, P., Bradley, D., Beeler, T., P{\'e}rez,
  P., Stamminger, M., Nie{\ss}ner, M., Theobalt, C.: State of the art on
  monocular 3d face reconstruction, tracking, and applications.
\newblock In: Computer Graphics Forum, vol.~37, pp. 523--550 (2018)

\bibitem{zoran2015learning}
Zoran, D., Isola, P., Krishnan, D., Freeman, W.T.: Learning ordinal
  relationships for mid-level vision.
\newblock In: Proceedings of the IEEE International Conference on Computer
  Vision, pp. 388--396 (2015)

\end{thebibliography}

\end{document}